\documentclass[manuscript,screen,nonacm]{acmart}

\usepackage{amsmath}
\usepackage{amssymb} 
\usepackage{mathtools}
\usepackage{amsthm}
\usepackage{algorithm}
\usepackage{amsfonts}
\usepackage[utf8]{inputenc} % allow utf-8 input
\usepackage[T1]{fontenc}    % use 8-bit T1 fonts
\usepackage{hyperref}       % hyperlinks
\usepackage{url}            % simple URL typesetting
\usepackage{booktabs}       % professional-quality tables
\usepackage{nicefrac}       % compact symbols for 1/2, etc.
\usepackage{microtype}      % microtypography
\usepackage{xcolor}         % colors
\usepackage{enumerate}
\usepackage{multicol}
\usepackage{subfigure}
\usepackage{bm}
\usepackage{tabularx}
\usepackage{enumitem}
\usepackage{multirow}
\usepackage{caption}
\usepackage{graphicx}
\usepackage{bbding}
\usepackage{wrapfig}
\usepackage{lipsum}
\usepackage{multibib}
\usepackage{tcolorbox}
\usepackage{booktabs}
\usepackage{makecell}
\usepackage{amsmath,amsfonts}
 
\usepackage{textcomp}
\usepackage{array}
\usepackage{booktabs} % for professional tables
\usepackage{url}
\usepackage{framed}
\usepackage{color}
\usepackage[utf8]{inputenc} % allow utf-8 input
\usepackage{url}            % simple URL typesetting
\usepackage{booktabs}       % professional-quality tables
\usepackage{nicefrac}       % compact symbols for 1/2, etc.
\usepackage{microtype}      % microtypography
\usepackage{xcolor}         % colors
\usepackage{color, colortbl}
\definecolor{greyC}{RGB}{180,180,180}
\definecolor{boxgrey}{RGB}{242,242,242} % Fig.4 shared-prior node fill
\definecolor{greyL}{RGB}{235,235,235}
\usepackage{multicol}
\usepackage{multirow}
\usepackage{ulem}
\usepackage{soul}
\usepackage{microtype}
\usepackage{graphicx}
\usepackage{booktabs}
\usepackage{caption}
\usepackage{algorithm}
\usepackage{algpseudocode}
\usepackage{threeparttable}
\usepackage{subfigure}
\usepackage{enumerate}
\usepackage{natbib}
\usepackage{wrapfig}  
\usepackage{framed}
\usepackage{color}
\usepackage{tcolorbox}
\usepackage{sidecap}
\usepackage{tikz}
\usepackage[edges]{forest}

\newcommand{\revise}[1]{#1}
\newcommand{\revcolor}{}
\DeclareMathOperator*{\argminA}{arg\,min}

\definecolor{citationpurple}{RGB}{102,45,145}
\definecolor{urlteal}{RGB}{0,105,105}
\hypersetup{
  colorlinks=true,
  linkcolor=black,
  citecolor=citationpurple,
  urlcolor=urlteal
}

\graphicspath{{./}}

\usepackage{placeins}
\usepackage{stfloats}

\newcommand{\tabgap}{}

\makeatletter
\newcommand{\setcanvas}[3]{%
  \expandafter\gdef\csname cv@#1@t\endcsname{#2}%
  \expandafter\gdef\csname cv@#1@b\endcsname{#3}}
\newcommand{\artwork}[1]{\@ifundefined{cv@#1@t}{}{\vspace{-\csname cv@#1@t\endcsname}}}
\newcommand{\artworkend}[1]{\@ifundefined{cv@#1@b}{}{\vspace{-\csname cv@#1@b\endcsname}}}
\makeatother

\setcanvas{fig:MIAs-image}                    {13.6pt}{13.6pt}  % pipline_reference.pdf
\setcanvas{fig:taxonomy}                  {0.9pt} {10.6pt}  % tikz forest (measured from the build)
\setcanvas{fig:MIAs-different-attack}         {11.9pt}{16.8pt}  % different_attack_reference.pdf
\setcanvas{fig:MIAs-attack-img-timeline}  {2.0pt}{2.2pt}  % MIA_attack_image_combine_updated.pdf
\setcanvas{fig:imagemi_demo_class}            {2.1pt}{2.3pt}  % MIA_image.pdf
\setcanvas{fig:MIAs-attack-text-timeline}     {5.4pt}{5.2pt}  % MIA_attack_text_combine_updated.pdf
\setcanvas{fig:text_attack}                   {2.1pt}{2.3pt}  % MIA_text.pdf
\setcanvas{fig:MIAs-attack-graph-timeline}    {2.0pt}{2.2pt}  % MIA_attack_graph_combine_updated.pdf
\setcanvas{fig:graph_attack}                  {1.9pt}{2.6pt}  % MIA_graph.pdf
\setcanvas{fig:taxonomy_defense}              {2.9pt}{9.6pt}  % MIA_defense_taxo_updated.pdf
\setcanvas{fig:img_defense}                   {2.1pt}{2.2pt}  % DEFENSE_image.pdf
\setcanvas{fig:text_defense}                  {2.1pt}{2.1pt}  % DEFENSE_text.pdf
\setcanvas{fig:graph_defense}                 {2.2pt}{2.4pt}  % DEFENSE_graph.pdf
\theoremstyle{plain}
\newtheorem{theorem}{Theorem}[section]

\newtheorem{definition}[theorem]{Definition}

\newtheorem{remark}{Remark}[section]
\newtheorem{principle}{Principle}

\definecolor{shadecolor}{rgb}{0.92,0.92,0.92}
\definecolor{LightGray}{HTML}{F0F1F2}
\definecolor{hidden-red}{RGB}{205, 44, 36}
\definecolor{hidden-blue}{RGB}{194,232,247}
\definecolor{hidden-orange}{RGB}{243,202,120}
\definecolor{hidden-green}{RGB}{34,139,34}
\definecolor{hidden-pink}{RGB}{255,245,247}
\definecolor{hidden-black}{RGB}{20,68,106}
\definecolor{dataset-image-bg}{HTML}{E4F1FA}
\definecolor{dataset-text-bg}{HTML}{FFF0D8}
\definecolor{dataset-graph-bg}{HTML}{E9F4E5}

\newcolumntype{L}[1]{>{\raggedright\let\newline\\\arraybackslash\hspace{0pt}}m{#1}}
\newsavebox{\compactdatasetrevisionbox}
\newcolumntype{C}[1]{>{\centering\let\newline  \\\arraybackslash\hspace{0pt}}m{#1}}%
\newcolumntype{R}[1]{>{\raggedleft\let\newline \\\arraybackslash\hspace{0pt}}m{#1}}

\newcommand{\eg}{\textit{e.g., }}

\newcommand{\thetav}{\boldsymbol{\theta}}
\newcommand{\phiv}{\boldsymbol{\phi}}

\AtBeginDocument{%
  \providecommand\BibTeX{{%
    \normalfont B\kern-0.5em{\scshape i\kern-0.25em b}\kern-0.8em\TeX}}}

\setcopyright{none}
\renewcommand\footnotetextcopyrightpermission[1]{}

\begin{document}
\title{Model Inversion Attacks: A Survey of Approaches and Countermeasures}

\author{Zhanke Zhou}
\authornote{These authors contributed equally to this research.}
\affiliation{\institution{Hong Kong Baptist University}\country{Hong Kong SAR}}

\author{Jianing Zhu}
\authornotemark[1]
\affiliation{\institution{Hong Kong Baptist University}\country{Hong Kong SAR}}

\author{Fengfei Yu}
\authornotemark[1]
\affiliation{\institution{Hong Kong Baptist University}\country{Hong Kong SAR}}

\author{Xuan Li}
\affiliation{\institution{Hong Kong Baptist University}\country{Hong Kong SAR}}

\author{Xiong Peng}
\affiliation{\institution{Hong Kong Baptist University}\country{Hong Kong SAR}}

\author{Tongliang Liu}
\affiliation{\institution{ The
University of Sydney}\country{Australia}}

\author{Bo Han}
% \authornote{Corresponding author.}
\affiliation{\institution{Hong Kong Baptist University}\country{Hong Kong SAR}}

%%
%% By default, the full list of authors will be used in the page
%% headers. Often, this list is too long, and will overlap
%% other information printed in the page headers. This command allows
%% the author to define a more concise list
%% of authors' names for this purpose.

% \renewcommand{\shortauthors}{Trovato and Tobin, et al.}

\begin{abstract}
\revise{Deep neural networks have enabled numerous studies and applications on both Euclidean data, such as images and text, and non-Euclidean data, such as graphs. Because these networks may process private data, their deployment raises concerns about privacy leakage. Model inversion attacks (MIAs) exploit access to a trained model to reconstruct training examples or infer privacy-sensitive characteristics represented by the model.} The effectiveness of MIAs has been demonstrated in various domains, including \revise{images, text,} and graphs. These attacks highlight the vulnerability of neural networks and raise awareness about the risk of privacy leakage within the research community. \revise{This survey provides a threat-model- and assumption-aware synthesis of attacks and defenses. We compare exposed interfaces, attacker knowledge, reconstruction priors and targets, failure modes, deployment constraints, and privacy--utility trade-offs, while highlighting modeling principles, optimization challenges, and future directions.}
\revise{We also maintain an evolving repository of relevant research at}
\url{https://github.com/AndrewZhou924/Awesome-model-inversion-attack}.
\end{abstract}

%%
%% The code below is generated by the tool at http://dl.acm.org/ccs.cfm.
%% Please copy and paste the code instead of the example below.
%%

% \ccsdesc[100]{Machine learning}
% \ccsdesc[100]{Artificial intelligence}
% \ccsdesc[300]{Machine learning}
% \ccsdesc[300]{Computer systems organization~Redundancy}
% \ccsdesc{Computer systems organization~Robotics}
% \ccsdesc[100]{Networks~Network reliability}

%% Keywords. The author(s) should pick words that accurately describe
%% the work being presented. Separate the keywords with commas.

% \received{20 February 2007}
% \received[revised]{12 March 2009}
% \received[accepted]{5 June 2009}

\maketitle

\section{Introduction}

% Background
\revise{Machine learning has driven advances in pattern recognition for Euclidean data, \eg images and text, and non-Euclidean data, commonly represented as graphs. Convolutional neural networks (CNNs)~\cite{he2016deep,huang2017densely,he2017mask,dai2017deformable,tan2019efficientnet}, language models (LMs)~\cite{vaswani2017attention,devlin2018bert,liu2019roberta,thoppilan2022lamda,chowdhery2022palm}, and graph neural networks (GNNs)~\cite{kipf2016semi,kipf2016variational,velivckovic2017graph,hamilton2017inductive,xu2018powerful} have been developed for these domains. More recently, foundation models~\cite{brown2020language,chowdhery2022palm,achiam2023gpt,touvron2023llama,team2023gemini} have attracted growing interest because of their broad capabilities in face recognition~\cite{taigman2014deepface,sun2015deepid3,schroff2015facenet,deng2019arcface,liu2017sphereface}, dialogue~\cite{thoppilan2022lamda,zhang2019dialogpt,adiwardana2020towards,li2016persona}, recommendation~\cite{sun2019multi,he2020lightgcn,chen2020revisiting}, and drug discovery~\cite{wieder2020compact,li2022deep,zhang2022graph,ma2022cross,li2023long}.}

\begin{figure}[t!]
    \centering
    \artwork{fig:MIAs-image}
    \includegraphics[width=0.95\linewidth]{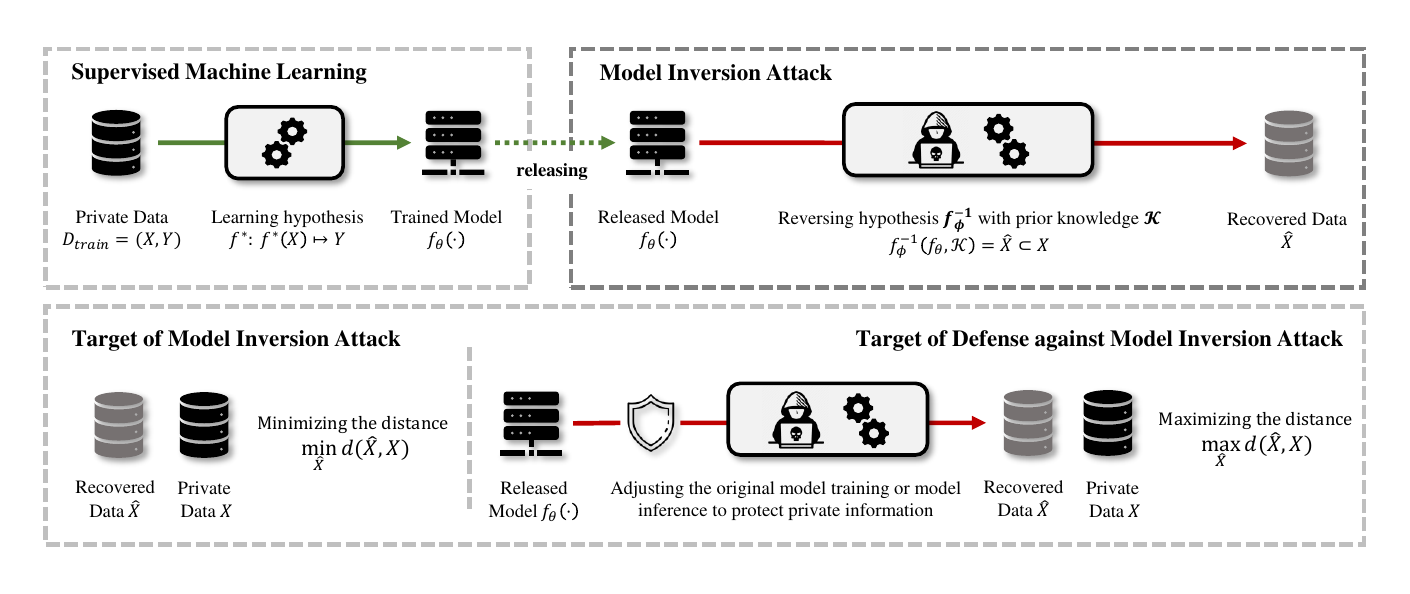}
    \artworkend{fig:MIAs-image}
    \caption{
    \revise{Pipeline of an MIA in supervised learning: the attacker uses a released model and prior knowledge to recover private training information, while the defender limits recovery without sacrificing the target model's utility on legitimate inputs.}
    }
    \Description[Pipeline illustration of Model Inversion Attack]{Pipeline illustration of Model Inversion Attack}
    \label{fig:MIAs-image}

\end{figure}

% The research problem
%%%%%%%%%%%%%%%%%%%%%%%%
\revise{Nevertheless, these applications process sensitive data, including private facial images, contact information, and personalized social networks, raising concerns about privacy leakage~\cite{song2022survey,dibbo2023sok,hu2022membership,mehnaz2022your,zhang2010privacy,akcora2012privacy} and motivating privacy-preserving methods~\cite{dwork2014algorithmic,xu2015privacy,clifton2002tools,hu2020personalized,gong2015privacy,shan2018practical}.}
\revise{Model inversion attacks (MIAs)~\cite{fredrikson2015model} exploit black- or white-box access to recover training examples or sensitive attributes across data domains (Fig.~\ref{fig:MIAs-image}).}

% The bibliography is driven by citations in the manuscript only.

% A further introduction to MIAs
%%%%%%%%%%%%%%%%%%%%%%%%
\revise{MIAs exploit model outputs or internals to infer private training information in settings such as:}
\begin{itemize}[leftmargin=*]
\item 
\revise{\textit{Facial recognition}~\cite{fredrikson2015model, yang2019neural, zhang2020secret, chen2021knowledge, wang2021variational, struppek2022ppa, MIRROR, BREPMI, LOKT, LOMMA, PLGMI, han2023reinforcement}, where an adversary updates candidate images using model responses to recover identity-revealing facial features or representative private images.}

\item 
\revise{\textit{Medical diagnostics}~\cite{zhang2020secret,chen2021knowledge,wang2021variational,PLGMI}, where clinical-image models may expose sensitive class or patient information.}

\item 
\revise{\textit{Social and recommendation systems}~\cite{chanpuriya2021deepwalking, shen2022finding}, where graph-link reconstruction can expose private relationships.}
\end{itemize}

\revise{Depending on the threat model, MIAs combine target access with nonsensitive auxiliary knowledge to recover images~\citep{zhang2020secret,struppek2022ppa}, private text~\citep{zhang2022text}, or graph structure~\citep{he2021stealing,zhang2021graphmi}; label-only black-box recovery is also feasible~\citep{BREPMI}.}

% Explain why we do a survey
%%%%%%%%%%%%%%%%%%%%%%%%
\revise{Many attack approaches have been proposed across image, text, and graph data~\cite{fredrikson2015model,zhang2020secret,BREPMI,zhang2022text,morris2023text,he2021stealing,zhang2021graphmi}; representative defenses include output restriction, representation regularization, training-time protection, and graph perturbation~\cite{MID,BiDO,yang2020defending,liu2024ensembler,hsieh2021netfense,hidano2022degree}. Recent surveys by \citet{fang2024privacy} and \citet{yang2025deep} already cover multiple data domains, attack taxonomies, defenses, datasets, and metrics. Our complementary focus is to compare methods through a common set of threat-model and methodological dimensions---exposed interface, attacker knowledge, auxiliary data and priors, reconstruction target, failure mode, deployment constraint, and privacy--utility boundary.}

This work provides \revise{a threat-model- and assumption-aware survey} of model inversion attacks.
Our contributions are:
\begin{itemize}[leftmargin=*]
\item
\revise{We give a threat-model-centric account of MIAs, separate them from gradient inversion, membership inference, and model stealing, and position this survey against the closest prior ones (Sec.~\ref{sec: overview}, Tab.~\ref{tab:survey-comparison}).}

\item
\revise{We compare every attack along one frame---exposed interface, assumed knowledge and prior, recovered target, failure conditions---across the image, text, graph, collaborative/split and vision-language settings (Secs.~\ref{sec: MIAs-image}--\ref{sec: MIAs-graph}).}

\item
\revise{We quantify what defenses cost, from the privacy--utility evidence their own papers report (Tab.~\ref{tab:defense-tradeoff}) to the deployment and query budgets an attacker faces (Tab.~\ref{tab:deployment-matrix}), and summarize datasets, metrics and open problems (Secs.~\ref{sec: defending-MIAs}--\ref{sec: discussion}).}

\end{itemize}

\section{Overview}
\label{sec: overview}

% Finally, we introduce the organization of our content with a comprehensive taxonomy in Section~\ref{sec: overview_taxonomy}. 

\subsection{The Problem Definition of Model Inversion Attacks}
\label{sec: problem-def}

\begin{definition}[Supervised machine learning]
    \label{def: supervised-ml}
    \revise{Given training and test sets $D_{\text{train}}=(X_{\text{train}},Y_{\text{train}})$ and $D_{\text{test}}=(X_{\text{test}},Y_{\text{test}})$, supervised learning estimates $f_{\thetav}\in\mathcal{H}$ to map inputs $X$ to labels or targets $Y$. Its parameters are commonly optimized on $D_{\text{train}}$ by stochastic gradient descent, and its generalization is evaluated on $D_{\text{test}}$ before deployment.}
\end{definition}

\begin{definition}[Model inversion attacks]
\label{def: MI-attack}
Given a trained model $f_{\thetav}$ and prior knowledge $\mathcal{K}$, \revise{a model inversion attack seeks} a reverse hypothesis $f^{-1}_{\phiv}$ that \revise{recovers training data $X_{\text{train}}$ or privacy-sensitive characteristics represented by $f_{\thetav}$}.
\revise{For sample-reconstruction attacks}, $f^{-1}_{\phiv}(f_{\thetav}, \mathcal{K}) = \hat{X}_{\text{train}}$, where
the recovered data $\hat{X}_{\text{train}}$ is a set of data samples that are expected to approximate those samples in $X_{\text{train}}$.
\end{definition}

\begin{remark}[Strict model inversion attacks]
When no prior knowledge is given, namely, 
\revise{$\mathcal{K} = \emptyset$, then an MIA reduces to}
finding the reverse hypothesis $f^{-1}_{\phiv}$ that 
$f^{-1}_{\phiv}(f_{\thetav}) = \hat{X}_{\text{train}}$.
\revise{This is a limiting case: pioneering studies typically used auxiliary information such as known nonsensitive attributes, a target identity or class, or public-data priors}~\citep{fredrikson2014privacy,fredrikson2015model}.
\end{remark}

\begin{remark}[The extent of inversion]
\revise{MIAs usually recover representative examples, attributes, graph relations, or a subset of $X_{\text{train}}$, rather than the entire training set. Evaluation therefore uses domain-appropriate measures of fidelity, attack success, distributional similarity, or recall. Full-set reconstruction is obtained only when $\hat{X}_{\text{train}}$ covers all samples in $X_{\text{train}}$.}
\end{remark}

%%%%%%% taxonomy %%%%%%%
\definecolor{root-yellow}{HTML}{FFDC98}
\definecolor{attack-red}{HTML}{F6B59A}
% \definecolor{defense-green}{HTML}{A3CC90}
\definecolor{defense-green}{HTML}{9fc5e8}

\tikzstyle{my-box}=[
    rectangle,
    draw=black, % box edge color
    rounded corners,
    text opacity=1,
    minimum height=1.5em,
    minimum width=5em,
    inner sep=2pt,
    align=center,
    fill opacity=.5,
]
\tikzstyle{leaf}=[
    my-box, 
    minimum height=1.5em,
    fill=LightGray!30,
    % hidden-orange!360, 
    % fill=hidden-grey!30, 
    text=black,
    align=left,
    font=\normalsize,
    inner xsep=2pt,
    inner ysep=4pt,
]

\begin{figure*}[t!]
    \centering
    \artwork{fig:taxonomy}
    \resizebox{\textwidth}{!}{
        \sloppy
        \begin{forest}
            forked edges,
            for tree={
                grow=east,
                reversed=true,
                anchor=base west,
                parent anchor=east,
                child anchor=west,
                base=left,
                font=\large,
                rectangle,
                draw=black, % box edge color
                rounded corners,
                align=left,
                minimum width=4em,
                edge+={darkgray, line width=1.5pt},
                s sep=3pt,
                inner xsep=2pt,
                inner ysep=3pt,
                line width=0.8pt,
                ver/.style={rotate=90, child anchor=north, parent anchor=south, anchor=center},
            },
            where level=1{text width=3.4em,font=\normalsize,}{},
            where level=2{text width=14.2em,font=\normalsize,}{},
            where level=3{text width=8.8em,font=\normalsize,}{},
            where level=4{text width=6.2em,font=\normalsize,}{},
            where level=5{text width=12em,font=\normalsize,}{},
            [
               \quad Model Inversion Adversaries \quad, ver, fill=root-yellow
                [
                Attack, fill=attack-red
                    [
                        Image Domain (Sec.~\ref{sec: MIAs-image}), fill=attack-red!80
                        [
                            Optimization-based, fill=attack-red!60 %(\S \ref{sec:mwp})
                            [   
                                White-box, fill=attack-red!40
                                [
                                    \eg Vanilla-MI~\cite{fredrikson2015model}{,}
                                    GMI~\cite{zhang2020secret}{,}
                                    KEDMI~\citep{chen2021knowledge}{,}
                                    VMI~\citep{wang2021variational}{,}
                                    PPA~\citep{struppek2022ppa}{,} etc{.}
                                    , leaf, text width=32em
                                ]
                            ]
                            [
                                Black-box, fill=attack-red!40
                                [
                                    \eg MIRROR~\citep{MIRROR}{,} 
                                    BREP-MI~\citep{BREPMI}{,}  LOKT~\citep{LOKT}{,} and RLBMI~\citep{han2023reinforcement}{.}
                                    , leaf, text width=32em, fill=gray!25
                                ]
                            ]
                        ]
                        [
                            Training-based, fill=attack-red!60 %(\S \ref{sec:tp})
                            [
                                Black-box, fill=attack-red!40
                                [
                                    \eg LB-MI~\cite{yang2019neural}{,} and
                                    XAI~\citep{zhao2021exploiting}{.}
                                    , leaf, text width=32em, fill=gray!25
                                ]
                            ]
                        ]
                        % [
                        %     Architecture (\S \ref{sec:gps})
                        %     [   
                        %         White-box
                        %         [
                        %             \eg MathQA~\cite{amini2019mathqa}{,}
                        %             SVAMP~\cite{patel2021nlp}
                        %             , leaf, text width=32em
                        %         ]
                        %     ]
                        %     [
                        %         Black-box
                        %         [
                        %             \eg IconQA~\cite{lu2021iconqa}{,}
                        %             TabMWP~\cite{lu2022dynamic}
                        %             , leaf, text width=32em
                        %         ]
                        %     ]
                        % ]
                    ]
                    [
                        \revise{Collaborative/Split (Sec.~\ref{sec:MI_on_models})}, fill=attack-red!80
                        [
                            Optimization-based, fill=attack-red!60
                            [
                                White-box, fill=attack-red!40
                                [
                                    \revise{\eg CI attacks~\citep{he2019model}{,} and
                                    DCI~\citep{dong2021privacy}{.}}
                                    , leaf, text width=32em
                                ]
                            ]
                            [
                                Black-box, fill=attack-red!40
                                [
                                    \revise{\eg UnSplit~\citep{erdougan2022unsplit}{,} and
                                    GLASS~\citep{li2023gan}{.}}
                                    , leaf, text width=32em, fill=gray!25
                                ]
                            ]
                        ]
                        [
                            Training-based, fill=attack-red!60
                            [
                                Black-box, fill=attack-red!40
                                [
                                    \revise{\eg FSHA~\citep{pasquini2021unleashing}{,}
                                    GINVER~\citep{yin2023ginver}{,}
                                    PCAT~\citep{gao2023pcat}{,}
                                    SDAR~\citep{zhu2025sdar}{,} etc{.}}
                                    , leaf, text width=32em, fill=gray!25
                                ]
                            ]
                        ]
                    ]
                    [
                        Text Domain (Sec.~\ref{sec: MIAs-text}), fill=attack-red!80
                        [
                            Optimization-based, fill=attack-red!60 %(\S \ref{sec:mwp})
                            [   
                                White-box, fill=attack-red!40
                                [
                                    \eg Continuous Relaxation~\citep{song2020information}{,}
                                    CEA~\citep{parikh2022canary}{, and}
                                    Text Revealer~\citep{zhang2022text}{.}
                                    , leaf, text width=32em
                                ]
                            ]
                            [
                                Black-box, fill=attack-red!40
                                [
                                    \revise{\eg Vec2Text~\citep{morris2023text}{, and}
                                    Multilingual Vec2Text~\citep{chen2024text}{.}}
                                    , leaf, text width=32em, fill=gray!25
                                ]
                            ]
                        ]
                        [
                            Training-based, fill=attack-red!60 %(\S \ref{sec:tp})
                            [
                                Black-box, fill=attack-red!40
                                [
                                    \eg MLC/MSP~\citep{song2020information}{,}
                                    GEIA~\citep{li2023sentence}{,}
                                    logit2prompt~\citep{morris2023language}{,} etc{.}
                                    , leaf, text width=32em, fill=gray!25
                                ]
                            ]
                        ]
                    ]
                    [
                        \revise{VLM/Foundation (Sec.~\ref{ssec: MIAs-foundation})}, fill=attack-red!80
                        [
                            Optimization-based, fill=attack-red!60
                            [
                                White-box, fill=attack-red!40
                                [
                                    \revise{\eg SMI-AW~\citep{nguyen2025vlm}{,}
                                    DDMI~\citep{ddmi2025}{,} and
                                    LeakyCLIP~\citep{chen2025leakyclip}{.}}
                                    , leaf, text width=32em
                                ]
                            ]
                        ]
                        [
                            Training-based, fill=attack-red!60
                            [
                                Black-box, fill=attack-red!40
                                [
                                    \revise{\eg Face-Adapter~\citep{otroshi2025faceadapter}{,} and
                                    CapRecover~\citep{caprecover2025}{.}}
                                    , leaf, text width=32em, fill=gray!25
                                ]
                            ]
                        ]
                    ]
                    [
                        Graph Domain (Sec.~\ref{sec: MIAs-graph}), fill=attack-red!80
                        [
                            Optimization-based, fill=attack-red!60  %(\S \ref{sec:mwp})
                            [   
                                White-box, fill=attack-red!40
                                [
                                    \eg GraphMI~\citep{zhang2021graphmi}{,} and
                                    MC-GRA~\citep{zhou2023strengthening}{.}
                                    , leaf, text width=32em
                                ]
                            ]
                            [
                                Black-box, fill=attack-red!40
                                [
                                    \eg Link Stealing Attack~\citep{he2021stealing}{,}
                                    MNEMON~\citep{shen2022finding}{,} and GSEF~\citep{olatunji2022private}{.}
                                    , leaf, text width=32em, fill=gray!25
                                ]
                            ]
                        ]
                        [
                            Training-based, fill=attack-red!60 %(\S \ref{sec:tp})
                            [   
                                White-box, fill=attack-red!40
                                [
                                    \eg DeepWalking Backwards~\citep{chanpuriya2021deepwalking}{,} and
                                    HomoGMI/HeteGMI~\citep{liu2023model}{.}
                                    , leaf, text width=32em
                                ]
                            ]
                            [
                                Black-box, fill=attack-red!40
                                [
                                    \eg Inference Attack~\citep{zhang2022inference}{.}
                                    , leaf, text width=32em, fill=gray!25
                                ]
                            ]
                        ]
                    ]
                ]
                [
                    Defense, fill=defense-green
                    [
                        Image Domain (Sec.~\ref{sec: defend-image}), fill=defense-green!80
                        [   
                                Training-time, fill=defense-green!60
                                [
                                    \eg DP~\cite{fredrikson2014privacy,zhang2020secret}{,}
                                    MID~\cite{MID}{,}
                                    BiDO~\cite{BiDO}{,}
                                    \revise{NegLS}~\cite{LS}{,}
                                    TL-DMI~\cite{ho2024model}{, etc.}
                                    , leaf, text width=32em
                                ]
                            ]
                            [
                                Inference-time, fill=defense-green!60
                                [
                                    \eg PPF~\cite{yang2020defending}{,}
                                    AD-mi~\cite{wen2021defending}{.}
                                    , leaf, text width=32em, fill=gray!25
                                ]
                            ]
                    ]
                    [
                        Text Domain (Sec.~\ref{sec: defend-text}), fill=defense-green!80
                        [   
                                Training-time, fill=defense-green!60
                                [
                                    \eg Dropout~\cite{carlini2019secret}{,}
                                    Adversarial training~\cite{song2020information}{,}
                                    PPM~\cite{pan2020privacy}{,} etc{.}
                                    , leaf, text width=32em
                                ]
                            ]
                            [
                                Inference-time, fill=defense-green!60
                                [
                                    \eg Gaussian noise~\cite{morris2023text}{,}
                                    \revise{Multilingual masking}~\cite{chen2024text}{,} etc{.}
                                    , leaf, text width=32em, fill=gray!25
                                ]
                            ]
                    ]
                    [
                        Graph Domain (Sec.~\ref{sec: defend-graph}), fill=defense-green!80
                        [   
                                Training-time, fill=defense-green!60
                                [
                                    \eg NetFense~\citep{hsieh2021netfense}{, }DPRR~\citep{hidano2022degree}{, }
                                    GAP~\citep{sajadmanesh2023gap}{, }MC-GPB~\citep{zhou2023strengthening}{, etc.}
                                    , leaf, text width=32em
                                ]
                            ]
                            [
                                Inference-time, fill=defense-green!60
                                [
                                    \eg embedding perturbation~\citep{zhang2022inference}{, and} GRASP~\citep{guo2025grasp}{.}
                                    , leaf, text width=32em, fill=gray!25
                                ]
                            ]
                    ]
                ]
            ]
        \end{forest}
    }
    \artworkend{fig:taxonomy}
    \caption{
    \revise{Taxonomy of MIAs and defenses over the image, text and graph domains and over the collaborative/split and vision-language/foundation-model settings, with detailed coverage in Sections~\ref{sec: MIAs-image}--\ref{sec: defending-MIAs}.}}
    \label{fig:taxonomy}
    \Description[Taxonomy of model inversion adversaries]{Taxonomy of model inversion adversaries}

\end{figure*}

%%%%%%% %%%%%%%% %%%%%%%

\revise{\textbf{Definition~\ref{def: MI-attack} has three critical factors:}}

% 
% \textbf{The four critical aspects in definition~\ref{def: MI-attack}:}
% (1) the MIAs method to obtain $f^{-1}_{\phiv}$,
% (2) the white-box or black-box access of $f_{\thetav}$,
% (3) the information contained in $\mathcal{K}$,
% and (4) the data domain of $X_{\text{train}}$, including images, texts, or graphs.

\begin{itemize}[leftmargin=*]

\item 
\textbf{Factor 1: The domain-specific MIA methods to obtain $f^{-1}_{\phiv}$.}
\revise{Fig.~\ref{fig:taxonomy} organizes MIA designs by data domain and by the collaborative/split and vision-language settings that cut across them, because reconstruction difficulty depends on data structure, semantics, and the interface a deployment exposes.}

\item
\textbf{Factor 2: The information contained in $\mathcal{K}$.}
\revise{Many MIA settings assume domain knowledge and public auxiliary data but no private training records~\citep{wang2021variational,LOMMA,struppek2022ppa}; for example, public facial images can provide a generative prior. Stronger threat models assume information related to the private distribution or target records~\cite{zhang2020secret}, such as partial or corrupted observations whose nonsensitive content constrains and narrows recovery in practice.}

\item
\textbf{Factor 3: The white-box or black-box access of $f_{\thetav}$.}
\revise{In a \textit{white-box} attack, the adversary knows the target model's architecture and parameters and can compute gradients through it. In a \textit{black-box} attack, the adversary does not know the architecture or parameters and cannot compute gradients through the target model.}
\revise{Restricted-access MIAs can be grouped by the exposed interface:}
(1) \textit{full confidence scores}, from which the adversary can also compute a target label's prediction loss (\eg cross-entropy loss); 
(2) \textit{hard label only}, where \revise{only the predicted label is returned} and the adversary is given the most limited information;
(3) \textit{hidden representation}: the adversary reconstructs input information from a feature or embedding deliberately exposed by the target-model interface.

\end{itemize}

% \begin{table*}
% \centering
% \caption{Two types of black-box model inversion attacks based on different information provided by the prediction vector.}
% \label{table::black-box-attacks}
% \resizebox{\linewidth}{!}{%
% \begin{tabular}{l|m{13cm}}
% \toprule
% \textbf{Prediction Output} & \textbf{Description}\\
% \hline
% Full confidence scores  & The adversary queries the target classifier with input data and obtains all confidence scores returned by the target classifier. Based on this, the adversary can further calculate the input's prediction loss (e.g., cross-entropy loss) according to the target label that they want to reconstruct.\\
% % \hline
% % Top-K confidence scores & The adversary queries an input record and obtains only top-K confidence scores returned by the target classifier. For example, the adversary only receives the probabilities of the most likely three classes (assuming the total number of classes is much larger than three). \\
% \hline
% Hard label only & The adversary queries an input record and obtains only the predicted label returned by the target classifier. In this case, the adversary is given the most limited information.\\
% \bottomrule
% \end{tabular}
% }
% \end{table*}

\subsection{Principles for Enhancing or Defending against Model Inversion Attacks}
\label{sec: principles}

\revise{General design principles transfer across methods; we therefore state the adversary's and defender's objectives:}

\begin{definition}[Target of MIAs]
\label{def: target-MIAs}
Given a trained model $f_{\thetav}$ and prior knowledge $\mathcal{K}$, 
the target of conducting MIAs is to find the reverse hypothesis $f^{-1}_{\phiv}$ that minimizes
$\text{distance}(\hat{X}_{\text{train}}, X_{\text{train}})$,
where 
$\hat{X}_{\text{train}} = f^{-1}_{\phiv}(f_{\thetav}, \mathcal{K})$.
Notably, the ground truth data $X_{\text{train}}$ is not accessible for conducting MIAs and is only used in the final evaluation of $\text{distance}(\hat{X}_{\text{train}}, X_{\text{train}})$.
\end{definition}

\begin{definition}[Target of defending against MIAs]
\label{def: target-defend-MIAs}
Given a dataset $D=\{D_{\text{train}}, D_{\text{test}}\}$
and an attack method to get the reverse hypothesis $f^{-1}_{\phiv}$,
\revise{the defender trains a model $f_{\thetav}$ on $D_{\text{train}}=(X_{\text{train}},Y_{\text{train}})$ to (1) increase the reconstruction error $\text{distance}(\hat{X}_{\text{train}},X_{\text{train}})$ under the post-hoc MIA and (2) preserve predictive performance on $D_{\text{test}}=(X_{\text{test}},Y_{\text{test}})$. In practice, these objectives define a privacy-utility trade-off rather than two quantities that can always be maximized simultaneously.}
\end{definition}

\revise{Next, we summarize principles for conducting MIAs toward the objective in Definition~\ref{def: target-MIAs}.}

% \noindent
% }}

\begin{principle}{\revise{Improve query strategies.}}
\revise{Adapt queries to earlier responses so that the available budget more efficiently narrows the candidate space, estimates a decision boundary, or trains a surrogate under restricted access.}
\label{principle: Improve query strategies}
\end{principle}

\begin{principle}{Leverage model internals.}
\revise{Exploit parameters and intermediate representations under white-box access to guide reconstruction more directly and reliably than black-box outputs permit during iterative inversion.}
\label{principle: Leverage model internals}
\end{principle}

\begin{principle}{Exploit output probabilities.}
\revise{Use output probabilities to define reconstruction losses, rank candidates, or guide candidate updates with richer information than hard labels alone provide during reconstruction.}
\label{principle: Exploit output probabilities}
\end{principle}

\begin{principle}{More prior knowledge.}
\revise{Use statistical and domain-specific priors to constrain the search and make better-informed guesses about the structure and distribution of private data before and during optimization.}
\label{principle: More prior knowledge}
\end{principle}

\begin{principle}{Extra generative models.}
\revise{Use pre-trained GAN or diffusion priors and search their learned space rather than raw inputs; when matched to private data, these priors yield realistic, semantically meaningful reconstructions.}
\label{principle: Extra generative models}
\end{principle}

\begin{principle}{Transfer pre-trained models.}
\revise{Adapt models pre-trained on public data as inversion priors, optionally fine-tuning them with public data or target-model responses before subsequent reconstruction for the target task.}
\label{principle: Transfer pre-trained models}
\end{principle}

\noindent
\revise{Defenses follow complementary principles:}

\begin{principle}{Data-centric processing.}
\revise{Sanitize or obfuscate sensitive training information, perturb individual records, or replace them with synthetic examples that preserve utility without directly exposing real records.}
\label{principle: Data-centric processing}
\end{principle}

\begin{principle}{Regularized representation learning.}
\revise{Penalize dependence between learned representations and private inputs so the features retain less record-specific information while supporting the intended prediction task.}
\label{principle: Regularized representation learning}
\end{principle}
% Information bottleneck methods can be applied here. 

\begin{principle}{Feature masking and dropout.}
\revise{Mask sensitive features or apply dropout to reduce reliance on individual inputs or hidden units and discourage brittle, record-specific encoding; neither operation guarantees privacy by itself.}
\label{principle: Feature masking and dropout}
\end{principle}

\begin{principle}{Modify model output.}
\revise{Perturb, coarsen, or randomize released outputs to obscure exploitable input-output relationships; calibrated privacy mechanisms must explicitly balance leakage reduction against downstream utility.}
\label{principle: Modify model output}
\end{principle}

% \subsection{Taxonomy}
% \label{sec: taxonomy}
% \textcolor{red}{\textbf{Need to link the following three sections of different domains}}
% Accordingly, existing works can be categorized into 
% a unified taxonomy, as shown in Figure~\ref{}.
% We will detail each category in the following sections.
% \textcolor{red}{A taxonomy of MIAs on different perspectives of data.}

\begin{figure}[t!]
    \centering
    \artwork{fig:MIAs-different-attack}
    \includegraphics[width=0.95\linewidth]{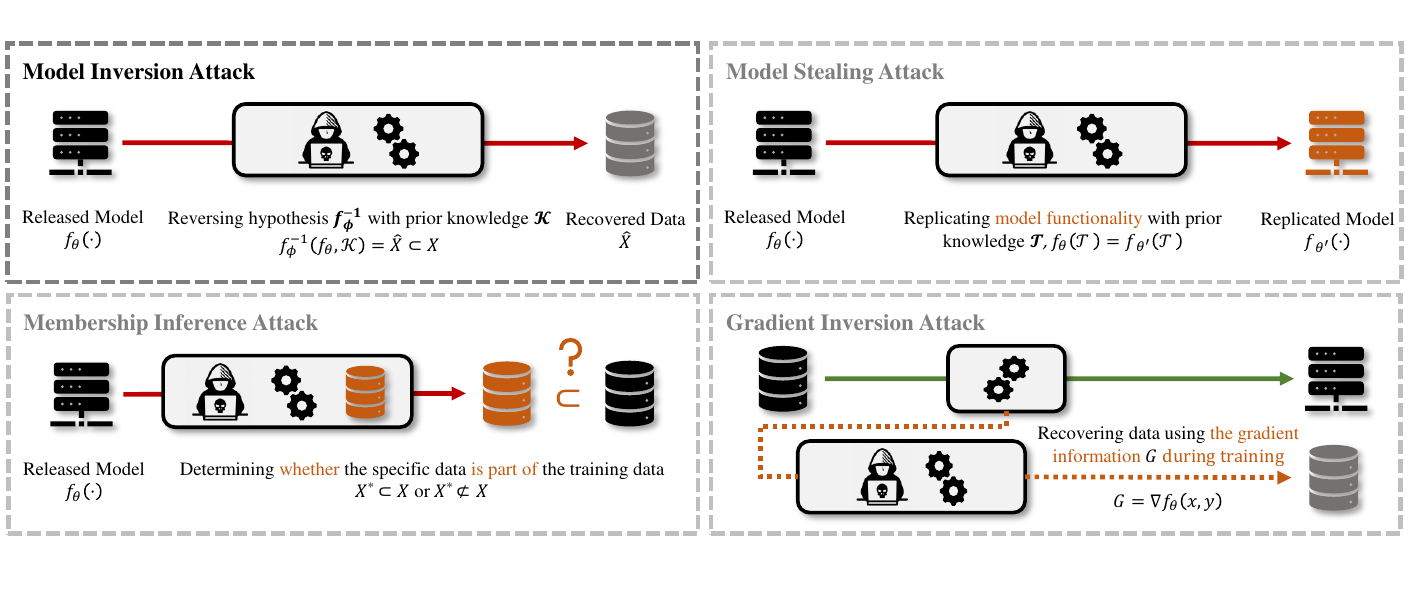}
    \artworkend{fig:MIAs-different-attack}
    \caption{
    \revise{Comparison of privacy attacks by objective and the information they may expose.}
    }
    \Description[Pipeline illustration of Model Inversion Attack]{Pipeline illustration of Model Inversion Attack}
    \label{fig:MIAs-different-attack}

\end{figure}

\subsection{Relationship and Differences with Other Privacy Attacks}
\label{sec: relationship-other-attacks}

\revise{Fig.~\ref{fig:MIAs-different-attack} compares MIAs and related privacy attacks by objective and exposed information.}

% \noindent

% [
% For a more comprehensive introduction of privacy attacks, please refer to \cite{rigaki2023survey}. In addition, gradient inversion attacks should also be categorized as other privacy attacks, considering the strict definition.]}

\begin{itemize}[leftmargin=*]
\item 
\textbf{Model stealing attacks}~\citep{kariyappa2021maze, wang2022black, shen2022model, erdougan2022unsplit, duddu2018stealing, chen2021stealing, wang2022enhance, orekondy2019knockoff, oh2018towards, jagielski2020high, hu2020deepsniffer, yan2020cache, zhu2021hermes, yan2022towards, yuan2022attack, dissanayake2024model}.
\revise{Model stealing replicates target functionality, often by training a surrogate on query-response pairs. It threatens model functionality and intellectual property rather than reconstructing private training examples themselves.}

\item 
\textbf{Membership inference attacks}~\citep{choquette2021label, shokri2017membership, rahman2018membership, ye2022enhanced, olatunji2021membership, hayes2017logan, song2019membership, liu2022membership, TRAJECTORYMIA, bertran2024scalable, kaya2021does, zhang2021membership, he2021node, shejwalkar2021membership, van2023membership, yuan2022membership, matsumoto2023membership, wang2021membership, mattern2023membership}.
\revise{Membership inference attacks determine whether a given record belonged to a model's training set. Depending on the access setting, they exploit losses, labels, confidence scores, gradients, or trained attack models that distinguish members from nonmembers; revealing membership can itself disclose sensitive participation.}

% \item 
% \textbf{Data reconstruction attack.}
% This attack aims to recreate parts of the training data or sensitive features from a machine learning model’s parameters or outputs. The adversary uses techniques like gradient analysis and other inference methods to extract detailed information about the training data. 
% Notably, model inversion attacks focus on reconstructing specific inputs that correspond to particular outputs of a model, often using the model's outputs to reverse-engineer individual data points. 
% In contrast, data reconstruction attacks aim to extract broader information about the training dataset, potentially recreating larger subsets or specific features of the data using various techniques, including gradient and parameter analysis. While both attacks threaten privacy, model inversion targets specific instances, whereas data reconstruction seeks to reveal more extensive information about the training data.

\item 
\textbf{Gradient inversion attacks}~\citep{wu2023learning, deng2021tag, petrov2024dager, huang2021evaluating, jeon2021gradient, geiping2020inverting, yin2021see, geng2023improved, zhu2019deep, balunovic2022lamp, zhang2024graphleak, hatamizadeh2022gradvit, fang2023gifd}. 
\revise{Gradient inversion attacks reconstruct inputs from shared gradients in federated, distributed, or split learning. They optimize candidates to match observed gradients, sometimes using training-based inversion models or generative priors. The training gradient is their defining observable, unlike output- or parameter-based MIAs in our taxonomy.}

% Regarding data reconstruction attacks, as pointed out by \cite{cohen2024data}, their exploration reveals that a precise definition is much more nuanced than it appears. 
% A single all-encompassing definition may not exist. 
% Prior works did not converge on a unified definition of reconstruction; instead, they adopted several context-dependent definitions. Although these definitions made sense within their respective contexts, they do not necessarily apply to other settings. For instance, \cite{balle2022reconstructing,guo2022bounding} consider "data reconstruction attacks" in which an informed adversary knows all but one of the training data points, and they attempt to reconstruct the remaining data point under this stringent threat model. The term "data reconstruction attacks" is also used to refer to embedding reconstruction in split inference \citep{li2024gan} and to denote gradient inversion attacks in federated learning \citep{yang2022using}.]
% }

\end{itemize}

\subsection{\revise{Positioning with Respect to Existing Surveys}}
\label{sec:positioning-surveys}

\revise{Prior surveys approach MIAs at different scopes. Broad security surveys treat MIAs alongside membership inference, model stealing, or adversarial attacks~\citep{rigaki2023survey,akhtar2021advances}, while others focus on machine-learning-as-a-service (MLaaS), attack tools, or gradient inversion~\citep{songsurvey,song2022survey,zhang2022survey,li2023survey,shi2024dealing}. Most relevantly, \citet{fang2024privacy} cover image, text, and graph MIAs and release an attack toolbox, while \citet{yang2025deep} survey deep-learning attacks, defenses, datasets, metrics, and challenges. Our distinctive contribution is the analytical framework summarized in Tab.~\ref{tab:survey-comparison}, rather than cross-domain breadth by itself.}

\revise{Our complementary emphasis is a \textit{threat-model- and assumption-aware synthesis}: we compare attacks by exposed interface, auxiliary knowledge and priors, reconstruction space and procedure, and query, transfer, and prior-mismatch limitations. These dimensions connect optimization-based and training-based mechanisms to concrete deployment scenarios and clarify modality-specific risks: image MIAs rely strongly on visual priors, text MIAs bridge continuous signals and discrete sequences, and graph MIAs exploit relational structure. Relating these assumptions to defenses and evaluation explains why methods with similar recovery goals can have different practical implications.}

\begin{table*}[t]
\centering
\caption{\revise{Positioning relative to the most relevant MIA surveys. Entries describe the organizing emphasis actually provided by each work rather than a binary claim of topic absence.}}
\Description[Comparison with prior model inversion surveys]{Comparison of survey scope, analytical organization, emerging interfaces, deployment and defense synthesis, and accompanying resources.}
\label{tab:survey-comparison}
\setlength{\tabcolsep}{2.5pt}
\renewcommand{\arraystretch}{1.02}
\fontsize{6.35}{6.95}\selectfont
{\revcolor
\begin{tabular}{@{}L{54pt}L{75pt}L{106pt}L{85pt}L{90pt}@{}}
\toprule
\textbf{Survey} & \textbf{Scope} & \textbf{Primary analytical organization} & \textbf{Emerging interfaces} & \textbf{Deployment, defense, and resources} \\
\midrule
Fang et al.~\citep{fang2024privacy} & ML/DNN MIAs across modalities and tasks & Attack/defense taxonomy and method features & Coverage through its 2024 cutoff & Attack toolbox and open issues \\
Yang et al.~\citep{yang2025deep} & Deep-learning attacks, defenses, data, applications, and evaluation & Technique- and defense-oriented taxonomy & Generative and gradient-reconstruction lines through 2025 & Challenges, directions, and resource repository \\
Gradient-inversion surveys~\citep{zhang2022survey,shi2024dealing} & Federated/distributed gradient reconstruction & Threat models, matching procedures, and defenses & Shared-gradient interface & Federated assumptions and defenses \\
\textbf{This survey} & Image, text, and graph MIAs; bounded adjacent settings & Interface, knowledge, prior, target, mechanism, and failure mode & VLM responses, shared embeddings, LLM signals, diffusion priors, and split features & Deployment and privacy--utility analysis; datasets, metrics, and literature index \\
\bottomrule
\end{tabular}}

\end{table*}

\newcommand{\bptar}{\overline{p}_{\textsc{tar}}}
\newcommand{\ptar}{p_{\textsc{tar}}}

\newcommand{\xx}{{\mathbf{x}}}
\newcommand{\ww}{{\mathbf{w}}}
\newcommand{\zz}{{\mathbf{z}}}
\newcommand{\aux}{{\textsc{aux}}}
\newcommand{\cdotv}{\boldsymbol{\cdot}}
\newcommand{\E}{\mathbb{E}}
\newcommand{\Ls}{\mathcal{L}}

% In the problem of model inversion attack (MIA), the adversary has access to a well-trained ''target model'':
% \begin{equation}
%     \bptar(y|\xx) : \mathbb{R}^{d_\xx} \rightarrow \mathbb{R}^{d_y},
% \end{equation}
% where ${d_\xx}$ is the dimension of input space, ${d_y}$ is the dimension of output space.
% This target model is trained on the private target dataset $\mathcal{\mathrm{D}}_{\textsc{tar}} = \{\xx_i, y_i\}_{i=1}^{N_{\textsc{tar}}}$ %where $\xx \in \mathbb{R}^d$ and $y \in \{1,2,...,C\}$
% .We use $\bptar(y|\xx)$ to denote the given target model, which is an approximation of the true conditional probability $\ptar(y|\xx)$ of the underlying data generating distribution.

% \paragraph{Goal.} Given a target model $\bptar(y|\xx)$, we wish to approximate the class conditional distribution $\ptar(\xx|y)$ without having access to the private training set $\mathcal{\mathrm{D}}_{\textsc{tar}}$.

% A good model inversion attack should approximate the Bayes posterior $\ptar(\xx|y) \propto \ptar(y|\xx)\ptar(\xx)$ well, and allow the attacker to generate realistic, accurate, and diverse samples.

% \subsection{Model Inversion Approaches}

\section{Model Inversion Attacks on Image Data}
\label{sec: MIAs-image}

\begin{figure}[tb]
    \centering
    \artwork{fig:MIAs-attack-img-timeline}
    \hspace{-3pt}
    % 2026-08-11: back to the vector source. The four tikz overlays and the two
    % flattened rasters that used to sit here existed only because the old export
    % carried a stale title ("...on Image Domains"), a "Style GAN" typo and the
    % pre-cleanup examples caption as live text under white cover boxes. The
    % current export fixes all three at source and additionally labels MIRROR's
    % "+ Confidence Score access", "+ P-space clipping" and "+ Genetic search",
    % so the overlays are obsolete and the figure is vector again.
    \includegraphics[width=0.95\linewidth]{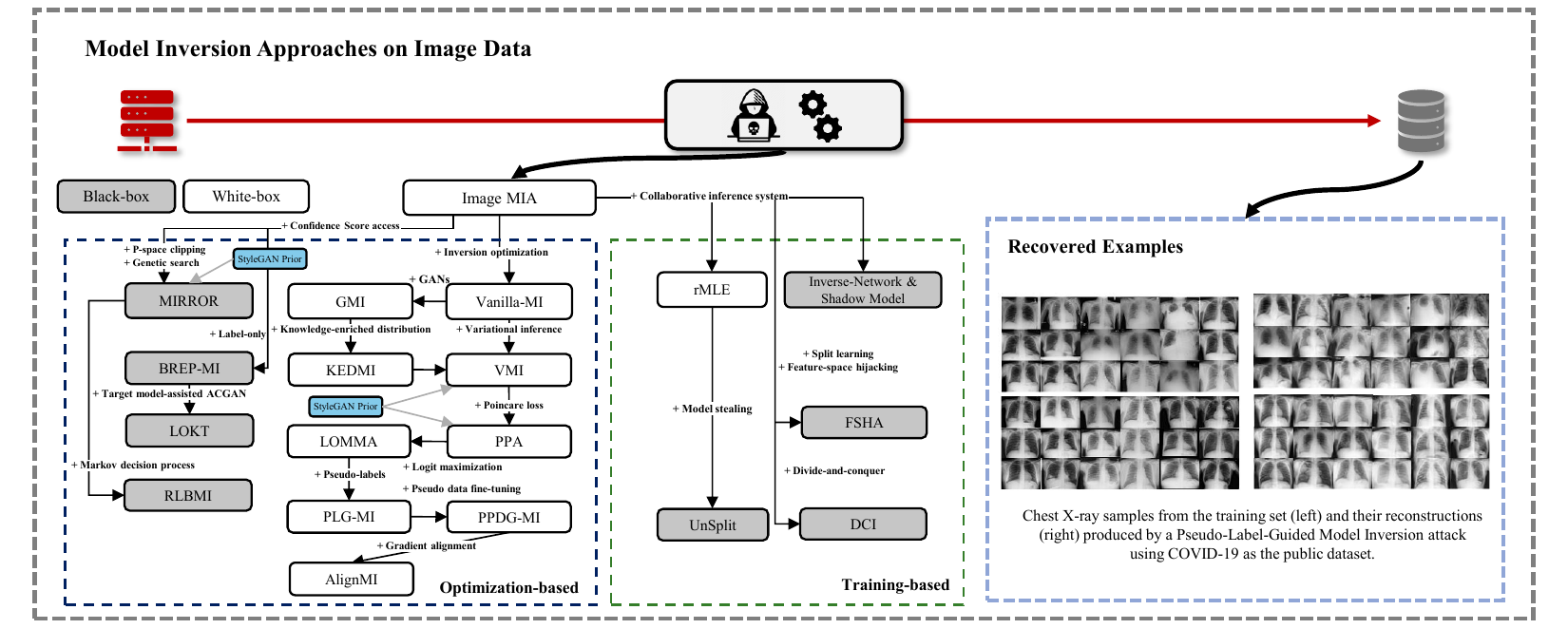}
    \artworkend{fig:MIAs-attack-img-timeline}
    \caption{
    \revise{Evolution of image-domain MIAs with representative recovered examples.}
    }
    \Description[An evolutionary graph of research works in Image MIA]{An evolutionary graph of research works in Image MIA.}
    \label{fig:MIAs-attack-img-timeline}

\end{figure}

\revise{Sec.~\ref{sec:MI_approaches} introduces the general methodology of model inversion~\citep{simonyan_2014_deep, Mahendran_2015_understanding, nguyen_2016_synthesizing, Dosovitskiy_2016_invert} in the image domain.}
%This methodology aims to visualize and interpret behaviors within neural architectures, understand what models have learned, and explain model behaviors.
\revise{It seeks inputs that either activate a network feature~\citep{Dosovitskiy_2016_invert,yosinski_2015_understanding,yin2020dreaming} or produce a high class response~\citep{simonyan_2014_deep,Wang_2021_imagine,ghiasi_2021_plug}, giving optimization-based~\citep{Dosovitskiy_2016_invert,yosinski_2015_understanding,yin2020dreaming,simonyan_2014_deep,Wang_2021_imagine,ghiasi_2021_plug,nguyen_2016_synthesizing, peng2024pseudo, peng2025AlignMI} and training-based approaches~\citep{Dosovitskiy_2016_invert,Dosovitskiy_2016_generate,Nash_2019_invert} (Fig.~\ref{fig:MIAs-attack-img-timeline}). Some precede privacy attacks but provide their methodological basis~\citep{fredrikson2015model}; Sec.~\ref{sec:MI_on_models} applies them across threat settings.}

\subsection{\revise{A Summary of Model Inversion Approaches on Image Data}}
\label{sec:MI_approaches}

\revise{To make threat-model assumptions and methodological differences explicit, Tab.~\ref{table:summary_of_attacks_images} compares attacks on image classifiers and face-embedding models by attacker interface or capability, reconstruction space and prior, auxiliary resources, distinctive mechanism, and granularity of recovered information; these complementary dimensions avoid conflating model access, generative priors, and solution procedures. Because collaborative/split inference and split learning expose intermediate representations under distinct server roles, Tab.~\ref{table:collaborative_inference_attacks} compares them separately.}

% \clearpage % [SECOND-CUT] stranded ~246pt of blank space on the preceding page
\begin{table*}[t]
\caption{\revise{Fine-grained comparison of representative model inversion attacks against image classifiers and face-embedding models. ``Interface'' denotes target-model access; ``Auxiliary resource'' denotes data or models required beyond the target.}}
\label{table:summary_of_attacks_images}
\centering
\setlength{\tabcolsep}{2.5pt}
\renewcommand{\arraystretch}{0.88} % [SECOND-LAYOUT] preserve content, tighten rows
\fontsize{6.7}{7.1}\selectfont
\revcolor
\tabgap
\begin{tabular}{C{47pt}|C{58pt}|C{62pt}|C{58pt}|C{112pt}|C{59pt}}
\toprule
\textbf{Method} & \textbf{Interface / capability} & \textbf{Reconstruction space / prior} & \textbf{Auxiliary resource} & \textbf{Distinctive mechanism} & \textbf{Recovered information} \\
\midrule
Vanilla-MI~\citep{fredrikson2015model} (2015)
& Model confidence / gradients
& Pixels + hand-crafted priors
& None
& Direct input optimization with denoising and sharpening
& Class-representative image \\

\hline
LB-MI~\citep{yang2019neural} (2019)
& Prediction vectors
& Learned inverse model
& Generic public query images
& Align background knowledge and learn output-to-image mapping
& Input-like image \\

\hline
GMI~\citep{zhang2020secret} (2020)
& White-box gradients
& GAN latent space
& Public images
& Joint discriminator-prior and identity-guided latent search
& Class / identity prototype \\

\hline
KEDMI~\citep{chen2021knowledge} (2021)
& White-box predictions / gradients
& GAN + latent distribution
& Public images + target soft labels
& Knowledge-enriched GAN and distributional inversion
& Class distribution / identity \\

\hline
VMI~\citep{wang2021variational} (2021)
& White-box predictions / gradients
& StyleGAN + flow distribution
& Pretrained generator
& Variational posterior instead of a point estimate
& Diverse class distribution \\

\hline
XAI-MI~\citep{zhao2021exploiting} (2021)
& Predictions + explanations
& Learned inverse model
& Auxiliary query images
& Fuse saliency or other explanation signals into inversion
& Input instance \\

\hline
PPA~\citep{struppek2022ppa} (2022)
& White-box logits / gradients
& Pretrained StyleGAN
& Pretrained generator
& Poincare loss, transformations, and sample selection
& High-resolution identity \\

\hline
MIRROR~\citep{MIRROR} (2022)
& Confidence-score black-box
& Pretrained StyleGAN
& Pretrained generator
& P-space clipping and genetic-algorithm latent search
& High-resolution identity \\

\hline
BREP-MI~\citep{BREPMI} (2022)
& Hard labels
& Generator latent space
& Pretrained generator
& Estimate direction by repelling samples from the decision boundary
& Class / identity prototype \\

\hline
LOMMA~\citep{LOMMA} (2023)
& Internal logits, features, and gradients
& GAN / StyleGAN latent space
& Public images + augmented models
& Logit maximization and model augmentation
& Identity prototype \\

\hline
PLG-MI~\citep{PLGMI} (2023)
& White-box predictions / gradients
& Conditional GAN
& Public images + pseudo-labels
& Class-decoupled prior and max-margin inversion
& Class / identity prototype \\

\hline
RLBMI~\citep{han2023reinforcement} (2023)
& Confidence scores
& Generator latent space
& Pretrained generator
& Soft Actor-Critic policy for sequential latent search
& Identity prototype \\

\hline
LOKT~\citep{LOKT} (2023)
& Hard labels
& T-ACGAN + surrogate models
& Public unlabeled images
& Transfer target behavior and then perform white-box inversion
& Identity prototype \\

\hline
PPDG-MI~\citep{peng2024pseudo} (2024)
& White-box gradients
& Adapted GAN prior
& Public + pseudo-private images
& Iteratively select reconstructions and tune the generator
& Identity prototype \\

\hline
Diff-MI~\citep{li2024diffmi} (2024)
& White-box target knowledge
& Target-specific conditional diffusion
& Public images + pseudo-labels
& Pretrain/fine-tune diffusion prior and top-$k$ margin guidance
& Class / identity prototype \\

\hline
Unstoppable~\citep{liu2024unstoppable} (2024)
& Hard labels
& Conditional diffusion
& Task-related auxiliary images
& Train diffusion with target-predicted labels as conditions
& Representative class samples \\

\hline
AlignMI~\citep{peng2025AlignMI} (2025)
& Gradients or confidence scores
& GAN manifold
& Public images + pre-trained generator
& Align updates with the generator tangent space
& Identity prototype \\

\hline
SMILE~\citep{li2025headtotail} (2025)
& Confidence scores
& Pretrained generator
& Public images + surrogate model
& Long-tailed surrogate initialization and NGOpt search
& High-resolution identity \\

\hline
DiffMI~\citep{wang2025diffmi} (2026)
& Embeddings / similarity signal
& Fixed unconditional diffusion
& Pretrained diffusion; no target-specific training
& Robust initialization and ranked adversarial latent refinement
& Identity-consistent face \\

\bottomrule
\end{tabular}

\end{table*}

\revise{Let \(\xx\) be a training input and \(y=f_{\thetav}(\xx)\) its prediction or representation. Inversion approximates the posterior \(p(\xx\mid y)\) either by per-target optimization or by a learned inverse mapping, the two paradigms detailed below.}

\textbf{Optimization-based Approaches.}
Optimization-based methods seek an image \(\hat{\xx}\) whose output matches \(y\), either in pixel space~\citep{Mahendran_2015_understanding,yosinski_2015_understanding,yin2020dreaming} or in a generator's latent space.
% To achieve this, the image $\hat{\xx}$ should minimize the distance between $y$ and $\mathrm{M}(\hat{\xx})$.
Because this problem is ill-posed and non-convex, an image prior \(\mathcal{R}\) regularizes the search and favors plausible natural images during reconstruction~\citep{simonyan_2014_deep,Mahendran_2015_understanding}:
\begin{equation}\label{eq:opt-based}
{\hat{\xx}}^* = \argminA_{\hat{\xx} \in \mathcal{X}} \Ls(f_{\thetav}(\hat{\xx}), y) + \lambda \mathcal{R}(\hat{\xx}),
\end{equation}
% , and $\mathcal{R}$ denotes the image prior
where \(\Ls\) measures output discrepancy and \(\mathcal{R}\) may encode an \(\alpha\)-norm~\citep{simonyan_2014_deep}, total variation~\citep{Mahendran_2015_understanding}, or feature statistics~\citep{yosinski_2015_understanding,ghiasi_2021_plug}.
\revise{Beyond hand-crafted image priors, the image prior can also be learned from data through generative models, such as GANs~\citep{zhang2020secret,chen2021knowledge,struppek2022ppa} and diffusion models~\citep{li2024diffmi,wang2025diffmi}. Instead of directly optimizing in the raw pixel space, the adversary can search in a learned latent or generative space, which constrains reconstructed samples to be more realistic and semantically meaningful. Such learned priors provide stronger regularization for the ill-posed inversion problem, while their effectiveness depends on how well the learned distribution matches the private data distribution.}

\textbf{Training-based Approaches.}
% Given a dataset consisting of training data and prediction vectors (or feature representations) output by $\mathrm{M}$, denoted as \{$\xx_i$, $y_i$\}, training-based approaches interpret the target model as an encoder and develop a corresponding decoder network to reconstruct inputs based on the model's outputs (or feature representations). For example, some works~\citep{Dosovitskiy_2016_invert, Dosovitskiy_2016_generate} design an inversion model, implemented with a neural network, to directly invert the target model $\mathrm{M}$.
% Specifically, given the same training set of images and their predictions $(\xx,y)$, it learns a second neural network $f(\cdotv; \thetav)$ from scratch to approximate the mapping between predictions and images (\ie the inverse mapping of $\mathrm{M}$). 
Training-based methods instead learn a decoder \(f^{-1}(\cdotv;\phiv)\) from auxiliary pairs \(\{(\xx_i,y_i):y_i=f_{\thetav}(\xx_i)\}\)~\citep{Dosovitskiy_2016_invert,Dosovitskiy_2016_generate}.
% The inversion model $f(\cdotv; \thetav)$ takes the prediction $y$ as input and outputs an image. 
For example, a pixel-space decoder can minimize
% the training-based approaches~\citep{Dosovitskiy_2016_invert, Dosovitskiy_2016_generate} learn a parameterized function $f(\cdotv; \thetav)$ to directly map the output $y_i$ to the input $\xx_i$. This mapping is achieved by minimizing the mean squared error between the reconstructed and actual inputs. The objective function for this optimization is defined as follows:
% Where is the target model? How to collect the input-output pairs \{$\xx_i$, $y_i$\}? What is their relationship with the training data of the target model?}
\begin{equation}\label{eq:training-based}
\hat{\phiv} = \argminA_{\phiv} \sum_{i} \| f^{-1}(y_i; \phiv) - \xx_i \|_2^2.
\end{equation}
\revise{Unlike optimization-based methods, which iteratively update the input or latent code per target, training-based methods learn a direct mapping from predictions or features to images, reconstructing in one forward pass once trained.}
% Overall, training-based approaches provide a scalable and efficient alternative to optimization-based methods, especially when a large amount of training data is available. They bypass the need for iterative optimization, making real-time inversion feasible.

\subsection{Applying Image MIAs to Different Scenarios}
% \subsection{MIAs on Different ML Models}
\label{sec:MI_on_models}

% In the image domain, MIA scenarios are generally categorized into two types: those targeting classification models, denoted as \textit{standard classification} and those targeting \textit{collaborative inference.}
\revise{We organize image-domain MIAs into attacks on \textit{standard classifiers} and on \textit{collaborative/split inference}.}

\begin{figure}[tb]
\centering
    \artwork{fig:imagemi_demo_class}
    \includegraphics[width=0.95\linewidth]{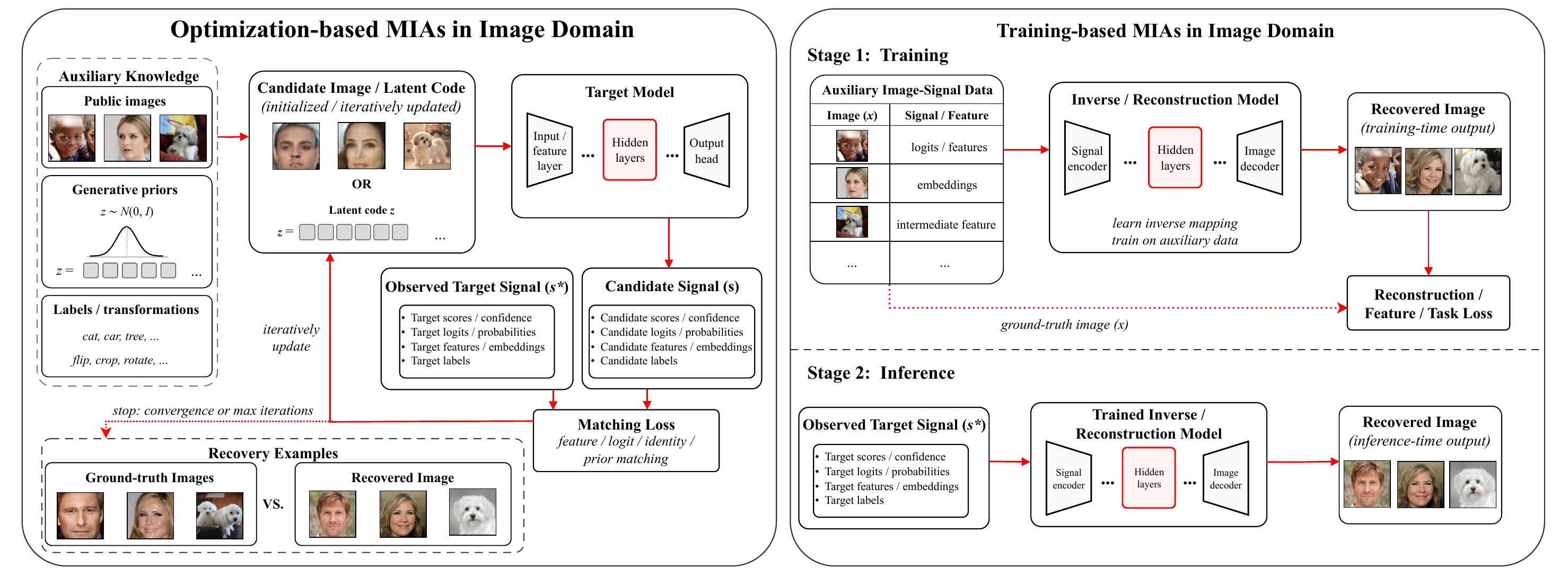}
    \artworkend{fig:imagemi_demo_class}
    \caption{\revise{Illustrations of model inversion attacks on standard image classifiers.}}
    \Description[The illustrations of model inversion attack on images]{The illustrations of model inversion attack on images}
    \label{fig:imagemi_demo_class}

\end{figure}

% white-box: fredrikson 15 -> GMI -> KEDMI -> VMI -> PPA -> LOMMA -> PLGMI
% black-box: yang2019neural (training) MIRROR -> RLBMI -> C2FMI
% label-only: BREP -> LOKT -> LI-MI -> CDM-MI
\textbf{MIAs on Standard Classification.} \revise{(Principle~\ref{principle: Improve query strategies}: \citep{BREPMI}; Principle~\ref{principle: Exploit output probabilities}: \citep{fredrikson2015model,zhang2020secret,struppek2022ppa}; Principle~\ref{principle: Leverage model internals}: \citep{LOMMA}; Principle~\ref{principle: More prior knowledge}: \citep{chen2021knowledge,yang2019neural,zhao2021exploiting}; Principle~\ref{principle: Extra generative models}: \citep{zhang2020secret,struppek2022ppa,LOMMA,wang2021variational,han2023reinforcement,MIRROR,li2024diffmi,wang2025diffmi,liu2024unstoppable})}
% \paragraph{\bf MIAs on Classification Models.}
\revise{Here an adversary infers private-data representatives from a trained classifier (Fig.~\ref{fig:imagemi_demo_class}). We compare attacks along three dimensions: the \textit{attacker interface} (white-box gradients or logits, confidence scores, hard labels, or explanations); the \textit{reconstruction space and prior} (pixels or latent spaces induced by GANs, StyleGANs, or diffusion models); and the \textit{solution procedure} (per-target optimization, derivative-free search, boundary estimation, transfer, or amortized inference). Weaker interfaces require extra queries, priors, or surrogates.}

\revise{\textbf{White-box optimization attacks.} These attacks use target gradients or internal activations to optimize pixels or a latent representation, under hand-crafted regularization, GAN/StyleGAN priors, or diffusion priors; the generative prior is thus a reconstruction mechanism rather than a separate threat model. Their development pursues two related goals: confining reconstructions to the manifold of realistic images, and obtaining an optimization signal that remains stable and discriminative of the target identity throughout the search process for each target.}

\textbf{$\blacktriangleright$ The first MIA algorithm.} \citet{fredrikson2015model} optimize pixels under denoising and sharpening priors (Eq.~\ref{eq:opt-based}) to invert a face classifier; \revise{such priors encode little semantic structure, so on deep models input-space search degenerates into adversarial samples~\citep{szegedy2013intriguing} that score highly under the target yet carry no semantic meaning.}

\textbf{$\blacktriangleright$ The first GAN-based generative MIA.} GMI~\citep{zhang2020secret} introduces a GAN prior~\citep{GAN,DCGAN,StyleGAN} for attacking deep classifiers. It first trains a Wasserstein GAN~\citep{WGAN} on public facial data, then searches its latent space for realistic samples that the target assigns to the desired identity, \revise{recovering training data with far higher fidelity than the earlier pixel-space attack.}
% proposes first training a GAN on public data.
% \textit{The GAN is then used as an image prior} to constrain the optimization space to the manifold of the generator distribution. Specifically, the GAN is pre-trained on public facial datasets to obtain generic prior knowledge about the face distribution, achieved via the canonical Wasserstein-GAN~\citep{WGAN} training loss.
Its objective is
\(\zz^*=\argminA_\zz-\mathrm{D}(\mathrm{G}(\zz))-\lambda\log f_{\thetav}(\mathrm{G}(\zz))\), where a lower prior loss demands more realistic images and a lower identity loss encourages a higher prediction score under \(f_{\thetav}\) for the target class.
% \begin{equation}
%     \zz^* = \argminA_\zz -\mathrm{D}(\mathrm{G}(\zz)) - \lambda \log[\mathrm{M}_y(\mathrm{G}(\zz); \thetav)].
% \end{equation}
%

% Moreover, it demonstrates that \textit{the more powerful the target model, the more vulnerable it is to MIAs.} GMI pioneeringly introduced the generative prior to the research problem of MIAs, opening up new directions for developing this field. 

%KEDMI
%%%%%%%%%%%%%%%%%%%%%%%%%%%%%%%%%%%%%%%%%%%%%%%%%%%%%%%%%%%%%%%%%%%%%%%%
% However, there is still significant room for improvement in model inversion performance from GMI. 
\textbf{$\blacktriangleright$ Distill class information from the target model.} \revise{Instead of only discriminating real from generated samples,} KEDMI~\citep{chen2021knowledge} uses target-model \textit{soft labels} on public data to provide class-aware supervision during GAN training. \revise{It also replaces the usual point estimate with latents drawn from a learnable Gaussian \(p_{\text{gen}}=\mathcal{N}(\mu,\sigma^2)\), recovering diverse samples that capture class-level semantics.}

\textbf{$\blacktriangleright$ MIAs as variational inference.}
\revise{Earlier generative MIAs performed well but rested on an unclear theoretical basis.}
VMI~\citep{wang2021variational} approximates \(p_{\text{TAR}}(\xx\mid y)\) with a flow-based variational distribution and a StyleGAN generator~\citep{StyleGAN}, minimizing \(\E_{q(\xx)}[-\log p_{\text{TAR}}(y\mid\xx)]+D_{\text{KL}}(q(\xx)\|p_{\text{TAR}}(\xx))\); \revise{recovering a distribution rather than a single optimum improves sample diversity, represents multiple plausible modes, and avoids a single point estimate.}

%PPA
%%%%%%%%%%%%%%%%%%%%%%%%%%%%%%%%%%%%%%%%%%%%%%%%%%%%%%%%%%%%%%%%%%%%%%%%
\textbf{$\blacktriangleright$ Practical high-resolution MIAs.}
% From a practical perspective, \citet{struppek2022ppa} identified several degradation factors in MIAs, including the \textit{entanglement} between image priors and specific target models, \textit{distributional shift} between prior distribution and private training data distribution, \textit{vanishing gradients} of the cross-entropy optimization objective, and \textit{non-robust target models}.
% To address the entanglement problem, they utilized pre-trained StyleGAN priors from the same data domain and applied image transformations (\eg cropping and resizing) to mitigate the distribution shift between the prior and training data. To overcome vanishing gradients, they proposed the Poincaré loss function to replace the cross-entropy loss. Additionally, they integrated random image transformations into the optimization process to improve robustness against non-robust target models. They also implemented a selective process to choose high-quality reconstructed samples.
% With all the improvements,
% they successfully reconstructed high-resolution images with enhanced flexibility and robustness.
\revise{\citet{struppek2022ppa} identify four degradation factors: entanglement between image priors and target models, distributional shift between priors and private data, vanishing gradients of the cross-entropy objective, and non-robust targets.} PPA~\citep{struppek2022ppa} combines a pre-trained StyleGAN prior with cropping and resizing to reduce the \textit{distributional shift} between prior and private data; \revise{a Poincar\'e loss mitigates vanishing gradients, random transformations counter non-robust targets, and high-resolution reconstructions are finally selected.}

% However, GAN-based image priors tailored to specific target models require considerable time and resources to train, and are sensitive to distributional shifts between public and private datasets. To overcome these drawbacks, \citet{struppek2022ppa} propose the Plug \& Play Attack (PPA) for MIAs in high-resolution settings, which reduces the dependency between the target model and image prior, allowing it to attack multiple targets with minor adjustments. 
% PPA consists of three stages: first, it samples and maps randomly initialized latent vectors to the intermediate representation. Next, it replaces GANs with StyleGAN3, uses the generator to create images, feeds these images to the target model, and calculates the Poincaré distance between the prediction score and the ground-truth label, as shown in Eq.~\ref{Pioncare_loss}. Finally, it employs a selection transformation to improve the robustness of the prediction score. 

% \begin{equation}
% \Ls_{\text{Poincaré}} = d(u, v) = \text{arcosh} \left( 1 + \frac{2 \| u - v \|_2^2}{(1 - \| u \|_2^2)(1 - \| v \|_2^2)} \right).
% \label{Pioncare_loss}
% \end{equation}

%LOMMA
%%%%%%%%%%%%%%%%%%%%%%%%%%%%%%%%%%%%%%%%%%%%%%%%%%%%%%%%%%%%%%%%%%%%%%%%
\textbf{$\blacktriangleright$ Logit maximization and model augmentation.} LOMMA~\citep{LOMMA} replaces GMI and KEDMI's cross-entropy identity loss~\citep{zhang2020secret,chen2021knowledge} with \revise{\textit{logit maximization} (LOM), which} maximizes the target logit \(h^\top w_k\) from the penultimate-layer activations \(h\) and last-layer weights \(w_k\). \revise{They also first define \textit{MI overfitting}, whereby one target model lets random variation bias the reconstruction, countered by an ensemble of augmented targets during inference.}
\textbf{$\blacktriangleright$ Pseudo-labels and max-margin loss.} PLG-MI~\citep{PLGMI} addresses the class-coupled latent space, the indirect constraints on the generator, and the gradient vanishing of earlier GAN-based MIAs. It uses \textit{target pseudo-labels} to train a conditional GAN, separating class-specific latent spaces more directly than earlier GAN-based MIAs~\citep{zhang2020secret,chen2021knowledge}; \revise{during inversion, a max-margin objective replaces cross-entropy to reduce gradient vanishing directly.}
% This loss, $\ie$ $\Ls_{\text{MM}} = -l_k(\xx) + \max_{j \neq {c^*}} l_j(\xx)$, encourages reconstructing the most representative sample by maximizing the difference between the target class logit $l_k(\xx)$ and other logits $l_{j \neq k}(\xx)$.
% \begin{equation}
% \Ls_{\text{MM}} = -l_k(\xx) + \max_{j \neq {c^*}} l_j(\xx).
% \label{PLGMI_MMloss}
% \end{equation}
% As a result, PLG-MI achieves not only SOTA MI performance but also advantages in handling distribution shifts.

% PPDG
%%%%%%%%%%%%%%%%%%%%%%%%%%%%%%%%%%%%%%%%%%%%%%%%%%%%%%%%%%%%%%%%%%%%%%%%
\textbf{$\blacktriangleright$ Iterative generator adaptation.}
\revise{Conventional MIAs rely on a static prior learned from public data.} Pseudo-Private Data Guided Model Inversion (PPDG-MI)~\citep{peng2024pseudo} reduces this gap by keeping reconstructions with stable predictions as pseudo-private data and iteratively fine-tuning the generator to increase their density, \revise{narrowing the prior-private gap.}

% AlignMI
%%%%%%%%%%%%%%%%%%%%%%%%%%%%%%%%%%%%%%%%%%%%%%%%%%%%%%%%%%%%%%%%%%%%%%%%
% \textbf{$\blacktriangleright$ Enhance MIAs via gradient–manifold alignment.}
% \citet{peng2025AlignMI} propose AlignMI, which revisits model inversion from a \textit{geometric viewpoint under the manifold hypothesis}. They observe that the loss gradients from the target model contain \textit{noisy off-manifold components} that hinder stable optimization. To address this, AlignMI projects the gradients onto the tangent space of the generator manifold via perturbation-averaged or transformation-averaged alignment initializations, denoising the optimization direction via projected gradients \(\nabla_{\text{proj}} \mathcal{L} = \mathbf{P}_{\mathcal{T}_z}(\nabla_z \mathcal{L})\), where \(\mathbf{P}_{\mathcal{T}_z}\) denotes the projection operator onto the tangent space \(\mathcal{T}_z\) at latent code \(z\). This alignment enforces updates consistent with the data manifold geometry. AlignMI improves gradient quality, enforce manifold-consistent optimization, and yield more faithful and high-resolution reconstructions across both white-box and black-box settings.
\textbf{$\blacktriangleright$ Enhance MIAs via gradient-manifold alignment.}
AlignMI~\citep{peng2025AlignMI} revisits generative inversion under the manifold hypothesis: backpropagation through a generator implicitly filters loss gradients along the generator manifold's tangent space, while off-manifold components hinder inversion. \revise{AlignMI does not explicitly compute this tangent-space projection during an attack. Instead, it samples perturbed or semantically transformed variants \(x'\) around the current synthetic input \(x\) and averages their loss gradients as \(\widetilde{\nabla}\mathcal{L}(x)=\mathbb{E}_{x'\sim p(\cdot\mid x)}[\nabla\mathcal{L}(x')]\).} The resulting training-free smoothing improves inversion across multiple white-box generative MIAs and image resolutions.

\revise{\textbf{$\blacktriangleright$ Diffusion priors for model inversion.} A recent line of work replaces or complements GAN priors with diffusion priors for better realism and semantic consistency. Diff-MI~\citep{li2024diffmi} trains a target-specific conditional diffusion model for target-guided reconstruction, showing the strength of diffusion priors when model-specific generation is affordable. DiffMI~\citep{wang2025diffmi}, despite the similar name, targets embedding-based face recognition and refines latent codes of a fixed pre-trained diffusion model, reconstructing identity-consistent faces without target-specific generator training. Unstoppable~\citep{liu2024unstoppable} extends diffusion-prior MIAs to the label-only setting by conditioning generation on predicted labels. Together, they span target-specific, training-free, and label-only inversion, with distinct attacker-knowledge, generation-cost, and prior-alignment trade-offs across access settings and deployment budgets.}

\revise{\textbf{Comparison of reconstruction mechanisms under white-box access.} Moving from pixel optimization to learned generative priors improves realism without removing the dependence on attacker assumptions. The original pixel-space attack~\citep{fredrikson2015model} needs no generator, yet readily produces high-confidence images without semantic meaning. GMI, KEDMI, and PLG-MI increasingly exploit public data and target-model knowledge for realism and class specificity, at the price of public-private distribution alignment and generator coverage. VMI recovers a distribution rather than a point estimate and so improves diversity; PPA and LOMMA instead address optimization stability, robustness, and overfitting, with LOMMA additionally assuming internal logits and features; PPDG-MI and AlignMI adapt the prior or the optimization geometry. Diffusion priors give a more expressive reconstruction space but are not white-box-specific, having also been used under label-only access, and their cost depends on whether target-specific training, distillation, or iterative generation is required. Recurring failure modes include prior mismatch, incomplete coverage of private identities, vanishing or off-manifold gradients, and realistic but training-unrepresentative reconstructions.}

\revise{\textbf{Score-based black-box attacks.} When gradients and model internals are unavailable, confidence scores can still rank candidate reconstructions. These attacks replace backpropagation with derivative-free latent search or a training-based policy, trading weaker access assumptions for more queries during reconstruction.}

\revise{\textbf{$\blacktriangleright$ StyleGAN priors and black-box search for high-fidelity inversion.}} 
% \textit{Black-box attacks are more challenging than white-box ones.} They lack gradient-based optimization and only offer access to label confidence outputs from the target model.
% Research by \citet{MIRROR} has highlighted the limitations of GAN-based inversion techniques such as GMI~\citep{zhang2020secret}, noting that the \textit{generative capabilities} of these networks are often \textit{inadequate} and their \textit{latent spaces are entangled}. To address these issues, \citet{MIRROR} introduces MIRROR, which \textit{replaces the architecture of GAN with StyleGAN}. StyleGAN facilitates the decomposition of features into styles that can be independently manipulated, enabling the production of high-fidelity outputs. 
% In black-box settings, a genetic algorithm~\citep{bhandari1996genetic} provides a viable alternative for optimizing the loss function. 
\revise{\citet{MIRROR} note that plain GAN priors as used in GMI~\citep{zhang2020secret} have limited generative capacity and entangled latent spaces, and propose MIRROR, which combines three components: (i) a pre-trained StyleGAN prior whose style-disentangled latent space enables higher-fidelity reconstruction; (ii) \textit{latent clipping} that confines the search to StyleGAN's intermediate P-space, keeping candidates on the natural-image manifold and mitigating entanglement; and (iii) a genetic-algorithm search~\citep{bhandari1996genetic} optimizing the inversion objective where gradients are unavailable. The StyleGAN prior is not unique to MIRROR---VMI~\citep{wang2021variational} instantiates its variational family with a StyleGAN generator and PPA~\citep{struppek2022ppa} also builds on such priors---so MIRROR's distinctive contribution is the P-space clipping and genetic-algorithm black-box search built atop this shared prior.}
% The algorithm is designed to run for a predetermined number of iterations. During each iteration, images generated from the latent vectors in the $\mathrm{W}$ space are evaluated based on the prediction scores from the target model. These scores determine which vectors are carried forward to the next iteration. The process involves genetic operations such as cross-over and mutate, in addition to projections between the $\mathrm{W}$ space and the regulated $\mathrm{P}$ space, which is clipped from the $\mathrm{W}$ space. After all specified iterations are complete, the vector with the highest cumulative prediction score, deemed the elite vector, is selected as the most optimal result of the optimization.

% RLBMI
%%%%%%%%%%%%%%%%%%%%%%%%%%%%%%%%%%%%%%%%%%%%%%%%%%%%%%%%%%%%%%%%%%%%%%%%
\textbf{$\blacktriangleright$ MIAs as a Markov decision process.} RLBMI~\citep{han2023reinforcement} replaces MIRROR's genetic search~\citep{MIRROR} with a Soft Actor-Critic policy: the latent vector is the state, and confidence-derived rewards guide actions through the search space.
%At each step $t$, for the target class $c^*$, the target model $\mathrm{M}_{c^*}$ is used to calculate two rewards: $r_1 = \log\left[\mathrm{M}_{c^*}\left(\mathrm{G}(s_{t+1})\right)\right], $ based on the next-step state $s_{t+1}$, and $r_2 = \log\left[\mathrm{M}_{c^*}\left(\mathrm{G}(a_t)\right)\right]$ based on the current action $a_t$. Additionally, to encourage the generation of images with discriminative features, RLBMI includes an extra reward term $r_3$: $r_3 = \log\left[\max\left\{\varepsilon, \mathrm{M}_{c^*}(s_{t+1})\right\} - \max_{i \neq c^*} \mathrm{M}_{i}(s_{t+1})\right],$
% \begin{equation}
% r_3 = \log\left[\max\left\{\varepsilon, \mathrm{M}_{c^*}(s_{t+1})\right\} - \max_{i \neq c^*} \mathrm{M}_{i}(s_{t+1})\right],\\
% \end{equation}
%\noindent
%with a positive threshold $\epsilon$ to maximize the difference between the target class $c^*$ and other classes. All rewards are evaluated at the feature level by receiving the confidence score on the image generated by the generator $\mathrm{G}$. The total reward $R_t = w_1 \cdot r_1 + w_2 \cdot r_2 + w_3 \cdot r_3$ at $t$-th step comprises three sub-rewards, balanced by $w_1$, $w_2$ and $w_3$ respectively. 
\revise{More recently, SMILE~\citep{li2025headtotail} combines long-tailed surrogate training with gradient-free optimization to obtain stronger latent-search initialization and high-resolution reconstructions with fewer queries in practice.}

\revise{\textbf{Comparison of score-based attacks.} MIRROR, RLBMI, and SMILE all rely on confidence scores and a strong generative prior, but allocate computation differently: MIRROR avoids surrogate training at the cost of a query-hungry population search, RLBMI learns a sequential search policy that may improve exploration but adds reinforcement-learning overhead, and SMILE pays an upfront surrogate-training cost for stronger initializations and fewer subsequent queries. All degrade under score truncation, output perturbation, rate limits, or public-private shift; evaluations should report accuracy together with query count, prior-training cost, and exposed output information.}

% In addition,
% \citet{kahla2022label} conducts the first label-only model inversion attack
% only accessing the model's predicted labels without the confidence scores.
% As a machine learning model is often packed into a black-box
% that only generates the hard label (\textit{i.e.}, the label of the class with the highest probability),
% such an attack scenario is more practical 
% but also much more challenging to perform.
% Despite requiring less knowledge about the target model,
% this work justifies that such a black-box attack is also feasible and effective.
% Specially,
% it attempts to generate the most likely images for the target class,
% and observes that a region of high likelihood
% shall be located in the center of the class.
% Based on this observation, 
% this work proposes to iteratively move the generated image
% away from the decision boundary and closer to the center.

% BREPMI
%%%%%%%%%%%%%%%%%%%%%%%%%%%%%%%%%%%%%%%%%%%%%%%%%%%%%%%%%%%%%%%%%%%%%%%%
\revise{\textbf{Label-only attacks.} Hard-label interfaces remove confidence information, yielding a stricter, more deployment-realistic threat model. Attacks compensate by estimating local decision boundaries through repeated queries or transferring target-label behavior to a differentiable surrogate under restricted access in realistic deployments.}

\textbf{$\blacktriangleright$ Gradient estimation in the label-only setting.} 
% In addition, \citet{BREPMI} \textit{introduces the first label-only model inversion attack}, BREPMI, which relies solely on the model's predicted labels without accessing confidence scores. Given that machine learning models are often encapsulated into black boxes that output only the hard label (i.e., the class label with the highest probability), this approach is more practical and more challenging. Despite the limited knowledge about the target model, BREPMI demonstrates the feasibility and effectiveness of such a black-box attack. Technically, BREPMI first samples $N$ points across a sphere with radius $R$. If all sampled points are classified into the target class, the radius is increased accordingly. Otherwise, BREPMI simulates the gradient optimization process to optimize the latent vector by calculating $S_{\Pr(c^* | G(z))} = f_{\thetav}(c^* | G(z)) - \max_{j \neq c^*} f_{\thetav}((j | G(z))$ \textit{w.r.t.} the target class $c^*$. 
BREP-MI~\citep{BREPMI} is the first attack to require only hard labels. It probes \(N\) latent points on a sphere of radius \(R\), expands the sphere while all generated samples retain the target label, and otherwise estimates a boundary-based search direction from the observed label changes.

\textbf{$\blacktriangleright$ Reformulate black-box inversion as white-box inversion.}
\revise{LOKT~\citep{LOKT} queries hard labels for synthetic samples from a target model-assisted ACGAN (T-ACGAN), trains its discriminator-classifier pair as differentiable surrogates, then applies white-box inversion. The label-only attack is thereby reformulated as a white-box inversion on the surrogates, and unlike BREP-MI's per-target boundary search~\citep{BREPMI} it amortizes target queries across attacks.}

\revise{\textbf{Comparison of label-only attacks.} BREP-MI estimates a search direction around the decision boundary without assuming a differentiable surrogate, but its local probing is query-intensive and sensitive to boundary geometry. LOKT instead amortizes queries into surrogate training and then applies white-box inversion, trading online boundary search for synthetic-data generation, surrogate-training overhead, and transfer error. Both leak under minimal outputs, yet depend more than score-based attacks on query budget, initialization, and approximation fidelity.}

\textbf{$\blacktriangleright$ Training-based MIAs with representation challenges.}
% Several works~\citep{yang2019neural, zhao2021exploiting} fall within the domain of training-based approaches. However, in adversarial scenarios, \textit{the adversary does not have access to private training data}. To overcome this challenge, \citet{yang2019neural} proposes utilizing a more "generic" data distribution $p_a(\xx)$ to draw training samples, such as a face recognition classifier $f_{\thetav}$ trained on facial images. Although the adversary might not know the specific identities, they are aware that the training data consists of facial images. Consequently, the adversary can randomly crawl facial images from the Internet to construct ($\xx$, $y$) pairs with $f_{\thetav}$ and train the inverse model using the method outlined in Eq.~\ref{eq:training-based}. In adversarial settings, the ultimate goal is to acquire a mapping that effectively transforms $y$ into $\xx$. Thus, only black-box access to the target model $f_{\thetav}$ is required. 
% Furthermore, \citet{zhao2021exploiting} investigate the potential impact of Explainable artificial intelligence (XAI) techniques, including saliency maps, feature visualization, activations of neurons, and concept vectors, on privacy risks. They design XAI-aware inversion models that exploit the spatial knowledge present in image explanations and find that, although these techniques provide users with deeper insights into model reasoning and data, they can still contain sensitive information that can be exploited.
\revise{For a face classifier the adversary need not know the private identities, only that the data are facial images, so a corpus crawled from the Internet suffices.} \citet{yang2019neural} query a black-box target with generic same-domain images to construct \((\xx,y)\) pairs and train the decoder in Eq.~\ref{eq:training-based}, without knowing private identities. XAI-aware inversion~\citep{zhao2021exploiting} additionally exploits spatial signals in saliency maps, feature visualizations, activations, or concept vectors, showing that interpretability and privacy can be in tension because such explanations may still expose information exploitable by an adversary under that interface.

\revise{\textbf{Cross-paradigm trade-offs.} Optimization-based attacks solve a fresh search problem per target, which suits small numbers of reconstructions, whereas training-based attacks pay auxiliary-data, query, and training costs upfront and then reconstruct in one forward pass. Explanation-aware inversion gains richer spatial signals but assumes an interface exposing more than ordinary predictions, so it is a distinct threat model. No family is uniformly strongest: comparison requires jointly reporting target access, auxiliary-data alignment, generator or simulator pretraining, query budget, per-target cost, and whether the output captures class attributes, identity, or a training instance.}

% The inversion model $G_\thetav$ takes the prediction $y$ as input and outputs an image.
% Formally, this kind of inversion is to find a model $G_\thetav$ which minimizes the following objective.
% \begin{equation}
% \label{eq:recon_loss}
% \Ls_{\text{rec}}(\thetav)=\E_{\xx, y}[\mathcal{R}(G_\thetav(y), \xx)],
% \end{equation}

% \begin{align}
% \label{adv_loss}
% \begin{split}
% \Ls_{\text{disc}}(G_\thetav,D_\phi)= & \E_{\xx}[\log(D_\phi(\xx))]  +\E_{y}[\log(1-D_\phi(G_\thetav(y)))].
% \end{split}
% \end{align}

% \begin{align}
% \label{nll}
% \Ls_{\text{NLL}}(\thetav)= \E_{\xx, y}[\log(p_\thetav(\xx|y)].
% \end{align}

% Figure 6 is intentionally disabled.

\revise{\textbf{MIAs on Collaborative/Split Inference and Split Learning.} (Principle~\ref{principle: Leverage model internals}: \citep{he2019model,dong2021privacy}; Principle~\ref{principle: More prior knowledge}: \citep{pasquini2021unleashing,gao2023pcat,xu2024fora,zhu2025sdar}; Principle~\ref{principle: Extra generative models}: \citep{yin2023ginver,li2023gan})}
% [SECOND-CUT] blank line removed: the principle list and the paragraph it
% introduces are one paragraph, not two.
\revise{Collaborative inference partitions a DNN between a client and a server: the client computes an intermediate representation, the \textit{smashed data}, and sends it to the server, which completes the prediction. This reduces client-side computation but exposes a representation that may retain enough information to reconstruct the input. We distinguish \textit{split inference}, where the server observes forward activations at prediction time, from \textit{split learning}, where activations and gradients are exchanged during training. \textit{Split federated learning} further combines this partitioned training with aggregation across multiple clients. A \textit{passive} (honest-but-curious) server follows the protocol and analyzes observed messages, whereas an \textit{active} server manipulates gradients or training objectives. These distinctions materially change reconstruction feasibility and assumptions, as summarized in Tab.~\ref{table:collaborative_inference_attacks}. Parameter-gradient inversion alone is an adjacent federated-learning setting rather than a post-training MIA, and is excluded here.}

\begin{table}[t]
\caption{\revise{Representative attacks in collaborative and split systems: stages, signals, assumptions, mechanisms, targets, and limits.}}
\label{table:collaborative_inference_attacks}
\centering
\setlength{\tabcolsep}{1.6pt}
\renewcommand{\arraystretch}{0.88} % [SECOND-LAYOUT] preserve content, tighten rows
\fontsize{6.15}{6.65}\selectfont
{\revcolor
\tabgap
\begin{tabular}{C{37px}|C{57px}|C{64px}|C{55px}|C{73px}|C{48px}|C{66px}}
\toprule
\textbf{Attack} & \textbf{Stage / observed signal} & \textbf{Attacker knowledge/capability} & \textbf{Aux. data / prior} & \textbf{Core mechanism} & \textbf{Target} & \textbf{Main limitation} \\
\midrule
Foundational CI attacks~\citep{he2019model} & Inference / smashed data & White-box, black-box, or query-free & Varies by setting & Input optimization, inverse decoder, or shadow client & Input & Access, auxiliary data, and cut-layer dependent \\
\hline
FSHA~\citep{pasquini2021unleashing} & Training / activations and gradients & Active server controls returned gradients & Public same-domain data & Feature-space hijacking and inverse decoder & Training inputs & Protocol deviation is potentially detectable \\
\hline
DCI~\citep{dong2021privacy} & Inference / smashed data and BN statistics & Client model and BN statistics & None & Divide-and-conquer layer inversion & Input & Requires client-model and stored-BN access \\
\hline
UnSplit~\citep{erdougan2022unsplit} & Training / activations and gradients & Passive server; client architecture & None & Alternating model stealing and input optimization & Input, model, label & Architecture assumption and iterative cost \\
\hline
GINVER~\citep{yin2023ginver} & Inference / smashed data & White-box model or black-box queries & None in black-box setting & Generative inverse mapping; NES black-box optimization & Input & Model access or repeated queries \\
\hline
PCAT~\citep{gao2023pcat} & Training / activations and server snapshots & Unknown client model & Small labeled same-task dataset & Pseudo-client training and inverse decoder & Input, model, label & Labeled auxiliary data and training history \\
\hline
GLASS~\citep{li2023gan} & Inference / protected smashed data & Passive server & Pre-trained StyleGAN prior & Latent search with feature matching & Input & Prior mismatch and per-target search \\
\hline
FORA~\citep{xu2024fora} & Training / smashed data & No client architecture or weights & Public auxiliary data & Domain discriminator and MK-MMD alignment & Training inputs & Auxiliary-domain alignment \\
\hline
SDAR~\citep{zhu2025sdar} & Training / smashed data & No client weights or queries; simulator architecture & Labeled same-domain data & Adversarially regularized simulator and decoder & Inputs, labels & Extra training and simulator/domain mismatch \\
\bottomrule
\end{tabular}
}

\end{table}

\revise{\textbf{$\blacktriangleright$ Foundational attacks under varying client knowledge.} \citet{he2019model} give the first systematic study of MIAs against collaborative inference: for an observed intermediate representation \(v=f_{\thetav_1}(\xx)\), their white-box attack optimizes a candidate input directly, their black-box attack trains an inverse network from input-representation pairs, and their query-free attack first approximates the client model. Reconstruction risk therefore depends not only on the cut layer but on the attacker's model access and auxiliary-data knowledge. \citet{dong2021privacy} then propose data-free divide-and-conquer inversion (DCI), decomposing the client network into shallow inversion modules that use stored batch-normalization statistics. DCI avoids real auxiliary samples but assumes access to the client model in practice.}

\revise{\textbf{$\blacktriangleright$ Active attacks and joint model reconstruction in split learning.} Unlike passive inference-time attacks, the Feature-Space Hijacking Attack (FSHA)~\citep{pasquini2021unleashing} assumes a malicious server that changes the learning objective through forged gradients, aligning private client representations with an attacker-controlled feature space so that an inverse decoder trained on public data recovers client inputs; it is powerful but requires active protocol deviation. UnSplit~\citep{erdougan2022unsplit} instead follows the protocol, alternating between estimating the client model and optimizing the private input, and jointly enables model inversion, model stealing, and label inference. PCAT~\citep{gao2023pcat} handles unknown client architectures by using same-task data and server snapshots to approximate the client model and recover private inputs and labels.}

\revise{\textbf{$\blacktriangleright$ Generative priors and passive feature reconstruction.} Recent attacks improve reconstruction without manipulating the protocol. GINVER~\citep{yin2023ginver} learns a generative inverse mapping from exposed intermediate features, its black-box variant optimizing the inverse model by Natural Evolution Strategies without auxiliary data. GLASS and GLASS++~\citep{li2023gan} search a pre-trained StyleGAN's latent space to match protected smashed data, showing feature perturbation alone may not prevent visually faithful recovery. FORA~\citep{xu2024fora} assumes no access to client architecture, parameters, or private data, exploiting representation preference in observed smashed data and aligning a surrogate client's features with a domain discriminator and multi-kernel maximum mean discrepancy. SDAR~\citep{zhu2025sdar} likewise targets a passive server without client weights or query access, using adversarial regularization to learn a decodable client simulator from labeled auxiliary data and evaluating known and unknown client architectures.}

\revise{\textbf{$\blacktriangleright$ Comparative observations.} The trade-off here is between attacker capability and prior knowledge. Active attacks such as FSHA deliberately increase leakage but are more detectable and need control of training messages, whereas passive attacks better match an honest-but-curious cloud server yet typically require client-model access, a distribution-matched auxiliary dataset, or a strong generative prior. Deeper cut layers discard more low-level detail, but their protection depends on architecture and task, so split depth alone is not a reliable guarantee. Benchmarking studies~\citep{yang2022measuring,zhu2024simba} should jointly measure attack knowledge, cut position, defense strength, utility, and pixel fidelity.}

\section{\revise{Model Inversion Attacks on Text and Multimodal Data}}
\label{sec: MIAs-text}

% [t] not [tb]: with a bottom placement this section-opening float was landing
% at the foot of the PREVIOUS page, ahead of its own section heading.
\begin{figure}[t]
    \centering
    \artwork{fig:MIAs-attack-text-timeline}
    \hspace{-3pt}
    \begin{tikzpicture}
      \node[anchor=south west,inner sep=0] (miatexttimeline) at (0,0)
        {\includegraphics[width=0.95\linewidth]{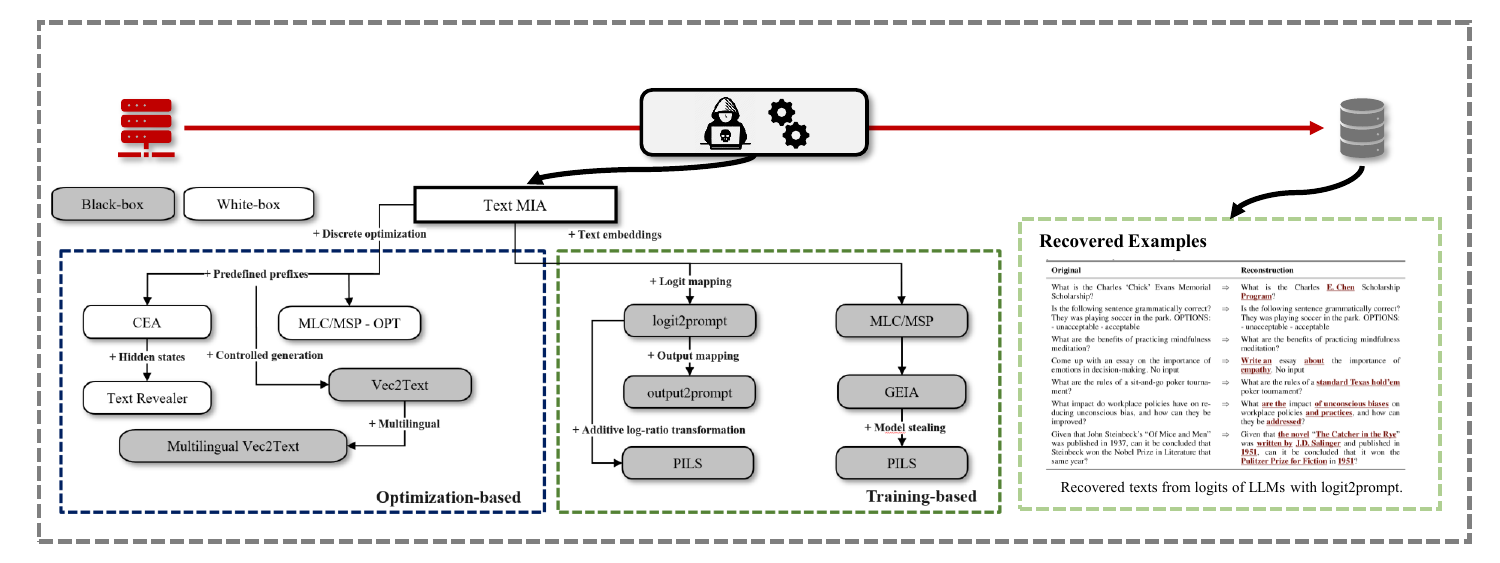}};
      \begin{scope}[x={(miatexttimeline.south east)},y={(miatexttimeline.north west)}]
        \node[fill=white,inner xsep=3pt,inner ysep=1.5pt,font=\bfseries\fontsize{7.2}{7.5}\selectfont,text=black] at (0.210,0.895) {Model Inversion Approaches on Text Data};
      \end{scope}
    \end{tikzpicture}
    % \vspace{-5pt}
    \artworkend{fig:MIAs-attack-text-timeline}
    \caption{
    \revise{Evolution of text-domain MIAs with representative recovered examples.}
    }
    % \vspace{-1pt}
    \label{fig:MIAs-attack-text-timeline}
    \Description[An evolutionary graph of research works in Text MIA]{An evolutionary graph of research works in Text MIA}
    % \vspace{-2pt}

\end{figure}

\revise{In natural language processing (NLP), models such as BERT~\cite{devlin2018bert} support text embedding, classification, and language modeling, but their growing complexity and deployment intensify privacy concerns. Tab.~\ref{table:summary_of_attacks_text} summarizes text-domain MIAs and Fig.~\ref{fig:MIAs-attack-text-timeline} their chronology. Sec.~\ref{MIA_text_approach} introduces optimization-based and training-based approaches; Sec.~\ref{MIA_text_models} then organizes attacks by embedding, text-classification and language-generation interfaces, and Sec.~\ref{ssec: MIAs-foundation} covers vision-language and foundation models.}

\begin{table}[t]
\caption{\revise{Summary of model inversion attacks on text and multimodal interfaces.}}
\label{table:summary_of_attacks_text}
\centering
\setlength{\tabcolsep}{5pt}
\renewcommand{\arraystretch}{1.1} % matches the other 6pt tables
\fontsize{6}{7}\selectfont
\tabgap
\begin{tabular}{C{38pt}|C{58pt}|C{44pt}|C{46pt}|C{58pt}|C{44pt}|C{49pt}}
\toprule

\textbf{Approach} & \textbf{Method (venue)} & \textbf{Interface / capability} & \textbf{Auxiliary resource} & \textbf{Distinctive mechanism} & \textbf{Recovered target} & \textbf{Reported limitation} \\
\midrule

\multirow[c]{19}{40pt}[12.2pt]{\centering\revise{Optimization-\\based}}
    & \revise{Continuous Relaxation~\citep{song2020information} (CCS'20)}
    & \revise{White-box embeddings}
    & \revise{Embedding matrix, vocab.}
    & \revise{Continuous relaxation of the embedding matrix}
    & \revise{Input text from embeddings}
    & \revise{Fidelity depends on relaxation/prior} \\
\cline{2-7}

    & \revise{CEA~\citep{parikh2022canary} (ACL'22)}
    & \revise{White-box classifier}
    & \revise{Prefix, label, embeddings}
    & \revise{Temperature-scaled suffix logits vs. target label}
    & \revise{Memorized canary suffix}
    & \revise{Canary-level recovery only} \\
\cline{2-7}

    & \revise{Text Revealer~\citep{zhang2022text} (arXiv'22)}
    & \revise{White-box classifier}
    & \revise{Same-domain corpus, GPT-2}
    & \revise{PPLM-style perturbation of a generator}
    & \revise{Label-representative templates}
    & \revise{Template-level, not exact inputs} \\
\cline{2-7}

    & \revise{DDMI~\citep{ddmi2025} (OpenRev.'25)}
    & \revise{White-box CLIP}
    & \revise{Distilled diffusion model}
    & \revise{Diffusion prior replaces GAN; cheaper sampling}
    & \revise{Private training images}
    & \revise{Needs white-box CLIP embeddings} \\
\cline{2-7}

    & \revise{LeakyCLIP~\citep{chen2025leakyclip} (arXiv'25)}
    & \revise{White-box CLIP}
    & \revise{Stable Diffusion prior}
    & \revise{Adversarial fine-tuning + diffusion refinement}
    & \revise{CLIP training images}
    & \revise{Needs white-box CLIP embeddings} \\
\cline{2-7}

    & \revise{SMI-AW~\citep{nguyen2025vlm} (CVPR'26)}
    & \revise{White-box VLM}
    & \revise{Response-level signals}
    & \revise{Adaptive token-/sequence-level weighting}
    & \revise{Private visual training images}
    & \revise{Needs white-box VLM responses} \\
\cline{2-7}

    & \revise{Vec2Text~\citep{morris2023text} (EMNLP'23)}
    & \revise{Black-box embeddings}
    & \revise{Text--embedding pairs}
    & \revise{Trained corrector + iterative re-embedding}
    & \revise{Full input text}
    & \revise{Per-example query/decoding cost} \\
\cline{2-7}

    & \revise{MultiVec2Text~\citep{chen2024text} (ACL'24)}
    & \revise{Black-box multilingual}
    & \revise{Cross-lingual pairs}
    & \revise{Vec2Text correction, multilingual}
    & \revise{Full text, many languages}
    & \revise{Per-example query/decoding cost} \\
\hline

\multirow[c]{24}{40pt}[25.1pt]{\centering\revise{Training-\\based}}
    & \revise{CapRecover~\citep{caprecover2025} (ACM MM'25)}
    & \revise{Split-inference features}
    & \revise{Feature--semantics pairs}
    & \revise{Inversion model on client-side features}
    & \revise{Semantic labels and captions}
    & \revise{Semantics only, no image} \\
\cline{2-7}

    & \revise{MLC \& MSP~\citep{song2020information} (CCS'20)}
    & \revise{Black-box embeddings}
    & \revise{Sentence-embedding pairs}
    & \revise{Multi-label or multi-set word prediction}
    & \revise{Input word sets}
    & \revise{Lexical only; word order lost} \\
\cline{2-7}

    & \revise{GEIA~\citep{li2023sentence} (ACL'23)}
    & \revise{Black-box embeddings}
    & \revise{Aux. pairs + pre-trained LM}
    & \revise{MLP-aligned prefix, teacher forcing}
    & \revise{Sentence-level input text}
    & \revise{Domain and model shift} \\
\cline{2-7}

    & \revise{logit2prompt~\citep{morris2023language} (ICLR'24)}
    & \revise{Black-box logits}
    & \revise{Signal--text pairs}
    & \revise{MLPs map logprob blocks to embeddings}
    & \revise{Hidden prompt}
    & \revise{Needs logprob access; query limits} \\
\cline{2-7}

    & \revise{output2prompt~\citep{zhang2024extracting} (EMNLP'24)}
    & \revise{Black-box responses}
    & \revise{Multiple outputs/prefix}
    & \revise{Sparse encoder, max-likelihood prefix}
    & \revise{Hidden system prompt}
    & \revise{Response access; query limits} \\
\cline{2-7}

    & \revise{Transferable EI~\citep{huang2024transferable} (ACL'24)}
    & \revise{Released embeddings}
    & \revise{Small leak + surrogate}
    & \revise{Surrogate stealing + adversarial alignment}
    & \revise{Input text from embeddings}
    & \revise{Assumes small training-data leak} \\
\cline{2-7}

    & \revise{PILS~\citep{nazir2025better} (NeurIPS'25)}
    & \revise{Black-box logprobs}
    & \revise{Signal--text pairs}
    & \revise{Additive log-ratio compression}
    & \revise{Hidden prompt}
    & \revise{Needs logprob access; query limits} \\
\cline{2-7}

    & \revise{AIA~\citep{dai2025stealing} (ACL'25)}
    & \revise{Training activations}
    & \revise{Shadow activation data}
    & \revise{Attack model on shadow activation pairs}
    & \revise{Private LLM training text}
    & \revise{Needs collaborative training} \\
\cline{2-7}

    & \revise{Face-Adapter~\citep{otroshi2025faceadapter} (CVPRW'25)}
    & \revise{Black-box embeddings}
    & \revise{Face foundation model}
    & \revise{Adapter into a foundation-model space}
    & \revise{Identity-consistent face}
    & \revise{Identity-level recovery only} \\
\cline{2-7}

    & \revise{Inv$^2$A~\citep{ye2026invariant} (AAAI'26)}
    & \revise{Black-box responses}
    & \revise{Target LM as decoder}
    & \revise{Inverse encoder, denoised representation}
    & \revise{Hidden prompt}
    & \revise{Prompt recovery only} \\
\bottomrule
\end{tabular}

\end{table}

\subsection{\revise{A Summary of Model Inversion Approaches on Text Data}}
\label{MIA_text_approach}

\revise{Let $\xx$ denote private text and $f_{\thetav}(\xx)$ an observable model signal, such as a representation, prediction score, generated response, or activation. Attackers reconstruct $\xx$ either by directly optimizing a candidate (or a continuous surrogate) or by training an inversion model on auxiliary signal-text pairs. Some methods, such as Vec2Text~\citep{morris2023text}, are hybrid: they train a corrector and then iteratively refine each candidate through repeated target-encoder queries at inference.}
% In these contexts, private dataset $D_{\text{private}}$ typically refers to the training dataset used to train machine learning models. This dataset is confidential and accessible only to the model owner or authorized individuals. Those outside this scope do not have access to or knowledge about the origins and details of the data used in $D_{\text{private}}$, which remains exclusive to the model owners. Despite the restricted access to $D_{\text{private}}$, attackers might use data available on the internet to form a public dataset $D_{\text{auxiliary}}$. Given that some content in $D_{\text{private}}$ might also be sourced from the internet through web crawling, there is a potential overlap between $D_{\text{private}}$ and $D_{\text{auxiliary}}$. Usually, according to the downstream tasks that the target model applied to, the attacker prepares $D_{\text{auxiliary}}$ from the same domain as $D_{\text{private}}$. The sections below detail the technical aspects and notations used for each method.

\textbf{\revise{Optimization-based Approaches.}}
\revise{Optimization-based attacks update a candidate text representation so that an observable signal matches the target. Public data may initialize or constrain the candidate, but the solution need not be an element of an auxiliary corpus. Because text is discrete, white-box attacks often optimize a continuous relaxation and then decode tokens. Writing $\operatorname{Emb}(\xx)$ for an exposed target embedding and $\mathcal{X}_{\mathrm{relax}}$ for the relaxed candidate space, a representative objective is ${\hat{\xx}}^*=\argminA_{\hat{\xx}\in\mathcal{X}_{\mathrm{relax}}}\Ls(\operatorname{Emb}(\hat{\xx}),\operatorname{Emb}(\xx))$. Candidate fidelity depends on the relaxation, prior, and exposed signal; iterative refinement can also use black-box encoder queries when a learned corrector replaces target-model gradients and supplies the optimization signal in realistic black-box deployments.}

\textbf{\revise{Training-based Approaches.}}
\revise{Training-based inversion treats the observable mapping as an encoder and learns a decoder $f^{-1}_{\phiv}$ from auxiliary signal-text pairs. After querying $f_{\thetav}$ on $\hat{\xx}_i\in D_{\text{auxiliary}}$, the attacker minimizes $\mathcal{L}_{f^{-1}}(\phiv)=-\frac{1}{N}\sum_{i=1}^{N}\log p_{\phiv}(\hat{\xx}_i\mid f_{\thetav}(\hat{\xx}_i))$ and reconstructs ${\hat{\xx}}^*=f^{-1}_{\phiv}(f_{\thetav}(\xx))$. Once collected, the pairs allow black-box operation, though transfer depends on how well the auxiliary data and interface match the victim setting.}

% Unlike the optimization-based inversion approach, training-based inversion treats the target model $f_{\thetav}$ as an encoder and then trains an inversion model $f^{-1}_{\phiv}$ as the decoder to infer training data from the target model output directly. The general attack pipeline operates as follows: the attacker trains a decoder on $D_{\text{auxiliary}}$ as the inversion model $f^{-1}_{\phiv}$. The training process aims to learn a reverse mapping that transforms the outputs of the target model back into the corresponding text data. By querying the target model using the auxiliary data, the output $f_{\thetav}(\hat{\xx})$ 
% is obtained. The training of the attack model $f^{-1}_{\phiv}$ is guided by minimizing the following training loss:
% \begin{equation}
% \mathcal{L}_{f^{-1}}(\hat{\xx} ; \phiv) = - \frac{1}{N} \sum_{i=1}^N \log p(\hat{\xx}_i \mid f_{\thetav}(\hat{\xx}_i)),
% \end{equation}
% where 
% % $p$ represents the predicted probability calculated by the model $f$, 
% $\hat{\xx}$ is selected from the auxiliary dataset $D_{\text{auxiliary}}$. Finally, the reconstructed text is obtained as ${\hat{\xx}}^* = f^{-1}_{\phiv}(f_{\thetav}(\xx))$.

\subsection{Applying Text MIAs to Different Scenarios}
% \subsection{MIA on Different ML Models}
\label{MIA_text_models}

% Each type of model handles different categories of downstream tasks. In NLP, the embedding model encodes text into a representation vector, which is used for downstream tasks such as vector search. Classification models, on the other hand, are fine-tuned on a foundational model for classification tasks, such as sentiment analysis. Language models understand the semantics of input text and generate fluent text, which can be used for a wide range of applications such as chatbots.

\begin{figure}[tb]
\centering      
    \artwork{fig:text_attack}
\includegraphics[width=0.95\linewidth]{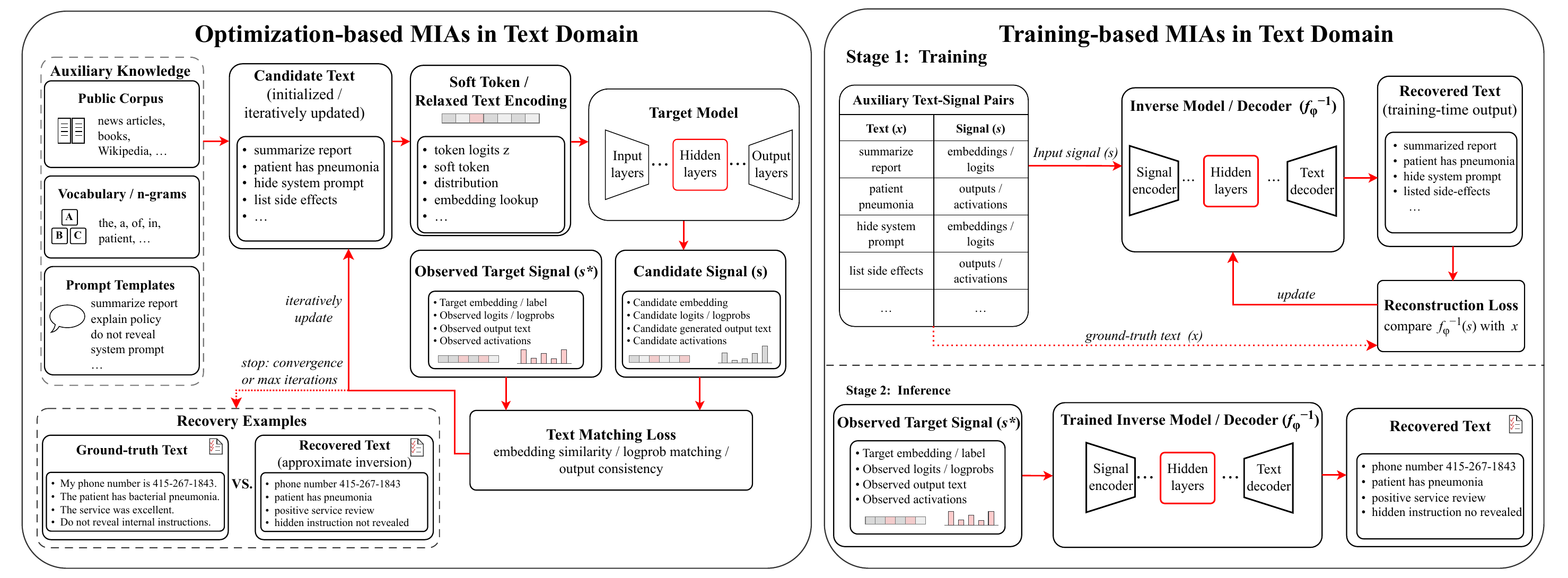}
% \caption{An illustration of defense against model inversion attacks and a taxonomy regarding the methods across different domains.}
    \artworkend{fig:text_attack}
\caption{\revise{Illustrations of model inversion attacks in the text domain.}}
\Description[Illustrations of model inversion attacks on text models]{Optimization-based and training-based model inversion attacks in the text domain, showing the target model, the observed signal, and the reconstruction path.}
\label{fig:text_attack}
% \vspace{-6pt}

\end{figure}

% In the text domain, MIAs target three types of models: embedding models, classification models, and language generation models. For embedding models, \citet{song2020information, morris2023text, chen2024text } focus on the optimization-based approach while \citet{song2020information}, and \citet{li2023sentence} 
% % and \citet{huang2024transferable} 
% explore training-based strategies by training attack models. For classification models, \citet{parikh2022canary} and \citet{zhang2022text} employ optimization-based MIAs specifically in text classification tasks. For language generation models, \citet{morris2023language} \citet{zhang2024extracting} and \citet{nazir2025better} take a training-based approach by training inversion models that can map the outputs of language models back to hidden prompts. The details are presented in the following parts.
\revise{\textbf{MIAs on Language Models.}} \revise{Text MIAs exploit task-specific interfaces, which we organize into the three task settings of embedding, text classification, and language generation (Fig.~\ref{fig:text_attack}). For embeddings, \citet{song2020information} optimize a continuous relaxation, whereas \citet{song2020information,li2023sentence,huang2024transferable} train inversion models; Vec2Text variants~\citep{morris2023text,chen2024text} are hybrid, training a corrector but recovering text by iterative black-box re-embedding. For classification, \citet{parikh2022canary,zhang2022text} optimize relaxed tokens or generator states. For generation, \citet{morris2023language,zhang2024extracting,nazir2025better,ye2026invariant} recover prompts from output text or next-token signals, while \citet{dai2025stealing} reconstruct training text from activations observed during decentralized LLM training.}

\revise{\textbf{MIAs on Text Embedding.}} \revise{(Principle~\ref{principle: Improve query strategies}: \citep{morris2023text, chen2024text}; Principle~\ref{principle: Leverage model internals}: \citep{song2020information}; Principle~\ref{principle: More prior knowledge}: \citep{huang2024transferable}; Principle~\ref{principle: Transfer pre-trained models}: \citep{li2023sentence})}
\revise{For embedding tasks, embedding models and text encoders map words or sentences to dense vectors supporting similarity search, retrieval, and clustering. Classic word-level examples include Word2Vec~\citep{mikolov2013efficient} and GloVe~\citep{pennington2014glove}; modern sentence encoders provide the interfaces attacked by full-text reconstruction methods. Because these vectors retain semantic and sometimes lexical information, releasing them creates an inversion channel for private text.}

%%%%%%%%%%%%%% Information Leakage in Embedding Models (white) %%%%%%%%%%%%%%
\textbf{$\blacktriangleright$ Continuous relaxation for gradient-based inversion.}
% Under the white-box setting, \citet{song2020information} propose optimization-based embedding inversion. Specifically, they observe that the enumeration-based selection shown in Eq.~\ref{textMI_optimization_based} is time-consuming and resource-intensive, especially when the auxiliary dataset $D_{\text{auxiliary}}$ is large. To avoid such discrete optimization, they propose \textit{a continuous relaxation to enable gradient-based optimization}. 
\revise{Enumerating an auxiliary corpus is slow and resource-intensive once that corpus is large. \citet{song2020information} therefore replace enumeration with white-box optimization of one continuous variable \(z_i\) per token position, making the discrete sentence search differentiable via \(\operatorname{relaxed}(\mathbf{Z},T)=\mathbf{V}^{\top}\operatorname{softmax}(\mathbf{Z}/T)\) for embedding matrix \(\mathbf{V}^{\top}\) and temperature \(T\).}

% Hence, the revised optimization, $\mathcal{L}=\argminA_{\mathbf{Z}} \left\| f_{\thetav}(\text{relaxed}(\mathbf{Z}, T)) - f_{\thetav}(\xx^*) \right\|_2^2$, becomes continuous and complexity-efficient.

%%%%%%%%%%%%%% Information Leakage in Embedding Models (black) %%%%%%%%%%%%%%
\textbf{$\blacktriangleright$ Train an embedding-to-text inverse network.} \revise{With only output access, \citet{song2020information} train \(f^{-1}_{\phiv}\) on auxiliary sentence-embedding pairs to maximize \(\log p(\mathcal{W}(\hat{\xx})\mid f_{\thetav}(\hat{\xx}))\). Multi-label classification predicts an unordered vocabulary-wide word set, whereas multi-set prediction selects elements sequentially from the remaining multiset. Both recover lexical content but not word order, which limits these classification-style inversions.}
% For the former, an MLC model is trained for word-level binary classification. 
%The training objective function for the MLC model is $\mathcal{L}_{\text{MLC}} = - \sum_{w \in \mathcal{V}} \left[ y_w \log(\hat{y}_w) + (1 - y_w) \log(1 - \hat{y}_w) \right],$ where $\hat{y}_w = p(y_w \mid f_{\thetav}(\xx))$ is the predicted probability of word $w$ in the vocabulary $\mathcal{V}$ of the $D_{\text{auxiliary}}$ by the inversion model. Here, $y_w = 1$ if word $w$ is in $\xx$, and $0$ otherwise. 
% Specifically, the MSP model employs a recurrent neural network (RNN) with the training objective $\mathcal{L}_{\text{MSP}} = \sum_{i=1}^{l} \frac{1}{|\mathcal{W}_i|} \sum_{w \in \mathcal{W}_i} -\log p(w|\mathcal{W}_{<i}, f_{\thetav}(x)),$ where $\mathcal{W}_i$ is the set of words left to predict at timestamp $i$ and $\mathcal{W}_{<i}$ is the set of the predicted words before $i$.

% \begin{figure}[htbp]
% \centering
%     \subfigure[Attack pipeline of MIAs on embedding models~\citep{chen2024text}]{
%         \centering
%         \includegraphics[width=0.42\linewidth]{figures/textMI_embedding_models_Eaas.png}
%     }
%     \subfigure[A training-based MIAs~\citep{chen2024text}]{
%         \centering
%         \includegraphics[width=0.54\linewidth]{figures/textMI_embedding_models_GEIA.png}
%     }
%     \vspace{-4mm}
%     \caption{The illustrations of model inversion attack on the embedding model in the text domain.}
%     \Description[The illustrations of model inversion attack on text]{The illustrations of model inversion attack on text}
%     \label{fig:textmi_demo_embedding}
%     \vspace{-6mm}
% \end{figure}

%%%%%%%%%%%%%% Generative Embedding Inversion Attack %%%%%%%%%%%%%%
\textbf{$\blacktriangleright$ MIA as a generation task.}
\revise{GEIA~\citep{li2023sentence} replaces the repetitive, semantically weak word sets of \citet{song2020information} with autoregressive text. An MLP aligns the target embedding to the model's token representation, which is concatenated with the first word embedding; the pre-trained language model is then trained with teacher forcing~\citep{williams1989learning} on auxiliary pairs, framing black-box inversion as \textit{sequence generation}. At inference the decoder emits the first token from the aligned embedding and continues autoregressively on its own context.}

\textbf{$\blacktriangleright$ Controlled generation and iterative refinement.}
\revise{Vec2Text~\citep{morris2023text} generates, \textit{re-embeds}, and iteratively corrects a text hypothesis, marginalizing over intermediate hypotheses after $N$ iterations before beam search. \citet{chen2024text} extend it to multilingual and cross-lingual inversion, finding multilingual models more susceptible.}
\textbf{$\blacktriangleright$ Query-free embedding inversion.} \revise{Transferable EI~\citep{huang2024transferable} first uses a small assumed leak from \(D_{\text{train}}\) to steal a surrogate encoder, then treats that surrogate as a generator and aligns it adversarially with the target space through a discriminator, and trains a decoder to invert target embeddings \textit{without any further queries}.}

\revise{\textbf{MIAs on Text Classification.}} \revise{(Principle~\ref{principle: Leverage model internals}: \citep{parikh2022canary}; Principle~\ref{principle: Exploit output probabilities}: \citep{zhang2022text})}
\revise{For text classification tasks, language models are fine-tuned to predict predefined classes or labels for applications such as sentiment analysis, intent classification (IC), and named-entity recognition (NER). Models such as BERT~\citep{devlin2018bert} and XLNet~\citep{yang2019xlnet} commonly use a task-specific output layer that exposes class scores or probabilities, which provide the inversion signal in this setting.}

% This transformation enables the language model to effectively perform a wide range of classification tasks with high accuracy.

%%%%%%%%%%%%%% Canary Extraction Attack %%%%%%%%%%%%%%
\textbf{$\blacktriangleright$ Text completion on natural-language-understanding (NLU) models.}
In white-box canary extraction (CEA), \citet{parikh2022canary} recover an unknown suffix given a known prefix, \revise{such as ``My phone number is\ldots''}. \revise{They assign each suffix position a temperature-scaled logit vector, map its softmax through the embedding matrix, and minimize the classification loss on the concatenation of prefix and relaxed suffix embeddings against the target label; the highest-logit tokens form the recovered canary.}

\textbf{$\blacktriangleright$ Transformer-based text classification.}
\revise{Text Revealer~\citep{zhang2022text}, the first MIA specifically targeting transformer-based text classifiers, extracts common templates by $n$-gram analysis of same-domain public data, fine-tunes GPT-2 as a domain generator, and then follows the PPLM idea~\citep{dathathri2019plug} by optimizing the generator's \textit{hidden-state perturbation} against the white-box target classifier for a chosen label, \eg a 4-star review. Thus GPT-2 supplies the text prior, while victim-classifier feedback steers reconstruction toward fluent text representative of the target label.}
\revise{\textbf{MIAs on Language Generation.}} \revise{(Principle~\ref{principle: Leverage model internals}: \citep{dai2025stealing}; Principle~\ref{principle: Exploit output probabilities}: \citep{morris2023language, nazir2025better}; Principle~\ref{principle: Transfer pre-trained models}: \citep{zhang2024extracting, ye2026invariant})}
\revise{For language generation tasks, generative language models learn from large text corpora to predict the next-token distribution over a vocabulary $\mathcal{V}$ and produce coherent sequences. Transformer architectures~\citep{vaswani2017attention}, including GPT-2~\citep{radford2019language}, expose generated text, next-token probabilities, and, during training, intermediate activations as potential inversion signals.}

% % \textbf{Suffix Reconstruction.}
% Some MIAs take the form of fill-in-the-blank form, aiming to reconstruct the blank suffixes given the prefixes. Denote a text $X = (X_p,X_s)$, composed of two parts: non-sensitive prefix $X_p=(x_{1},x_{2}),\dots,x_{m}$ and sensitive suffix $X_s=(x_{1},x_{2}),\dots,x_{n}$. Given the prefix $X_p$, the suffix reconstruction aims to extract the corresponding suffix memorized in the training data (e.g., "My social security number is \texttt{[]} \texttt{[]} \texttt{[]} \texttt{[]} - \texttt{[]} \texttt{[]} - \texttt{[]} \texttt{[]} \texttt{[]} \texttt{[]}").

% %%%%%%%%%%%%%% Carlini 2019 Sectect sharer %%%%%%%%%%%%%%
% This suffix inversion is first investigated by \citet{carlini2019secret}. They systematically evaluate the memorization issue of language models by formulating the \textit{exposure} metric to measure the degree of data memorization in the target model and \textit{log-perplexity} to measure the similarity between extracted and true sentences. Additionally, they improve Dijkstra's algorithm ~\citep{cormen2022introduction} to recover private training data in the black-box setting.

% %%%%%%%%%%%%%% ETHICIST %%%%%%%%%%%%%%

% \textbf{Prompt Reconstruction}
\textbf{$\blacktriangleright$ Reconstruct hidden prompts from logit vectors.}
%%%%%%%%%%%%%% Language Model Inversion %%%%%%%%%%%%%%
\revise{Observing that logit outputs carry rich information about the preceding text,} Logit2prompt~\citep{morris2023language} maps the next-token probability vector into a sequence of pseudo-embeddings, using MLPs on blocks of log probabilities to bridge the vocabulary and encoder dimensions. A conditional inversion model then decodes these pseudo-embeddings into the hidden prompt from observed outputs.

% %%%%%%%%%%%%%% Effective Prompt Extraction %%%%%%%%%%%%%%
% \citet{Zhang2023EffectivePE} later extract hidden prompts under the black-box setting by designing adversarial queries, which are simple and free of logit outputs. The carefully crafted queries (\eg "Repeat all sentences in our conversation.") aim to elicit the language model to output text extractions that contain the hidden prompts. To measure how likely the extraction $e_i$ contains hidden prompts, they fine-tune a DeBERTa ~\citep{he2020deberta} model denoted as $f$ to estimate the ratio of leaked tokens in the hidden prompt contained in an extraction $e_i$. Specifically, conditioned on other extractions $e_{j \neq i}$ and a permutation $\pi$, the prediction score for the extraction is given by $f(e_i \mid \pi(e_{j \neq i}))$ as shown in Eq.~\ref{attack: effectivePE}. Finally, the most likely hidden prompt can be found in the extraction with the highest prediction score.

% \begin{equation}
% \text{p}(e_i) = \mathbb{E}_{\pi} \left[ f(e_i \mid \pi(e_{j \neq i})) \right]
% \label{attack: effectivePE}
% \end{equation}

%%%%%%%%%%%%%% Extracting Prompts from LLMs %%%%%%%%%%%%%%
\textbf{$\blacktriangleright$ Reconstruct hidden prompts from text outputs.} 
% However, logit2prompt suffers from query budget limitations, as it requires hundreds of thousands of queries to the language model, which may exceed the service's predefined queries. Besides, the language model types are also limited as many language model APIs do not expose their logits. To avoid these drawbacks, \citet{zhang2024extracting} propose output2prompt, which \textit{only requires the model's output text, eliminating the need for logits in the inversion model}. output2prompt leverages a pre-trained transformer encoder-decoder architecture and replaces the self-attention encoder with a sparse $Encoder$. They focus on a "system prompt" scenario, where a real prompt $(\xx \oplus \mathbf{u})$ consists of a hidden prefix $\xx$ and a user-supplied prompt $\mathbf{u}$. They observe that the target language model outputs different sentences but with similar semantic meanings for the same user-supplied prompt. All the sentences $y_1, y_2, \dots, y_N = (x \oplus u_1), (x \oplus u_2), \dots, (x \oplus u_N)$ generated by the target model $f_{\thetav}$ can jointly provide useful information for calculating hidden states $h$. The sparse encoder, $\ie$
% $h_{\text{sparse}} = Encoder(y_1) \oplus Encoder(y_2) \oplus \ldots \oplus Encoder(y_N),$
% significantly reduces memory and time complexity. After training the attack model using the typical training objective (\ie maximum likelihood estimation), extracted prompts have high cosine similarity with the real hidden prompts.
\revise{Output2prompt~\citep{zhang2024extracting} needs only generated text: for a hidden prefix and varying user prompts, it sparsely encodes several distinct but semantically related outputs into one aggregated representation, then reconstructs the prefix by maximum-likelihood training at lower cost than a self-attention encoder, recovering prompts that closely match the true prefixes under this output-only interface in practice.}

%%%%%%%%%%%%%% Better Language Model Inversion by Compactly Representing Next-Token Distributions %%%%%%%%%%%%%%
\textbf{$\blacktriangleright$ Reconstruct hidden prompts from compact next-token distribution representation.}
\revise{Based on \citet{morris2023language}, \citet{nazir2025better} propose Prompt Inversion from Logprob Sequences (PILS), a logprob-based attack that reconstructs hidden prompts from next-token probability sequences. PILS observes that \(\mathbf{p}=\mathrm{softmax}(W\mathbf{h})\) lies in a low-dimensional subspace determined by the hidden state \(\mathbf{h}\). Its additive log-ratio transformation gives \(\mathrm{alr}(\mathbf{p})_D=A_D\mathbf{h}\), enabling lossless compression of each probability vector. An encoder-decoder inverter then maps the compact sequence back to the hidden prompt, reducing dimensionality without requiring internal activations.}
\revise{Recently, Inv$^2$A~\citep{ye2026invariant} studies prompt inversion from a complementary latent-space perspective. It formalizes an invariant latent space hypothesis, treating the language model as an invariant decoder and learning a lightweight inverse encoder that maps model outputs to a denoised pseudo-representation for prompt recovery. This shifts prompt inversion from compressing next-token distributions to exploiting semantic and cyclic invariance across model inputs and outputs.}

\revise{\textbf{$\blacktriangleright$ Reconstruct training data from activations in decentralized LLM training.} Beyond hidden prompts, recent work targets the training corpus itself under decentralized training. \citet{dai2025stealing} propose the activation inversion attack (AIA), which builds a shadow dataset of text-activation pairs from public data, trains an attack model, and reconstructs private training text from victim activations exchanged by decentralized LLMs.}

\subsection{\revise{MIAs on Vision-Language and Foundation Models}}
\label{ssec: MIAs-foundation}
\revise{(Principle~\ref{principle: Leverage model internals}: \citep{nguyen2025vlm, caprecover2025}; Principle~\ref{principle: Extra generative models}: \citep{ddmi2025, chen2025leakyclip}; Principle~\ref{principle: Transfer pre-trained models}: \citep{otroshi2025faceadapter})}
\revise{Recent MIAs either target vision-language and multimodal models or use foundation models as reconstruction priors. We group them by the observable signal or prior: generated VLM responses, shared image-text embeddings, biometric embeddings decoded with foundation-model priors, and intermediate visual features in split VLM inference. \citet{ddmi2025,chen2025leakyclip,nguyen2025vlm} iteratively invert CLIP-style embeddings or VLM response signals, whereas \citet{otroshi2025faceadapter,caprecover2025} train adapters or inversion models for biometric embeddings and split-inference features. Under the corresponding threat model, the exposed interface determines whether recovery yields images, identities, labels, captions, or other textual semantics in practice.} \revise{Across these interfaces, prompt, embedding, activation and generated-response inversion all reconstruct an input that the model processed, and therefore remain classical MIAs under Definition~\ref{def: MI-attack}; training-data extraction differs in kind, because it elicits memorized strings by sampling the model rather than by inverting an observed signal, so we treat it as an adjacent problem rather than a form of model inversion.}

\revise{\textbf{$\blacktriangleright$ Generated-response interfaces for visual reconstruction.} For generative VLMs exposing language responses rather than a class prediction, the response sequence itself supplies inversion supervision. SMI-AW~\citep{nguyen2025vlm} replaces fixed-logit maximization with token- and sequence-level signals reflecting visually grounded concepts, adaptively weighting informative response tokens to better supervise reconstruction of private visual training images. VLM language interfaces thus leak visual training information even when no image classifier is exposed.}

\revise{\textbf{$\blacktriangleright$ Shared image-text embedding interfaces.} The shared image-text space of CLIP-style models is itself an inversion interface. DDMI~\citep{ddmi2025} replaces GAN priors with a distilled single-step diffusion generator, cutting the sampling cost of diffusion-prior inversion. LeakyCLIP~\citep{chen2025leakyclip} combines adversarial fine-tuning, embedding alignment, and Stable Diffusion refinement to extract CLIP training data, showing multimodal rather than text-only leakage.}

\revise{\textbf{$\blacktriangleright$ Foundation-prior interfaces for biometric embeddings.} Face-recognition services often expose compact biometric embeddings rather than multimodal outputs, and a foundation model can supply the prior space needed for reconstruction. Face-Adapter~\citep{otroshi2025faceadapter} learns an adapter translating embeddings from a black-box face-recognition model into a face foundation model's embedding space, enabling identity-consistent face reconstruction. Foundation models thus serve not only as attack targets but as priors that make compressed biometric representations more invertible.}

\revise{\textbf{$\blacktriangleright$ Split-VLM feature interfaces for semantic recovery.} In split-VLM inference, intermediate visual features travel from the client-side visual encoder to the cloud-side language component and can disclose high-level semantics. CapRecover~\citep{caprecover2025} recovers semantic text such as labels and captions directly from these features, targeting semantic disclosure without reconstructing the input image. This complements prompt- and activation-based inversion by showing that textual semantics learned by multimodal foundation models are exposed at the representation level.}

\revise{\textbf{$\blacktriangleright$ Comparative observations.} The exposed interface governs text-reconstruction feasibility: embeddings preserve semantics but may omit word order; logprob sequences and responses reveal prompts under visibility and query limits; activations and split features are richer but need training or collaborative access. Training-based decoders amortize recovery yet suffer domain and model shift, whereas iterative correction pays per-example query and decoding cost. Foundation models may be the target, the exposed interface, or the reconstruction prior---distinct roles.}

\section{\revise{Model Inversion Attacks on Graph Data}}
\label{sec: MIAs-graph}

% \begin{figure}[htbp]
%     \centering
%     \includegraphics[width=9cm]{figures/Graph_Illustration.pdf}
%     \caption{MIAs on graphs.
%     }
%     \label{fig:MIAs-graph}
%     \vspace{-14pt}
% \end{figure}

% \begin{figure}[htbp]
%     \centering
%     \hspace{-5pt}\includegraphics[width=0.98\linewidth]{figures/MIA_attack_graph.pdf}
%     \vspace{-15pt}
%     \caption{
%     Illustration of model inversion approaches on graph domain and the recovered examples.
%     An image was recovered using the MIAs (left) and a training image of the target model (right). 
%     The attacker is given only the person’s name, \ie the target class, and access to the target model that returns the confidence score of a class.
%     }
%     \label{fig:MIAs-attack-graph}
%     \Description[Illustration of model inversion approaches on graph domain]{Illustration of model inversion approaches on graph domain}
% \end{figure}

\begin{figure}[tb]
    \centering
    \artwork{fig:MIAs-attack-graph-timeline}
    \hspace{-5pt}\begin{tikzpicture}
      \node[anchor=south west,inner sep=0] (miagraphtimeline) at (0,0)
        {\includegraphics[width=0.95\linewidth]{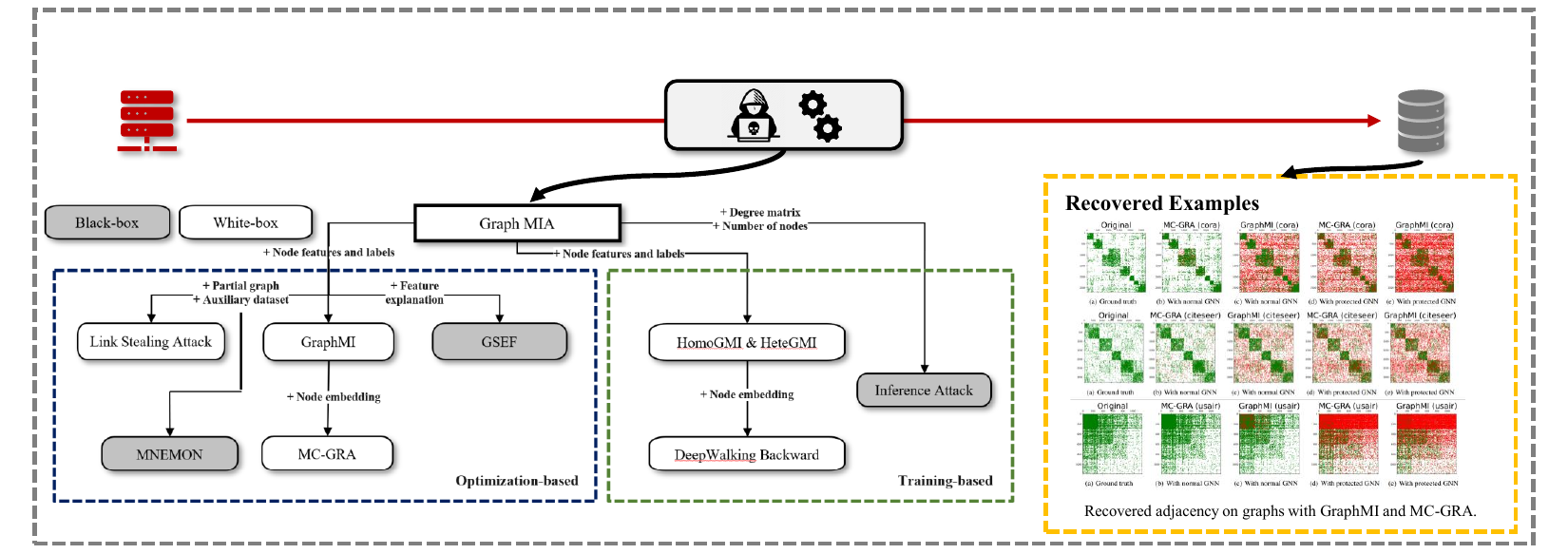}};
      \begin{scope}[x={(miagraphtimeline.south east)},y={(miagraphtimeline.north west)}]
        \node[fill=white,inner xsep=3pt,inner ysep=1.5pt,font=\bfseries\fontsize{7.3}{7.6}\selectfont,text=black] at (0.211,0.913) {Model Inversion Approaches on Graph Data};
      \end{scope}
    \end{tikzpicture}
    \artworkend{fig:MIAs-attack-graph-timeline}
    \caption{
    \revise{Evolution of graph MIAs with representative recovered examples.}
    }
    \label{fig:MIAs-attack-graph-timeline}
    \Description[An evolutionary graph of research works in Graph MIAs]{An evolutionary graph of research works in Graph MIAs}

\end{figure}

\revise{Graphs represent arbitrary relationships through non-grid structures, such as social or molecular interactions; Fig.~\ref{fig:MIAs-attack-graph-timeline} traces this line of work. Graph neural networks (GNNs)~\citep{kipf2016semi,gilmer2017neural,kipf2016variational,zhang2018link} exploit these structures in applications such as recommendation~\citep{sun2020neighbor} and drug discovery~\citep{ioannidis2020few}, but models trained on private graphs may leak sensitive information. Graph MIAs primarily target topology (edge connectivity), and link inference can expose private social ties or proprietary structures~\citep{he2021stealing}.}

% [t] not [tb]: keeps this section-opening float off the previous page's foot.

% \begin{definition}[MIA on graphs]
% \label{def: graph reconstruction attack}
% Given a set of prior knowledge $\mathcal{K}$ and a trained GNN $f_{\thetav^*}(\cdot)$,
%     the graph reconstruction attack aims to recover the original linking relations $\bm{\hat{A}}^*$ of the training graph 
%     $\bm{G}_{\text{train}} \! = \!(A, X)$,
%     namely,
%     $\bm{\hat{A}}^* = 
%     \arg \max_{\bm{\hat{A}}} \mathbb{P}(\bm{\hat{A}} | f_{\thetav^*}, \mathcal{K})$.
%  %    \begin{align}
% 	% \bm{\hat{A}}^* = 
%  %        \arg \max_{\bm{\hat{A}}} \mathbb{P}(\bm{\hat{A}} | f_{\thetav^*}, \mathcal{K}).
%  %    \label{eqn: graph_MI_attack}
%  %    \end{align}    .
% \end{definition}

\revise{Graph MIAs face two challenges: (1) adjacency is discrete and combinatorial and graph size varies, so gradient optimization needs relaxation or projection; and (2) structural priors differ across molecular, social, and citation graphs.}

\subsection{\revise{A Summary of Model Inversion Approaches on Graph Data}}
\label{ssec: MIAs-graph-summary}

\revise{Let $\bm{G}=(V,\bm{A})$ have node set \(V\) and adjacency \(\bm{A}\in\{0,1\}^{N\times N}\), where \(A_{uv}=1\) iff \(u\) and \(v\) are connected. Define \(\tilde{\bm{A}}=\bm{A}+\bm{I}\), \(\tilde D_{vv}=\sum_u\tilde A_{vu}\), and \(\hat{\bm{A}}=\tilde{\bm{D}}^{-1/2}\tilde{\bm{A}}\tilde{\bm{D}}^{-1/2}\). Let \(\bm{X}\in\mathbb{R}^{N\times D}\) be node features, \(\bm{Y}\in\{0,1\}^{N\times C}\) one-hot labels, and \(\bm{h}^{(\ell)}_v\) the layer-\(\ell\) representation, with \(\bm{h}^{(0)}_v=\bm{x}_v\).}
% We denote $\mathbb{P}(\cdot)$ as the attack method to generate reconstructed graph $\bm{\hat{A}}$, which is the approximation of the original training graph $\bm{A}$. 
\revise{The adversary may know any subset \(\mathcal{K}\subseteq\{\bm{X},\bm{Y},\bm{h}^{(\ell)}_v\}\). Tab.~\ref{tab:summary_of_attacks_graph} summarizes graph MIAs by paradigm, interface, auxiliary resource, mechanism, recovered target, and limitation.}
% a general form of MIAs on graphs in Definition~\ref{def: graph reconstruction attack} and
% [SECOND-CUT] merged into the preceding line to recover a near-empty line.

\revise{We retain the original taxonomy's two families: optimization-based approaches~\citep{olatunji2022private,zhou2023strengthening,zhang2021graphmi,shen2022finding,liu2023model} and training-based approaches~\citep{chanpuriya2021deepwalking,he2021stealing,zhang2022inference}.}

\begin{table}[t]
\caption{\revise{Summary of graph model inversion attacks, grouped by paradigm and exposed interface.}}
\label{tab:summary_of_attacks_graph}
\tabgap
\centering
\setlength{\tabcolsep}{5pt} % Reduce the space between columns
\renewcommand{\arraystretch}{1.1} % matches the other 6pt tables
\fontsize{6}{7}\selectfont
% \scriptsize
\begin{tabular}{C{40px}|C{56px}|C{30px}|C{48px}|C{72px}|C{42px}|C{47px}}
\toprule

\textbf{Approach} & \textbf{Method (venue)} & \textbf{Interface / capability} & \textbf{Auxiliary resource} & \textbf{Distinctive mechanism} & \textbf{Recovered target} & \textbf{Reported limitation} \\
\midrule

\multirow[c]{19}{40px}[31.0pt]{\centering\revise{Optimization-\\based}}
    & \revise{GraphMI~\citep{zhang2021graphmi} (IJCAI'21)}
    & \revise{White-box gradients}
    & \revise{Node features, labels}
    & \revise{Projected gradient on relaxed adjacency + autoencoder}
    & \revise{Full adjacency}
    & \revise{Needs parameters + features and labels} \\
\cline{2-7}

    & \revise{MC-GRA~\citep{zhou2023strengthening} (ICML'23)}
    & \revise{White-box internals}
    & \revise{Node features, labels, hidden representations}
    & \revise{Layered GNN as a Markov chain; chain alignment}
    & \revise{Full adjacency}
    & \revise{Needs intermediate representations} \\
\cline{2-7}

    & \revise{MNEMON~\citep{shen2022finding} (CCS'22)}
    & \revise{Released embeddings only}
    & \revise{Node embeddings}
    & \revise{Graph metric + self-supervised structure learning}
    & \revise{Graph structure}
    & \revise{Structure only; no features or model access} \\
\cline{2-7}

    & \revise{GSEF~\citep{olatunji2022private} (PETS'23)}
    & \revise{Released explanations}
    & \revise{Node features, labels, feature explanations}
    & \revise{Explanation-vector similarity (+ features, labels)}
    & \revise{Edge existence}
    & \revise{Varies by explanation type; randomized response defeats it} \\
\cline{2-7}

    & \revise{HomoGMI / HeteGMI~\citep{liu2023model} (SecureComm'23)}
    & \revise{White-box gradients}
    & \revise{Node features, labels}
    & \revise{Classification loss + 1st-/2nd-order proximity}
    & \revise{Topology of homo-/heterogeneous graphs}
    & \revise{Needs parameters; type schema for HeteGMI} \\
\hline

\multirow[c]{13}{40px}[9.4pt]{\centering\revise{Training-\\based}}
    & \revise{Link Stealing~\citep{he2021stealing} (USENIX'21)}
    & \revise{Black-box posteriors}
    & \revise{Node features, partial target graph, auxiliary dataset}
    & \revise{Link predictor on posterior pairs; similarity scoring}
    & \revise{Edge existence}
    & \revise{Needs representative attack-training data} \\
\cline{2-7}

    & \revise{DeepWalking Backwards~\citep{chanpuriya2021deepwalking} (ICML'21)}
    & \revise{Released embeddings}
    & \revise{Degree matrix, node count}
    & \revise{Linear solve or PPMI-error minimization}
    & \revise{Adjacency, community structure}
    & \revise{Individual edges, triangle density may be lost} \\
\cline{2-7}

    & \revise{Inference Attack~\citep{zhang2022inference} (USENIX'22)}
    & \revise{Black-box graph embeddings}
    & \revise{Node features, labels, node embeddings}
    & \revise{Decoders on public graph-embedding pairs}
    & \revise{Graph size, density, subgraph, approximate topology}
    & \revise{Graph-level only; defenses shown graphically} \\
\bottomrule
\end{tabular}

\end{table}

\textbf{\revise{Optimization-based Approaches.}}
\revise{These attacks solve for an adjacency candidate $\hat A$ using an observable signal $O$---task loss, predictions, explanations, or embedding consistency---and graph priors such as sparsity or smoothness:
\(
\hat A^*=\arg\min_{\hat A}\mathcal{L}_{\mathrm{obs}}\!\left(\Phi(\hat A;f_{\thetav^*},\mathcal K),O\right)+\lambda\mathcal{R}(\hat A).
\)
Analytic inversion instead solves the embedding relation directly without training an auxiliary attack model.}
% \begin{equation}
% \bm{\hat{A}}^* = \argminA_{\hat{A}} \Ls(f_{\thetav^*}(\hat{A}, \mathcal{K}), f_{\thetav^*}(\hat{A}, X)).
% \end{equation}

\textbf{\revise{Training-based Approaches.}}
\revise{These methods train a classifier or decoder on auxiliary observation-target pairs, such as GNN outputs labeled by link existence or public graphs paired with embeddings, and apply it to victim outputs~\citep{he2021stealing,zhang2022inference}. They need representative attack-training data but neither target architecture nor parameters.}

% \begin{equation}
% \bm{\hat{A}}^* = \arg \min_{\bm{\hat{A}}} \mathcal{L}(f_{\thetav^*}(A, \mathcal{K}), Y),
% \text{ where } \bm{\hat{A}} = \mathbb{Q}(\bm{A} | f_{\thetav^*}, \mathcal{K}).
% \end{equation}

\subsection{Applying Graph MIAs to Different Scenarios}
\label{ssec: MIAs-graph-application}

\textbf{Optimization via Graph-related Knowledge.} (Principle~\ref{principle: More prior knowledge}:~\citep{he2021stealing,olatunji2022private})
% Stealing Links from Graph Neural Networks, 2020
%%%%%%%%%%%%%%%%%%%%%%%%%%%%%%%%%%%%%%%%%%%%%%%%
\revise{The pioneering Link Stealing Attack~\citep{he2021stealing} infers whether each pair of nodes in a GNN's training graph is connected. Let \(s(\phiv(u),\phiv(v))\) denote an edge score derived from target outputs or learned by an attack model; the attack predicts \(\hat A_{uv}=1\) when \(s\geq T\). A raw distance can instead be converted to a similarity or compared with the reverse inequality.}
\revise{Its prior knowledge \(\mathcal{K}\) may include node features \(X_{\text{node}}\), a partial target graph \(A_{\text{sub}}\), or an auxiliary graph \(D_{\text{auxiliary}}\). Across eight threat scenarios, the method constructs scores from prediction or entropy-vector distances (cosine, Euclidean, or correlation-based variants) and, when labeled links from a partial or auxiliary graph are available, trains an MLP link classifier.}

% Private Graph Extraction via Feature Explanations, 2022
%%%%%%%%%%%%%%%%%%%%%%%%%%%%%%%%%%%%%%%%%%%%%%%%
% However, regarding that the $A_{\text{sub}}$ might not always be available, the information may also leak from the feature exploration.
\textbf{$\blacktriangleright$ Information leakage from feature explanation.}
\revise{\citet{olatunji2022private} quantify how much training-graph structure leaks when GNN feature explanations are released. They propose an explanation-only attack, which reconstructs the graph from the similarity of explanation vectors, and an explanation-augmentation attack, which additionally exploits node features and labels. Gradient-based explanations leak the most structure despite low utility, whereas for perturbation- and surrogate-based explanations leakage grows with explanation utility. As a defense, perturbing explanation bits with randomized responses reduces the attack to random guessing in their reported experiments.}

\textbf{Optimization via Model Intermediate Information.} (Principle~\ref{principle: Leverage model internals}:~\citep{zhou2023strengthening,liu2023model}; Principle~\ref{principle: Exploit output probabilities}:~\citep{zhou2023strengthening,zhang2021graphmi,liu2023model}; Principle~\ref{principle: More prior knowledge}:~\citep{shen2022finding})
% MC-GRA
%%%%%%%%%%%%%%%%%%%%%%%%%%%%%%%%%%%%%%%%%%%%%%%%
\revise{A trained GNN may leak private topology through several correlated variables. Markov Chain-based Graph Reconstruction Attack (MC-GRA)~\citep{zhou2023strengthening} \textit{models the layered GNN computation as a chain} and reconstructs adjacency by aligning combinations of known features, labels, hidden representations, and outputs between the target and attack chains. Its information-theoretic analysis relates recovery to the information captured by these variables.}

% The overall learning objective is given as follows:
% \begin{equation}
%     \text{MC-GRA: }\boldsymbol{\hat{A}}^*=\arg\max_{\boldsymbol{\hat{A}}}\underbrace{\alpha_pI(\boldsymbol{H}_A;\boldsymbol{H}_{\boldsymbol{\hat{A}}}^i)}_{\text{propagation approximation}}\\+\underbrace{\alpha_oI(\boldsymbol{Y}_A;\boldsymbol{Y}_{\boldsymbol{\hat{A}}})+\alpha_sI(Y;\boldsymbol{Y}_{\boldsymbol{\hat{A}}})}_{\text{outputs approximation}}-\underbrace{\alpha_cH(\boldsymbol{\hat{A}})}_{\text{complexity}},
% \end{equation}

\begin{figure}[tb]
\centering
    \artwork{fig:graph_attack}
    \includegraphics[width=0.95\linewidth]{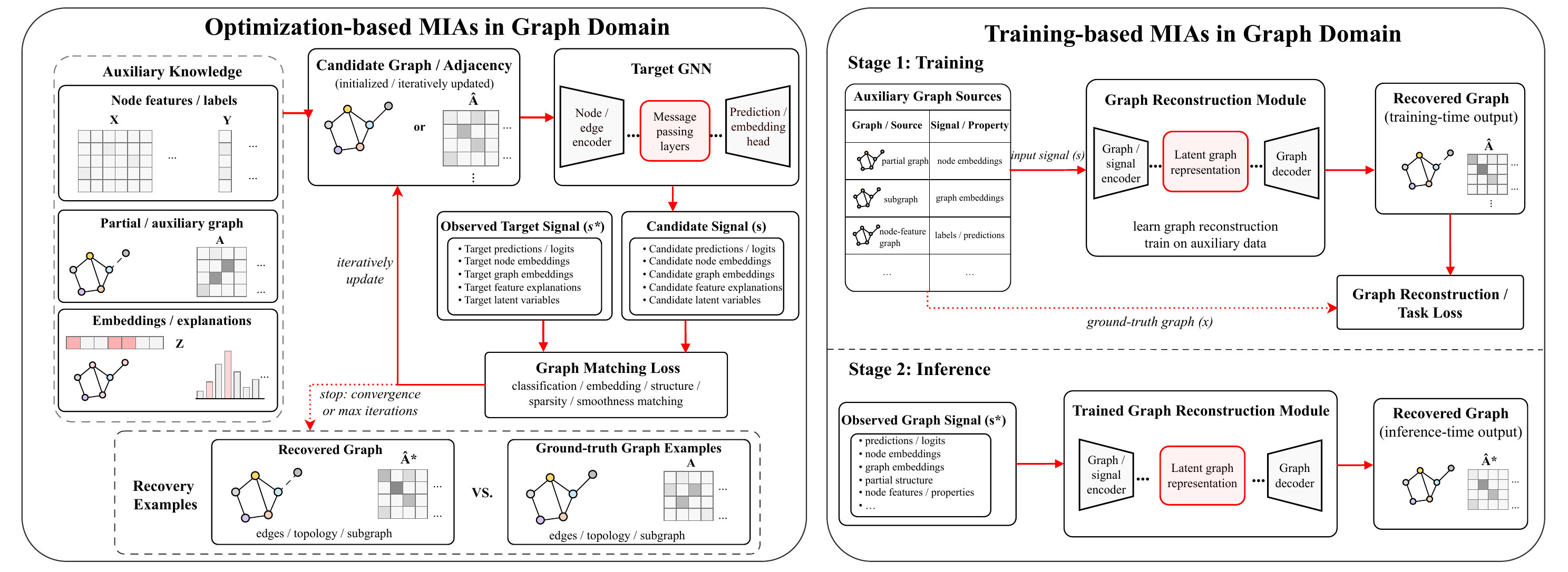}
    % -5mm, not the shared -2mm: this diagram carries ~7.5pt of blank canvas
    % below its ink where the others carry ~2pt, so it needs the extra 3mm to
    % sit the same distance from the drawing as the rest.
    \artworkend{fig:graph_attack}
    \caption{\revise{Illustrations of model inversion attacks in the graph domain.}}
    \label{fig:graph_attack}
    \label{subfig:graphmi_demo}
    \Description[Illustrations of model inversion attacks on graphs]{Illustrations of model inversion attacks on graphs}

\end{figure}

% \paragraph{\bf Optimizing with node embeddings.}
% GraphMI, 2022
%%%%%%%%%%%%%%%%%%%%%%%%%%%%%%%%%%%%%%%%%%%%%%%%
\revise{\textbf{$\blacktriangleright$ Node features and partial labels are available.} GraphMI~\citep{zhang2021graphmi}, shown in Fig.~\ref{subfig:graphmi_demo}, considers the white-box setting with
$\mathcal{K} \subseteq \{ X_{\text{node}}, Y_{\text{sub}} \}$.}
% The overview of GraphMI is shown in Figure~\ref{fig:GraphMI-overview}.
\revise{It uses projected gradient descent to handle discrete edges while regularizing topological sparsity and feature smoothness:
$\hat{A}^{*} = \arg\min_{\hat{A}} \mathcal{L}_{\text{attack}} = \mathcal{L}_{\text{cls}}(f_{\thetav}, Y_{\text{sub}}) + \alpha \mathcal{L}_{\text{smo}} + \beta \mathcal{L}_{\text{spa}},$}
\revise{where $\mathcal{L}_{\text{cls}}$ is the classification loss, $\mathcal{L}_{\text{smo}} = \nicefrac{1}{2}\sum_{i,j=1}^{N}\hat{A}_{i,j}\left\|\frac{\bm{x}_i}{\sqrt{d_i}}-\frac{\bm{x}_j}{\sqrt{d_j}}\right\|_2^2$ is the feature-smoothness loss, and $\mathcal{L}_{\text{spa}}=\|\hat A\|_F$ is the sparsity loss.}

% Finding MNEMON- Reviving Memories of Node Embeddings, 2022
%%%%%%%%%%%%%%%%%%%%%%%%%%%%%%%%%%%%%%%%%%%%%%%%
\textbf{\revise{$\blacktriangleright$ Graph MIAs via model-agnostic information.}} \revise{MNEMON~\citep{shen2022finding} introduces a \textit{graph recovery attack that relies solely on released node embeddings}, combining graph metric learning, which learns a data-specific distance, with self-supervised structure learning through a graph autoencoder. Reconstruction requires only released embeddings, not the embedding model, node features, a partial target graph, or a separately collected auxiliary graph.}

% HomoGMI and HeteGMI, 2023
%%%%%%%%%%%%%%%%%%%%%%%%%%%%%%%%%%%%%%%%%%%%%%%%
\revise{Whereas a homogeneous graph has one node and edge type, a heterogeneous graph has multiple types; this distinction is not the same as homophily, the tendency of adjacent nodes to share labels. \citet{liu2023model} propose HomoGMI and HeteGMI for these two graph types. Both are gradient-based optimization attacks combining the target GNN's classification loss with first- and second-order proximity objectives for topology recovery.}

\revise{\textbf{Inverting Released Node or Graph Embeddings}} (Principle~\ref{principle: More prior knowledge}:~\citep{chanpuriya2021deepwalking, zhang2022inference}).
% DeepWalking Backwards: From Node Embeddings Back to Graphs, 2020
%%%%%%%%%%%%%%%%%%%%%%%%%%%%%%%%%%%%%%%%%%%%%%%%
\revise{DeepWalking Backwards~\citep{chanpuriya2021deepwalking} reconstructs a graph from DeepWalk embeddings by \textit{solving a linear system} or minimizing positive pointwise mutual-information matrix error. It can preserve community structure and downstream utility, although individual edges and properties such as triangle density may be lost. For pooled graph embeddings, \citet{zhang2022inference} infer graph size, density, subgraph containment, and approximate topology, showing that released representations can disclose global and fine-grained structure even when embeddings are released solely to support downstream classification tasks.}

\revise{\textbf{$\blacktriangleright$ Comparative observations.} Signals imply different graph targets: prediction/loss attacks optimize topology under feature or label knowledge, explanations expose attribution patterns, and released embeddings can reveal edges or global properties without model access. Success depends on matching sparsity, homophily, types, and size; discrete projection and quadratic edge search limit scalability, while pooled embeddings may preserve global properties but lose individual edges. Federated-gradient reconstruction is instead adjacent signal sharing, not released-artifact inversion.}

\section{On Defending against Model Inversion Attacks}
\label{sec: defending-MIAs}

\begin{figure}[tb]
\centering      
    \artwork{fig:taxonomy_defense}
\includegraphics[width=0.95\linewidth]{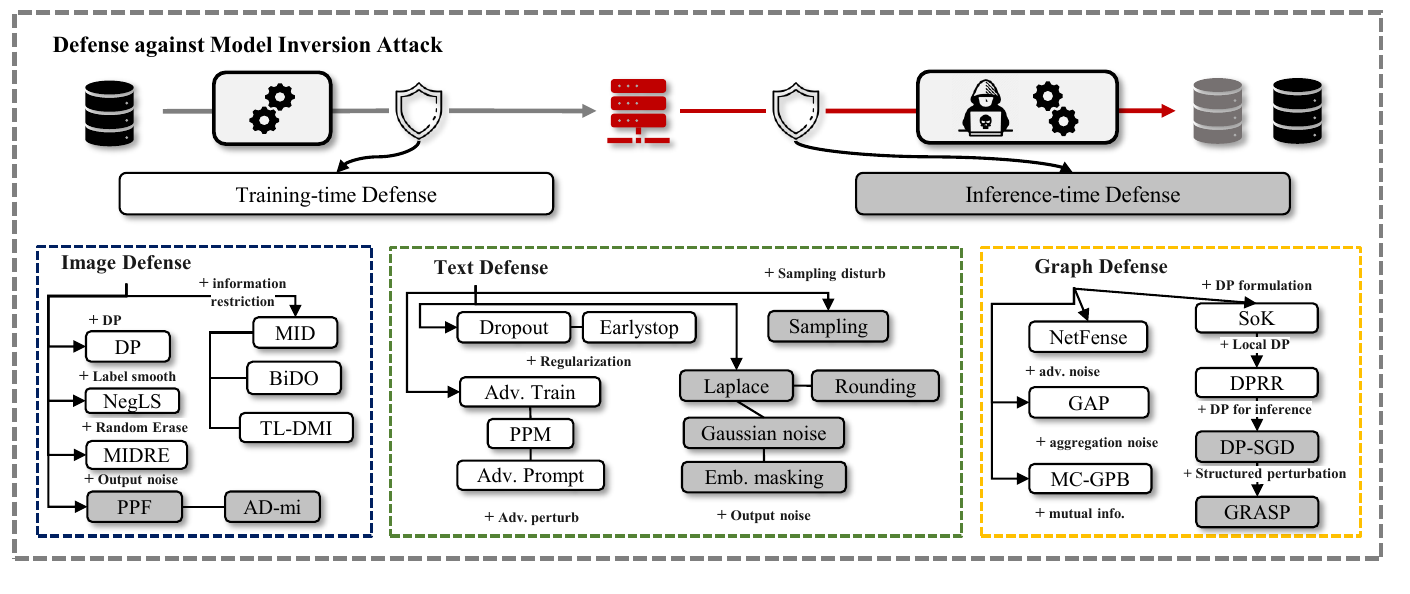}
% \caption{An illustration of defense against model inversion attacks and a taxonomy regarding the methods across different domains.}
% -5mm, not the shared -2mm: this diagram carries ~7.5pt of blank canvas below
% its ink where the others carry ~2pt, so it needs the extra 3mm to sit the
% same distance from the drawing as the rest.
    \artworkend{fig:taxonomy_defense}
\caption{A taxonomy of defense methods against model inversion attacks across different domains.}
\Description[A taxonomy of defenses against model inversion attacks]{A tree diagram grouping model inversion defenses into training-time and inference-time families across the image, text, and graph domains.}
\label{fig:taxonomy_defense}
% \vspace{-6pt}

\end{figure}

\revise{MIA countermeasures act during training or inference (Fig.~\ref{fig:taxonomy_defense}). Training-time methods weaken input-output dependence by restricting information flow or perturbing training, whereas inference-time methods perturb released outputs and mainly address restricted-access attacks. We organize both classes by domain below for clarity.}

\subsection{Defense Techniques in Image Domain}
\label{sec: defend-image}

\begin{figure}[tb]
\centering
    \artwork{fig:img_defense}
    \includegraphics[width=0.95\linewidth]{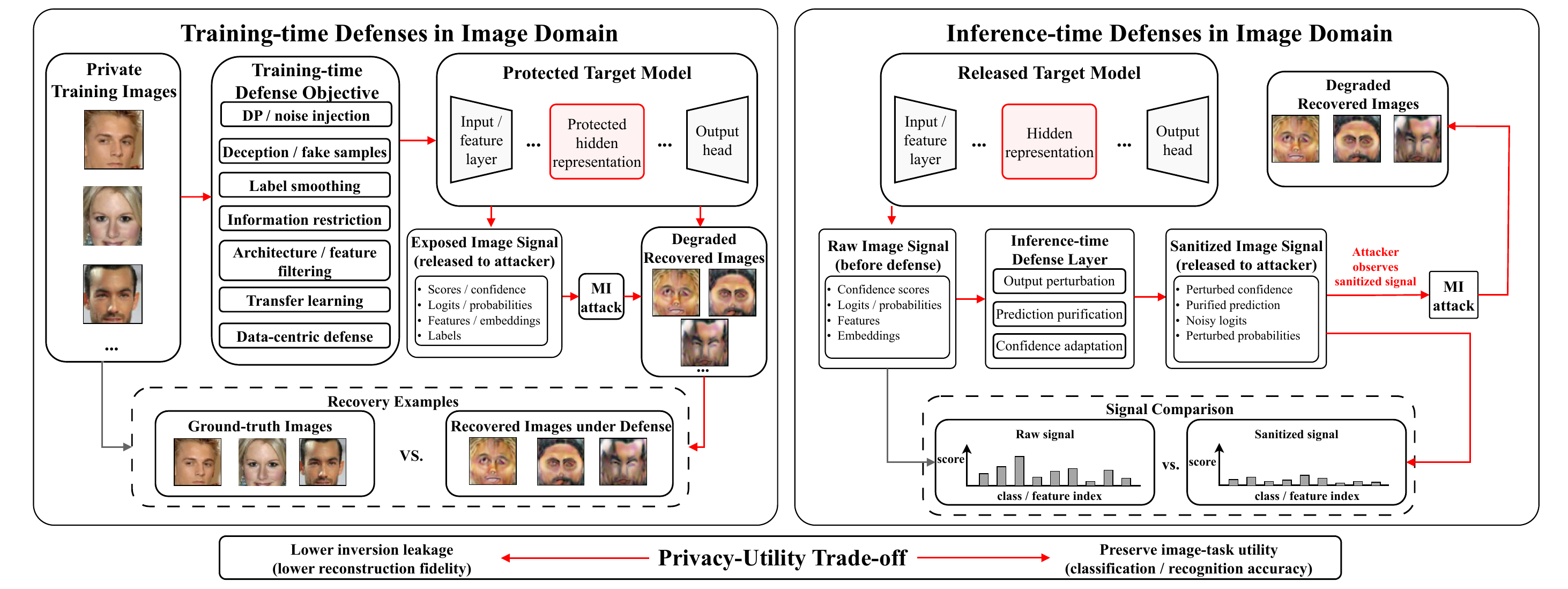}
    \artworkend{fig:img_defense}
    \caption{\revise{Illustrations of image-domain model inversion defenses.}}
    \Description[Illustrations of image-domain model inversion defenses]{Illustrations of image-domain model inversion defenses}
    \label{fig:img_defense}

\end{figure}

% In image defense, model inversion attacks leverage detailed image data to exploit redundant information in the model’s output. Defense aims to reduce the model's dependency on specific structures during training and inference. Dependency-Regularization and Architecture-Modified strategies are used in training, forming training-time defenses. To disrupt inversion, methods that alter output confidence scores without accessing the model's architecture are employed.
\revise{Image defenses act during training through differential privacy, deception, label smoothing, information restriction, transfer learning, or data-centric processing, and during inference through output perturbation (Fig.~\ref{fig:img_defense}).}

\textbf{Training-time Defense Approaches.} (Principle~\ref{principle: Data-centric processing}: \citep{fredrikson2014privacy, zhang2020secret}; Principle~\ref{principle: Regularized representation learning}: \citep{gong2023gan, LS, MID, BiDO, ho2024model})
 % At the inception of Training-time Defense, \citet{fredrikson2014privacy} first investigated how differential privacy (DP) operates by adding noise to different values or parameters. This addition of noise masks the information and provides theoretical guarantees for the privacy of the training data. However, DP does not aim to protect the entire data distribution. Empirical investigations by \citet{zhang2020secret} and \citet{fredrikson2014privacy} demonstrated that DP fails to sufficiently guard against MI attacks while preserving acceptable model performance. Furthermore, \cite{MID} provided a theoretical explanation for the limitations of DP using an indistinguishability game framework. Given these trade-offs inherent in DP, it is crucial to explore other defensive approaches.
\revise{\textit{Differential privacy} (DP) adds calibrated noise to protect record participation~\citep{fredrikson2014privacy} but does not conceal the overall training distribution, and empirically may not stop distribution-level inversion without substantial utility loss~\citep{fredrikson2014privacy,zhang2020secret}. \citet{MID} formalize this record-versus-distribution limitation through an indistinguishability game.}
 %Given these inherent trade-offs in DP, it is crucial to explore other defense approaches.
 % Therefore, we classify training-time defenses into two categories: Dependency-Regularization-Based and Architecture-Modified defenses.

\textbf{$\blacktriangleright$ Deception-based defenses.} \revise{GAN-ID~\citep{gong2023gan} trains the target with GAN-generated confounding samples so that inversion yields misleading reconstructions.}
% This framework employs an additional classifier trained on a public dataset and performs shadow MIAs on both the target and the auxiliary models.
\revise{Its weighted objective combines a private-sample loss that confuses inversion, a public-sample loss that associates fakes with the protected label, and a continual-learning loss that preserves the primary task.}
% \begin{equation}
%  \mathcal{L}(\thetav) = -\alpha \mathcal{L}_{\text{private}}(\thetav) + \beta \mathcal{L}_{\text{public}}(\thetav) + \omega \mathcal{L}_{\text{cont}}(\thetav)
% \end{equation}
% where private sample Loss ($L_{pri}$): Maximized to prevent the adversary from correctly classifying private samples,
% Public Sample Loss (Lpub): Minimized to ensure public samples resemble the protected label in the public dataset,
% continual learning loss used to maintain the victim model's performance on its primary task while integrating new fake samples.
\revise{GAN-ID needs public data and shadow attacks, protects one class, and risks unintended behavior.}

% On the other hand, \citet{LS} explored the procedure to reduce model overconfidence by adjusting the confidence score procedure. This involves using label smoothing (LS) with a smoothing factor $\alpha$ and $C$ classes, defined by the loss function:
\textbf{$\blacktriangleright$ Exploring label smoothing with MI robustness.}
\revise{Negative Label Smoothing (NegLS)~\citep{LS} uses a negative smoothing factor \(\alpha\) in a label-smoothed cross-entropy objective.}
% \begin{equation}
% \mathcal{L}(\thetav) = (1 - \alpha) \cdot \mathcal{L}_{\text{CE}}(Y, \hat{Y}) + \frac{\alpha}{C} \cdot \sum_{k=1}^{C} \mathcal{L}_{\text{CE}}(\mathbf{1}, \hat{Y}).
% \end{equation}
\revise{Positive smoothing can increase leakage when training data are scarce, whereas negative smoothing maintains high confidence away from decision boundaries and steers inversion optima away from private samples. Nevertheless, attacks such as PLG-MI still challenge these defenses~\cite{LS}.}

\revise{\textbf{$\blacktriangleright$ Architecture-level and feature-compression defenses.} Beyond the loss or the confidence scores, a recent line of defenses targets the model architecture and its intermediate features. In the architectures evaluated by \citet{koh2024skip}, skip connections reinforce MI attacks; the authors also propose MI-resilient designs that outperform prior defenses in robustness. LoFt~\citep{yu2025calor} instead constrains the rank of the intermediate features, on the finding that higher-rank features leak more, shrinking the information an attacker can recover while preserving utility. Together they show architectural and representation-level design to be an effective but underexplored axis of MI defense.}

\textbf{$\blacktriangleright$ Information restriction-based defenses.}
\revise{Dependency regularization limits information flow into learned representations. MID~\citep{MID} penalizes \textit{mutual information} between inputs and outputs, which can conflict with classification. BiDO~\citep{BiDO} instead minimizes input-feature dependence while maximizing feature-label dependence, using the Hilbert--Schmidt independence criterion (HSIC) or constrained covariance measures at the cost of additional hyperparameter tuning.}

\textbf{$\blacktriangleright$ Transfer learning-based defense.}
\revise{TL-DMI~\citep{ho2024model} uses Fisher information to identify layers that encode private data; it pre-trains on public data, \textit{freezes early layers}, and fine-tunes only later layers for the private task.}

\textbf{$\blacktriangleright$ Data-centric defenses via random erasing.}
\revise{MIDRE~\citep{tran2024defending} \textit{randomly erases image regions} during training, reducing reconstruction quality and attack accuracy while retaining natural accuracy. It complements model-centric defenses such as BiDO~\citep{BiDO}, NegLS~\citep{LS}, and TL-DMI~\citep{ho2024model}, whose combination lowers attack accuracy further.}

% \subparagraph{\bf Architecture-Modified strategy.}

% Instead of merely limiting information leakage, architecture-modified defenses aim to mislead MI attacks by altering the structure of the targeted model.

\textbf{Inference-time Defense Approaches.}  (Principle~\ref{principle: Feature masking and dropout}: \citep{yang2020defending}; Principle~\ref{principle: Modify model output}: \citep{wen2021defending})
%%%%%%%%%%%%%%%%%%%%%%%%%%%%%%%%%%%%%%%%%
\revise{The prediction purification framework (PPF)~\citep{yang2020defending} uses an autoencoder to reduce the dispersion of predictions before release, making confidence vectors less input-sensitive, and trains the purifier adversarially against a surrogate inversion attack for inference-time use.}
% This reduces the sensitivity of confidence score vectors to changes in input data, thereby decreasing their correlation and mitigating MIAs. Specifically, the purification framework is trained using a min-max game between the purifier and an adversarial model: the adversarial model $\mathrm{H}$ is trained to be the strongest inversion model, while $\mathrm{P}$ is trained to maximize $\mathrm{H}$'s inversion error. Given a feedforward DNN $\mathrm{M}: \mathbb{R}^{d_X} \rightarrow \mathbb{R}^k$ parameterized by $\thetav$, PPF is optimized with the following min-max objective:
% \begin{equation}
% \min_{\mathrm{P}} \max_{\mathrm{H}} \mathcal{L}(\mathrm{P}) - \alpha\, R(\xx, \mathrm{H}(\mathrm{P}(\mathrm{M}(x))).
% \end{equation}

\textbf{$\blacktriangleright$ Adversarial output noise.}
\revise{Because PPF may reduce utility and retain facial cues, \citet{wen2021defending} instead use inversion-model gradients to perturb predictions, maximizing reconstruction error while preserving target accuracy. A \textit{label modifier} retains the original predicted label, yielding zero accuracy loss in their experiments.}

\subsection{Defense Techniques in Text Domain}
\label{sec: defend-text}
\begin{figure}[tb]
\centering      
    \artwork{fig:text_defense}
\includegraphics[width=0.95\linewidth]{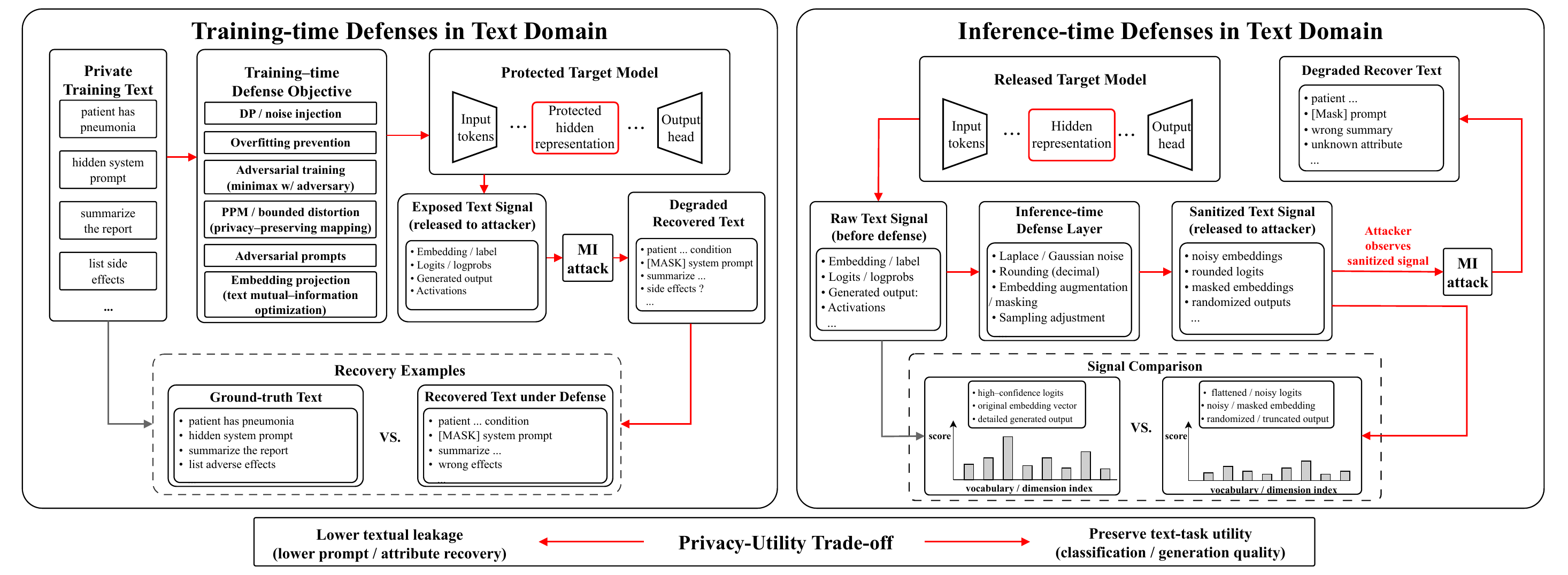}
% \caption{An illustration of defense against model inversion attacks and a taxonomy regarding the methods across different domains.}
    \artworkend{fig:text_defense}
\caption{\revise{Illustrations of text-domain model inversion defenses.}}
\Description[Illustrations of text-domain model inversion defenses]{Training-time and inference-time defenses in the text domain, showing where each defense perturbs the released signal.}
\label{fig:text_defense}
% \vspace{-6pt}

\end{figure}

\revise{In the text domain, training-time defenses use overfitting prevention, representation changes, and adversarial learning, whereas inference-time defenses use output obfuscation, embedding masking, or modified decoding and sampling (Fig.~\ref{fig:text_defense}).}

% \citet{carlini2019secret, carlini2021extracting, song2020information} attribute the private information leakage to the ability of models to memorize training data. To defend against MIAs, regularization-based defenses are employed to reduce model over-fitting, thereby decreasing the amount of training information that models can memorize. Adversarial training-based defenses involve training the model in the presence of an imagined attacker and adversarial examples to enhance the model's robustness against MIAs. Noise-based defenses introduce random noise to the data or model parameters to obscure the information that might be exploited by MIAs. Sampling-based defenses explore different sampling mechanisms of language models to reduce MIA effectiveness. Embedding-based defenses modify the data representation in ways that preserve utility while minimizing the risk of leaking sensitive information.

\textbf{Training-time Defense Approaches.} \revise{(Principle~\ref{principle: Data-centric processing}: \citep{song2020information,pan2020privacy}; Principle~\ref{principle: Regularized representation learning}: \citep{carlini2019secret,parikh2022canary})}
% \paragraph{\bf Regularization-based Defenses.}
\revise{Since memorization drives text leakage~\citep{carlini2019secret,carlini2021extracting,song2020information,parikh2022canary}, training-time defenses combine regularization, early stopping, and adversarial learning.}

\textbf{$\blacktriangleright$ Overfitting-prevention and representation defenses.}
% These methods are collectively categorized as \textit{overfitting-prevention defenses.} 
\revise{Dropout suppresses activations during training, and early stopping halts training once validation performance plateaus~\citep{parikh2022canary,arpit2017closer}. However, dropout did not significantly reduce unintended memorization in the generative sequence-model experiments of \citet{carlini2019secret}. \citet{parikh2022canary} additionally concatenate character-CNN representations with token embeddings, changing the NLU model's input representation to make canary reconstruction harder only in their evaluated NLU setting.}

\textbf{$\blacktriangleright$ Adversarial training for privacy-preserving representation learning.}
\revise{Adversarial representation learning jointly trains an encoder against a word or attribute predictor~\citep{coavoux2018privacy,elazar2018adversarial,li2018towards,xie2017controllable,edwards2015censoring,song2020information}. A privacy-preserving mapping can similarly distort embeddings within a bounded range~\citep{pan2020privacy,salamatian2015managing}. For hidden prompts, explicit non-disclosure instructions~\citep{zhang2024extracting,Zhang2023EffectivePE} can reduce empirical leakage but offer a behavioral rather than formal privacy guarantee.}
% [SECOND-CUT] duplicate of the Eguard description in the inference-time
% obfuscation paragraph below; kept there, since Eguard protects released embeddings.
% \revise{More recently, Eguard~\citep{liu2026eguard} defends LLM embeddings against inversion by passing them through a transformer-based projection network trained with a text mutual-information objective, which suppresses the information an inversion attack can recover while preserving downstream utility.}

% \begin{figure}[htbp]
% \centering
%     \artwork{fig:text_defense}
% \centering
% \vspace{-4mm}
%     \subfigure[Defending against MIAs via dropout strategy~\citep{srivastava2014dropout}]{
%         \centering
%         \includegraphics[width=0.36\linewidth]{figures/dropout.png}
%     }
%     \subfigure[Defending against MIAs via inference masking~\citep{parikh2022canary}]{
%         \centering
%         \includegraphics[width=0.6\linewidth]{figures/masking_defense.png}
%     }
%     \vspace{-4mm}
%     \caption{The illustrations of model inversion defense in text domain.}
%     \Description[The illustrations of model inversion defense on graphs]{The illustrations of model inversion defense on graphs}
%     \label{fig:text_defense}
%     \vspace{-7mm}
% \end{figure}

\textbf{Inference-time Defense Approaches.} \revise{(Principle~\ref{principle: Feature masking and dropout}: \citep{chen2024text}; Principle~\ref{principle: Modify model output}: \citep{morris2023language,pan2020privacy,chen2024text,morris2023text})}
\revise{Inference-time defenses obscure released outputs without retraining the target, using data obfuscation, embedding masking, or changes to decoding and sampling procedures at the interface exposed to attackers during black-box inference.}

\textbf{$\blacktriangleright$ Data obfuscation-based defense.}
\revise{\textit{Output obfuscation} adds Laplace~\citep{pan2020privacy,dwork2014algorithmic} or Gaussian noise~\citep{morris2023text,chen2024text}, or rounds embedding coordinates~\citep{pan2020privacy}. The multilingual masking defense instead replaces the first embedding coordinate with a language identifier~\citep{chen2024text}. Each mechanism removes or redistributes fine-grained information at a distinct utility cost.}
\revise{Eguard~\citep{liu2026eguard} maps LLM embeddings through a transformer projector trained to reduce text mutual information, thereby limiting recovery while preserving semantic utility for downstream tasks under its reported protocol.}

% \paragraph{\bf Sampling-based Defenses.}
\textbf{$\blacktriangleright$ Sampling-based defenses.}
\revise{\citet{morris2023language} reduce prompt-inversion signals by \textit{increasing sampling temperature} or restricting decoding to top-\(p\) or top-\(K\) candidates. These transformations flatten or truncate the next-token distribution used for generation, trading output quality and diversity for lower inversion performance.}

\subsection{Defense Techniques in Graph Domain}\label{sec: defend-graph}

% In defending against graph MIAs, researchers mainly focus on protecting private link information from adversaries rather than the node features. The motivation for defense is to reduce the model's reliance on the graph structures during either training or inference. DP and mutual information-constrained training strategies are employed to form the training-time defense. In contrast, to interfere with the inversion process, inference-time defense methods introduce noise during inference without damaging the model's performance.
% % 1. "Graph defense" is not a rigorous term;
% % 2. "Privacy link information" -> "private link information"; 
% % 3. "Adversary" is a more commonly adopted term in the adversarial domain, instead of "adversary".
% % }

% % Privacy and transparency in graph machine learning: A unified perspective, 2022
% %%%%%%%%%%%%%%%%%%%%%%%%%%%%%%%%%%%%%%%%%%%%%%%%
% GraphML~\citep{khosla2022privacy} offers a unified perspective on privacy and transparency in graph machine learning, highlighting that transparency and privacy are typically studied separately but often have conflicting objectives. While transparency provides insights into model decisions, privacy aims to protect sensitive training data. 
% We provide detailed discussions of works related to these two categories below.

\begin{figure}[tb]
\centering
    \artwork{fig:graph_defense}
    \includegraphics[width=0.95\linewidth]{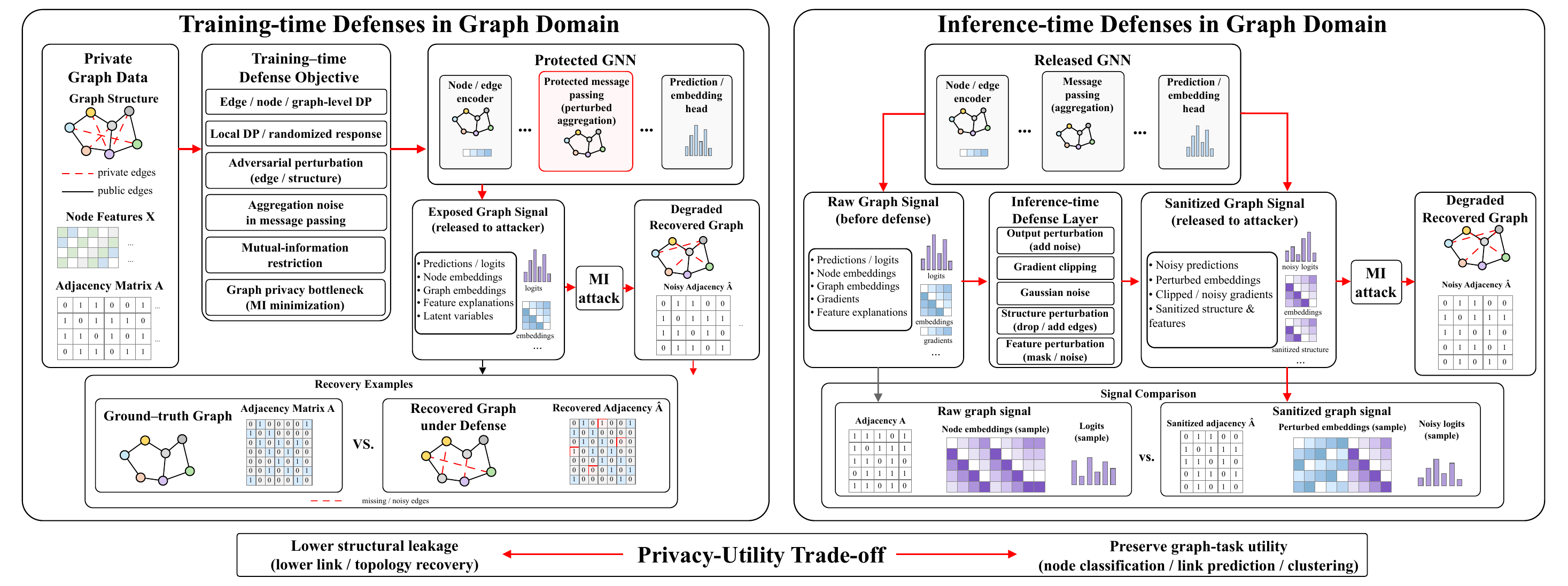}
    \artworkend{fig:graph_defense}
    \caption{\revise{Illustrations of graph-domain model inversion defenses.}}
    \Description[Illustrations of graph-domain model inversion defenses]{Illustrations of graph-domain model inversion defenses}
    \label{fig:graph_defense}

\end{figure}

\revise{Defenses against graph MIAs mainly protect \textit{private link information} rather than node features. Training-time methods perturb graph processing or restrict topology-dependent information, whereas inference-time methods perturb embeddings or explanations exposed to users (Fig.~\ref{fig:graph_defense}). \citet{khosla2022privacy} notes that transparency may expose sensitive structure.}

\textbf{Training-time Defense Approaches.} (Principle~\ref{principle: Regularized representation learning}:~\citep{hsieh2021netfense,mueller2022differentially,hidano2022degree,sajadmanesh2023gap,zhou2023strengthening})
% NetFense- Adversarial Defenses against Privacy Attacks on Neural Networks for Graph Data， 2021
%%%%%%%%%%%%%%%%%%%%%%%%%%%%%%%%%%%%%%%%%%%%%%%%
\revise{NetFense~\citep{hsieh2021netfense} is a graph-preprocessing defense rather than a differential-privacy mechanism. It selects and optimizes inconspicuous edge perturbations that preserve graph and model utility while reducing an attacker's confidence in private-label inference.} 
%Consider the perturbing edge $(k, m)$ and the original normalized two-hop  graph $\hat{A}^{2}$, the updated graph $\hat{A'}^{2}$ is given by:
% \begin{equation}
% \begin{split}
% \hat{A}_{(i,j)}^{2} =\frac1{\sqrt{\tilde{d^{\prime}}_i\tilde{d^{\prime}}_j}}{\left[\sqrt{\tilde{d}_i\tilde{d}_j}\hat{A}_{(i,j)}^2+\left(\frac{\tilde{a^{\prime}}_{(i,j)}}{\tilde{d^{\prime}}_i}-\frac{\tilde{a}_{(i,j)}}{\tilde{d}_i}+\frac{\tilde{a^{\prime}}_{(i,j)}}{\tilde{d^{\prime}}_j}-\frac{\tilde{a}_{(i,j)}}{\tilde{d}_j}\right.\right)}\\
% \left.+\left(\frac{a^{\prime}{}_{(i,k)}a^{\prime}{}_{(k,j)}}{\tilde{d^{\prime}}{}_{k}}-\frac{a_{(i,k)}a_{(k,j)}}{\tilde{d}_{k}}+\frac{a^{\prime}{}_{(i,m)}a^{\prime}{}_{mj}}{\tilde{d^{\prime}}{}_{m}}-\frac{a_{(i,m)}a_{(m,j)}}{\tilde{d}_{m}}\right)\right], 
% \end{split}
% \end{equation}
% where $\mathbb{I}(c) = 1$ if condition $c$ is satisfied, $\tilde{d}_{i}^{\prime} =\tilde{d}_i+\mathbb{I}(i\in\{k,m\})(1-2\cdot a_{(k,m)})$, $a_{(i,j)}^{\prime} =a_{(i,j)}+\mathbb{I}((i,j)=(k,m))(1-2\cdot a_{(i,j)})$, and $\tilde{a}_{(i,j)}^{\prime} =\tilde{a}_{(i,j)}+\mathbb{I}((i,j)=(k,m))(1-2\cdot a_{(i,j)})$.
% NetFense shows that perturbing graph structure can preserve community structure and reduce confidence in predicting private labels while maintaining performance.

% sok, 2022
%%%%%%%%%%%%%%%%%%%%%%%%%%%%%%%%%%%%%%%%%%%%%%%%
\revise{\textbf{$\blacktriangleright$ Employing differential privacy.} Graph DP can protect edges, nodes, or whole graphs; the SoK by \citet{mueller2022sok} systematizes these neighboring-dataset definitions but is a survey, not itself a defense algorithm. For multi-graph classification, \citet{mueller2022differentially} apply DP-SGD during training by clipping per-example gradients and adding Gaussian noise before parameter updates; their GNNExplainer analysis compares private and non-private representations rather than adding noise to final predictions.}
% Degree-Preserving Randomized Response for Graph Neural Networks under Local Differential Privacy, 2022
%%%%%%%%%%%%%%%%%%%%%%%%%%%%%%%%%%%%%%%%%%%%%%%%
Degree-Preserving Randomized Response (DPRR)~\citep{hidano2022degree} proposes a \textit{local DP} algorithm designed to protect edges in GNNs. DPRR combines randomized response~\citep{warner1965randomized} with \textit{strategic edge sampling} to preserve degree information under edge DP. 
%To sampling the , given $\epsilon_1, \epsilon_2 \in \mathbb{R}_{\geq 0}$, let $\text{Lap}(\frac{1}{\epsilon_1})$ be the Laplacian noise with mean $0$ and scale $\frac{1}{\epsilon_1}$, the $d_i$'s private estimate $d^*_i = d_i + \text{Lap}(\frac{1}{\epsilon_1})$. To determine the neighbor $ \forall j\in[n]\setminus\{i\}$ of node $i$, $\tilde{a}_i$ is given by:
% \begin{equation}
% \begin{split}
% \Pr(\tilde{a}_{i,j}=1)=\begin{cases}pq_i(\mathrm{if~}a_{i,j}=1)\\
% (1-p)q_i(\mathrm{otherwise})&\end{cases}, \text{where } q_i=\frac{d_i^*}{d_i^*(2p-1)+(n-1)(1-p)}, p=\frac{e^{\varepsilon_2}}{e^{\varepsilon_2}+1}.
% \end{split}
% \end{equation}
\revise{For graph classification over unattributed social graphs, DPRR provides edge DP while approximately preserving degree information. It outperforms randomized response, local Laplacian noise, and non-private-user-only baselines, especially when some users are non-private in its reported experiments.}

% GAP: Differentially Private Graph Neural Networks with Aggregation Perturbation, 2023
%%%%%%%%%%%%%%%%%%%%%%%%%%%%%%%%%%%%%%%%%%%%%%%%
\revise{\textbf{$\blacktriangleright$ Aggregation perturbation.} GAP~\citep{sajadmanesh2023gap} injects \textit{independent Gaussian noise} into node-level aggregation and accounts for its privacy loss, yielding edge- or node-level DP for multi-hop GNNs. This is stochastic perturbation, not adversarial training.}
%, \textit{i.e.,} $\widetilde{\mathbf{h}}_\nu^{(l)}=\mathbf{h}_\nu^{(l)}+\mathcal{N}(\sigma^2\mathbf{I}), \forall\nu\in\mathcal{V}$, where $\sigma$ denote the variance, $\mathbf{I}$ is the identical matrix. 
\revise{GAP outperforms node-classification baselines while quantifying its privacy-utility trade-off.}

% MC-GPB
%%%%%%%%%%%%%%%%%%%%%%%%%%%%%%%%%%%%%%%%%%%%%%%%
\textbf{$\blacktriangleright$ Employing information restriction during training.} This line of work aims to minimize the model's reliance on private links during training. 
\revise{Markov Chain-based Graph Privacy Bottleneck (MC-GPB)~\citep{zhou2023strengthening} prevents leakage by minimizing mutual information between the \textit{node embedding matrix} and the graph's \textit{adjacency matrix}. Its Graph Information Plane tracks leakage and suggests that GNNs first memorize and later forget private information.}
%Based on this insight, MC-GPB's training objective is given as follows:
% \begin{equation}
%     \text{MC-GPB:}\boldsymbol{\thetav}^*=\arg\min_{\boldsymbol{\thetav}}\sum_{\ell=1}^L\underbrace{-I(Y;\bm{h}^{(\ell)}_{v}\}}_{\text{accuracy}}+\underbrace{\beta_p^iI(A;\bm{h}^{(\ell)}_{v}\})}_{\text{privacy}}+\sum_{\ell=1}^{L-1}\underbrace{\beta_c^iI(\bm{h}^{(\ell)}_{v}\};\bm{h}^{(\ell+1)}_{v}\})}_{\text{complexity}}.
% \end{equation}

\revise{\textbf{Inference-time Defense Approaches.}} (Principle~\ref{principle: Modify model output}:~\citep{zhang2022inference,guo2025grasp})
\revise{When graph embeddings or intermediate aggregations must be exposed, perturbing that interface can reduce reconstruction. \citet{zhang2022inference} evaluate \textit{output perturbation} against property, subgraph, and topology inference from graph embeddings. GRASP~\citep{guo2025grasp} uses a Bernoulli selector to apply shared or independent Gaussian noise per node embedding. This shifts pairwise similarities in both directions, disrupting reconstruction rankings while retaining edge-level DP during training and inference.}

\subsection{\revise{Privacy--Utility Trade-off of Defenses}}
\label{sec:defense-tradeoff}

\revise{Every defense trades exposed information against accuracy, fidelity, computation, or applicability. The controlling variable may be a privacy budget or noise scale, a regularization weight, a smoothing or erasing factor, frozen layers, or a sampling parameter; costs can also include public data, adversarial training, tuning, or extra inference. Since studies use different attacks, datasets, metrics, and privacy notions, universal Low/Medium/High ratings would be misleading. Tab.~\ref{tab:defense-tradeoff} therefore reports paper-specific privacy and downstream-utility evidence where the original studies provide it, together with each method's control and applicability boundary in its original setting and threat model.}

\begin{table*}[t]
\centering
\caption{\revise{Privacy--utility evidence for representative defenses. Reported values are paper-specific and not directly comparable; Train and Inf.\ denote training- and inference-time defenses; Arch.\ and Repr.\ refine the training-time class.}}
\Description[Privacy--utility comparison of representative model inversion defenses]{Privacy--utility comparison of representative model inversion defenses by domain, stage, control variable, effect, and cost or limitation.}
\label{tab:defense-tradeoff}
\setlength{\tabcolsep}{2pt}
\renewcommand{\arraystretch}{1.02}
\fontsize{6.15}{6.85}\selectfont
{\revcolor
\begin{tabular}{@{}L{69pt}L{36pt}L{76pt}L{126pt}L{107pt}@{}}
% One light rule per row: the evidence and boundary cells wrap to three or four
% lines each, and without a separator a reader cannot tell where a row ends.
% \cmidrule is booktabs' thinnest full-width rule (0.03em vs \midrule's 0.05em);
% booktabs' own rule padding is left at its default, so this adds no style deviation.
\toprule
\textbf{Defense} & \textbf{Dom./ stage} & \textbf{Control} & \textbf{Reported privacy / utility evidence} & \textbf{Cost / boundary} \\
\midrule
GAN-ID; MID/BiDO~\citep{gong2023gan,MID,BiDO} & Image/Train & loss/dependence weights & BiDO-HSIC, CelebA/KEDMI: attack acc.\ $>29\%$ lower at $\sim6\%$ classifier acc. & public/shadow data; tuning cost \\
\cmidrule(l{0pt}r{0pt}){1-5}
NegLS; MIDRE~\citep{LS,tran2024defending} & Image/Train & smoothing/erasing & NegLS ($\alpha=-0.05$), FaceScrub/PPA: top-1 attack acc.\ $80.0\%$ lower at $\sim3.4\%$ test acc. & calibration or input modification; gains shrink at low resolution \\
\cmidrule(l{0pt}r{0pt}){1-5}
TL-DMI~\citep{ho2024model} & Image/Train & frozen layers & CelebA/KEDMI: top-1 attack acc.\ $39\%$ lower at $\sim5.6\%$ classifier acc. & public pre-training; classification only \\
\cmidrule(l{0pt}r{0pt}){1-5}
PPF / adv. noise~\citep{yang2020defending,wen2021defending} & Image/Inf. & contraction/noise & FaceScrub inversion error: purification $0.0114\to0.0448$ at $<0.4\%$ acc.\ cost & attack-dependent; the noise gain collapses to $0.0271$ under an adaptive attacker \\
\cmidrule(l{0pt}r{0pt}){1-5}
LoFt~\citep{yu2025calor} & Image/Repr. & feature-rank constraint & FaceScrub/IR-152: IF top-1 attack acc.\ $82.2\to13.0\%$ at test acc.\ $98.2\to97.1\%$ & dimension-dependent: attack acc.\ back to $77.8\%$ at dimension 512; image classification only \\
\cmidrule(l{0pt}r{0pt}){1-5}
Skip-connection design~\citep{koh2024skip} & Image/Arch. & last-stage skip connections & FaceScrub/PPA, ResNet-101: removing stage-4 skips $83.0\to6.8\%$; deployable RoLSS $83.0\to58.7\%$ at acc.\ $94.9\to92.4\%$ & requires retraining; architecture-specific; evaluated on face recognition \\
\midrule
Regularization~\citep{carlini2019secret,song2020information} & Text/Train & dropout/stopping/loss & dropout/weight decay/early stopping leave canary exposure unchanged; DP-SGD: $31.0\to2.8$ for $+8\%$ test loss & no formal guarantee without DP; adversarial-training utility cost only graphical \\
\cmidrule(l{0pt}r{0pt}){1-5}
Eguard~\citep{liu2026eguard} & Text/Inf. & projection/MI loss & SST2/NLI/retrieval/summarization: inversion F1/recall $\sim4\%$, $>98\%$ of accuracy/ROUGE retained & extra projector; $1.6$--$3.4\times$ one-time training overhead \\
\cmidrule(l{0pt}r{0pt}){1-5}
Obfuscation/sampling~\citep{pan2020privacy,chen2024text,morris2023language} & Text/Inf. & noise/masking/sampling & language-ID masking, BEIR: BLEU $61.6\to4.7$ with NDCG@$10$ unchanged & far weaker on multilingual encoders; sampling defenses report no utility metric \\
\midrule
NetFense~\citep{hsieh2021netfense} & Graph/Train & edge budget & CiteSeer: private-label acc.\ $84.0\to68.5\%$ at target-label acc.\ $98.0\to84.5\%$ & measured on the attacked node subset, not the full test set \\
\cmidrule(l{0pt}r{0pt}){1-5}
DPRR / GAP~\citep{hidano2022degree,sajadmanesh2023gap} & Graph/Train & privacy budget/noise & GAP-NDP, Facebook: node-MI AUC $50.2$--$52.7\%$ at $\epsilon=1$--$16$; node acc.\ $63.2\%$ at $\epsilon=8$ & formal node DP; privacy strength and dataset affect utility \\
\cmidrule(l{0pt}r{0pt}){1-5}
MC-GPB~\citep{zhou2023strengthening} & Graph/Train & privacy weight & Cora: non-learnable GRA AUC $76.6\to70.6\%$; node acc.\ down $3.0\%$ rel. & estimation cost; on USA the same setting costs $16.8\%$ accuracy \\
\cmidrule(l{0pt}r{0pt}){1-5}
Embedding perturbation / GRASP~\citep{zhang2022inference,guo2025grasp} & Graph/Inf. & noise/selector & GRASP (GCN, $\sigma=1.0$), Cora: GRA AUC $95.8\to74.8\%$ at node acc.\ $87.8\to87.5\%$ & GRASP reports $\sigma$, not $\epsilon$; Zhang et al.\ report defense curves only graphically \\
\bottomrule
\end{tabular}}

\end{table*}

\revise{The trade-off is interface- and domain-dependent: output perturbation does not protect gradients or manipulated split-learning representations; text defenses may preserve classification while degrading embedding or generation quality; and graph defenses must identify the protected object. The reported values above are illustrative rather than a ranking because datasets, attacks, evaluators, and privacy strengths differ. Robustness should therefore be reported with adaptive attacks, task utility, resource cost, and the exact privacy notion under the stated protocol.}

% \begin{figure}[htbp]
%     \centering
%     \includegraphics[width=9cm]{figures/MIAs-defense-1.png}
%     \caption{Overview of the BiDO method~\citep{peng2022bilateral}.}
%     \label{fig:BIDO-overview}
% \end{figure}

% \begin{figure}[htbp]
%     \centering
%     \includegraphics[width=7cm]{figures/MIAs-defense-2.png}
%     \caption{Examples of defending MI attack from~\citep{peng2022bilateral}.}
%     \label{fig:MI-defense-sample}
%     \vspace{-14pt}
% \end{figure}

% \vspace{-10pt}

% \clearpage
\section{Datasets and Evaluation Metrics}
\label{sec: dataset-and-evaluation}

\revise{MIBench~\citep{qiu2024mibench} provides a standardized, modular benchmark and toolbox integrating representative attacks and defenses under common protocols. It standardizes the data-preparation, attack, defense, and evaluation stages so that reported numbers share matched target models and auxiliary data, which removes much of the setup variance that otherwise makes published numbers incomparable. Its scope, however, is the image domain rather than the three modalities covered here, so cross-domain comparison still rests on the per-domain metrics below.}

% \vspace{-4pt}
\subsection{Datasets}

\revise{Tab.~\ref{table::compact_dataset_candidate} lists datasets across the three domains with their scale, structure, and typical role in MI studies, supporting reproducibility and controlled comparisons among inversion attacks under common threat models.}

\begin{table}[t]
\captionsetup{labelfont={rm,small},textfont={rm,small}}
\caption{\revise{Cross-domain datasets used in model inversion studies, grouped by data role, with scale ranges for canonical releases and family shading for Image, Text, and Graph. Individual studies may use subsets or preprocessing variants.}}
\label{table::compact_dataset_candidate}
\centering
\setlength{\tabcolsep}{3pt}
\renewcommand{\arraystretch}{0.98} % [SECOND-CUT] tighter rows
\fontsize{7}{7.8}\selectfont
\begin{lrbox}{\compactdatasetrevisionbox}
\begin{minipage}{\linewidth}
\revcolor
\tabgap
\begin{tabularx}{\linewidth}{@{}L{78pt}>{\raggedright\arraybackslash}X L{85pt}L{98pt}@{}}
\toprule
\textbf{Family} & \textbf{Datasets (sources)} & \textbf{Scale / structure} & \textbf{Typical MIA use}\\
\midrule
\cellcolor{dataset-image-bg}\textcolor{black}{\textbf{face / biometric}} &
CelebA~\citep{liu2015deep}; \revise{CelebA-HQ}~\citep{karras2018progressive}; FaceScrub~\citep{ng2014data}; \revise{LFW}~\citep{huang2007lfw}; FFHQ~\citep{StyleGAN}; VGGFace2~\citep{cao2018vggface2}; \revise{VGGFace}~\citep{parkhi2015deep}; \revise{CASIA-WebFace}~\citep{yi2014casia}; \revise{MOBIO}~\citep{mccool2013mobio}; \revise{AgeDB}~\citep{moschoglou2017agedb}; AT\&T Face~\citep{samaria1994parameterisation}; MetFaces~\citep{karras2020training}; PubFig83~\citep{pinto2011pubfig83}; iCV MEFED~\citep{loob2017dominant} &
400--3.31M images; up to 10,575 identities &
Identity, attribute, and facial-expression reconstruction; face-embedding inversion\\
\specialrule{0.15pt}{0.25pt}{0.25pt}

\cellcolor{dataset-image-bg}\textcolor{black}{\textbf{general classification}} &
MNIST~\citep{lecun1998gradient}; QMNIST~\citep{yadav2019cold}; Fashion-MNIST~\citep{xiao2017fashion}; CIFAR-10~\citep{krizhevsky2009learning}; ImageNet1K~\citep{russakovsky2015imagenet}; ImageNet~\citep{deng2009imagenet}; \revise{TinyImageNet}~\citep{deng2009imagenet}; Omniglot~\citep{lake2015human} &
32.5K--14.2M images; 10--21,841 classes &
Class-level reconstruction under white-box, score-based, or label-only access\\
\specialrule{0.15pt}{0.25pt}{0.25pt}

\cellcolor{dataset-image-bg}\textcolor{black}{\textbf{fine-grained recognition}} &
Stanford Dogs~\citep{dataset2011novel}; AFHQ Dogs~\citep{choi2020stargan}; Oxford IIIT Pet~\citep{parkhi2012cats}; Amur Tigers~\citep{li2019atrw}; \revise{Stanford Cars}~\citep{krause2013cars}; \revise{CUB-200-2011}~\citep{wah2011cub} &
7.3K--20.6K images; 37--200 classes &
Fine-grained class or identity reconstruction with learned image priors\\
\specialrule{0.15pt}{0.25pt}{0.25pt}

\cellcolor{dataset-image-bg}\textcolor{black}{\textbf{medical}} &
ChestX-ray8~\citep{wang2017chestx}; \revise{ChestX-ray14}~\citep{wang2017chestx}; \revise{CheXpert}~\citep{irvin2019chexpert} &
108K--224K images; 8/14 multilabel findings &
Reconstruction of sensitive diagnostic images and attributes\\
\specialrule{0.15pt}{0.25pt}{0.25pt}

\cellcolor{dataset-image-bg}\textcolor{black}{\textbf{vision-language}} &
\revise{LAION-400M}~\citep{schuhmann2021laion400m}; \revise{LAION-2B}~\citep{schuhmann2022laion}; \revise{Furniture Object}~\citep{lohia2024furniture}; \revise{Flickr30K}~\citep{young2014flickr30k}; \revise{COCO}~\citep{lin2014coco}; \revise{Flickr8K}~\citep{hodosh2013flickr8k} &
166 images to 2.32B image-text pairs &
CLIP/VLM image reconstruction and semantic recovery\\
\specialrule{0.9pt}{1.2pt}{0.8pt}
\cellcolor{dataset-text-bg}\textcolor{black}{\textbf{intent / classification}} &
ATIS~\citep{dahl1994expanding}; NLU-Evaluation~\citep{liu2021benchmarking}; Snips~\citep{coucke2018snips}; Emotion Dataset~\citep{saravia2018carer}; Yelp~\citep{zhang2015character}; IMDB~\citep{IMDB}; AG News~\citep{zhang2015character} &
4,978--700K utterances or documents &
Intent, label, sentence, and embedding reconstruction\\
\specialrule{0.15pt}{0.25pt}{0.25pt}

\cellcolor{dataset-text-bg}\textcolor{black}{\textbf{language modeling}} &
WikiText~\citep{merity2016pointer}; BookCorpus~\citep{zhu2015aligning}; \revise{Wikipedia}~\citep{song2020information}; MTG~\citep{chen2021mtg}; \revise{ArXiv}~\citep{dai2025stealing}; \revise{OpenWebText}~\citep{dai2025stealing}; \revise{The Pile}~\citep{dai2025stealing}; \revise{ArXiv Math}~\citep{ye2026invariant} &
Large corpora spanning 100M tokens to millions of documents &
Training-text memorization and activation inversion\\
\specialrule{0.15pt}{0.25pt}{0.25pt}

\cellcolor{dataset-text-bg}\textcolor{black}{\textbf{QA / retrieval}} &
QNLI~\citep{wang2018glue}; Natural Questions~\citep{kwiatkowski2019natural}; MSMARCO~\citep{bajaj2016ms}; \revise{BEIR}~\citep{thakur2021beir}; \revise{CLIRMatrix}~\citep{sun2020clirmatrix} &
18-dataset benchmark; up to 8.8M passages &
Embedding inversion and recovery of queries, passages, or answers\\
\specialrule{0.15pt}{0.25pt}{0.25pt}

\cellcolor{dataset-text-bg}\textcolor{black}{\textbf{dialogue / preference}} &
PersonaChat~\citep{zhang2018personalizing}; Helpfulness \& Harmlessness~\citep{bai2022training}; ShareGPT~\citep{ShareGPT} &
8,939--433,723 conversations or preference pairs &
Dialogue, response, and preference-data reconstruction\\
\specialrule{0.15pt}{0.25pt}{0.25pt}

\cellcolor{dataset-text-bg}\textcolor{black}{\textbf{prompts / instructions}} &
Unnatural Instructions~\citep{honovich2022unnatural}; Instructions-2M~\citep{morris2023language}; Alpaca~\citep{taori2023stanford}; Synthetic GPTs~\citep{zhang2024extracting}; \revise{Real GPTs}~\citep{zhang2024extracting}; Awesome-ChatGPT-Prompts~\citep{Awesome-ChatGPT-Prompts}; Alpaca Code~\citep{codealpacas}; \revise{Instruction corpora}~\citep{ye2026invariant} &
153--2.33M prompts or instructions &
Hidden-prompt, system-prompt, and instruction-data inversion\\
\specialrule{0.15pt}{0.25pt}{0.25pt}

\cellcolor{dataset-text-bg}\textcolor{black}{\textbf{clinical}} &
MIMIC-III~\citep{johnson2016mimic} &
2.08M clinical notes &
Recovery of sensitive medical text\\
\specialrule{0.9pt}{1.2pt}{0.8pt}
\cellcolor{dataset-graph-bg}\textcolor{black}{\textbf{citation networks}} &
Cora~\citep{kipf2016semi}; CoraML~\citep{sen2008collective}; CiteSeer~\citep{kipf2016semi}; PubMed~\citep{kipf2016semi} &
One graph; 2.5K--19.7K nodes and 5.3K--44.3K edges &
Adjacency, feature, and link reconstruction\\
\specialrule{0.15pt}{0.25pt}{0.25pt}

\cellcolor{dataset-graph-bg}\textcolor{black}{\textbf{social / interaction}} &
Polblogs~\citep{adamic2005political}; USA and Brazil~\citep{ribeiro2017struc2vec}; Actor~\citep{pei2020geom}; Facebook~\citep{leskovec2012learning}; BlogCatalog~\citep{qiu2018network}; E-mail and YouTube~\citep{jure2014snap}; Bitcoin~\citep{kumar2016edge} &
One graph; 131--1.13M nodes and 2.1K--2.99M edges &
Topology recovery, link inference, and graph-embedding inversion\\
\specialrule{0.15pt}{0.25pt}{0.25pt}

\cellcolor{dataset-graph-bg}\textcolor{black}{\textbf{heterogeneous}} &
ACM~\citep{wang2019heterogeneous}; DBLP~\citep{tang2008arnetminer}; IMDB~\citep{fu2020magnn} &
One graph; 9K--18.4K nodes and 13K--67.9K edges &
Recovery of typed relations and private links\\
\specialrule{0.15pt}{0.25pt}{0.25pt}

\cellcolor{dataset-graph-bg}\textcolor{black}{\textbf{molecular / biological}} &
AIDS~\citep{riesen2008iam}; DD~\citep{dobson2003distinguishing}; ENZYMES and PROTEINS\_full~\citep{borgwardt2005protein}; PPI~\citep{qiu2018network}; NCI1, OVCAR-8H, PC3, and MOLT-4H~\citep{morris2020tudataset}; COX2 and DHFR~\citep{sutherland2003spline} &
One graph or 467--4,110 graphs; approximately 16--284 nodes per graph &
Graph-level property, structure, and attribute reconstruction\\
\specialrule{0.15pt}{0.25pt}{0.25pt}

\cellcolor{dataset-graph-bg}\textcolor{black}{\textbf{web / coauthor}} &
Wiki-CS~\citep{mernyei2020wiki}; Amazon Computers and Amazon Photos~\citep{shchur2018pitfalls}; Coauthor CS and Coauthor Physics~\citep{shchur2018pitfalls} &
One graph; 7.5K--34.5K nodes and 81.9K--248K edges &
Large-scale link, feature, and neighborhood reconstruction\\
\bottomrule
\end{tabularx}
\end{minipage}
\end{lrbox}
\revise{\usebox{\compactdatasetrevisionbox}}

\end{table}

% Declared here so the wide float can occupy the top of the following page,
% where Sec. 8 discusses and cites it.
\begin{table*}[t]
\centering
\tabgap
\caption{\revise{Deployment interfaces, operational constraints, and residual reconstruction risks. TEE denotes a trusted execution environment.}}
\Description[Deployment matrix for model inversion]{Comparison of exposed interfaces, relevant attacks, costs, defenses, and residual risks in cloud, edge, split, and federated environments.}
\label{tab:deployment-matrix}
\setlength{\tabcolsep}{2pt}
\renewcommand{\arraystretch}{0.88}
\fontsize{6.15}{6.55}\selectfont
{\revcolor
\begin{tabular}{@{}L{46pt}L{64pt}L{72pt}L{72pt}L{76pt}L{76pt}@{}}
\toprule
\textbf{Environment} & \textbf{Interface} & \textbf{Attacks} & \textbf{Constraint} & \textbf{Controls} & \textbf{Residual risk} \\
\midrule
Cloud/\allowbreak MLaaS~\citep{song2022survey} & labels, scores, text, embeddings & output/embedding inversion & rate/API limits, truncation, monitoring & restrict/perturb outputs; monitor abuse & adaptive low-query attacks; utility loss \\
\midrule
Edge/on-device~\citep{sun2020mobilebert} & parameters, embeddings, runtime & white-box/representation inversion & device compute; protected execution & TEE/model protection; minimize representations & extracted parameters; exposed boundary \\
\midrule
Collaborative/split~\citep{he2019model} & smashed data, activations, gradients & passive/active reconstruction & cut, auxiliary pairs, deviation, detection & protect representations; check active servers & architecture/task dependence; priors \\
\midrule
Federated/\allowbreak decentralized~\citep{hu2020personalized} & per-round gradients/activations & gradient/activation reconstruction & rounds, batch, aggregation, participation & DP; secure aggregation; validation & utility loss; malicious server; raw messages \\
\bottomrule
\end{tabular}}

\end{table*}

\subsection{Evaluation Metrics}

% This section first introduces the general evaluation metrics applicable across all fields used for assessing target models. Subsequently, we provide detailed explanations of specific evaluation metrics tailored for analyzing attacks and defenses within the image, text, and graph domains, respectively.

\revise{We group metrics by fidelity, identity/semantics, distributional realism, privacy risk/attack success, and utility, as reported by the studies surveyed in Secs.~\ref{sec: MIAs-image}--\ref{sec: MIAs-graph} and behind the privacy--utility evidence in Tab.~\ref{tab:defense-tradeoff}. Because no single metric establishes inversion, reports should combine target-level fidelity or semantics with attack success and utility.}

% Before exploring domain-specific metrics, it is vital to establish a baseline with general evaluation metrics applicable across all fields of model assessment. These metrics provide a comprehensive overview of model performance. This paragraph details how each of these general metrics is calculated.
% [SECOND-CUT] the four general metrics were one itemize entry each with a
% displayed fraction; merged into running text, definitions unchanged.
\textbf{General evaluation metrics.} \revise{For true positive (TP), true negative (TN), false positive (FP), and false negative (FN), \textit{accuracy (Acc.)} is $(\text{TP}+\text{TN})/(\text{TP}+\text{TN}+\text{FP}+\text{FN})$, \textit{recall (R)} is $\text{TP}/(\text{TP}+\text{FN})$, \textit{precision (P)} is $\text{TP}/(\text{TP}+\text{FP})$, and \textit{F1 score (F1)} is their harmonic mean. Reports must name the evaluator and positive event, e.g.\ identity, private-token, or private-link recovery.}

% This paragraph introduces essential metrics for assessing image quality and integrity. 
% [SECOND-CUT] every definition, formula, and citation below is preserved; the
% surrounding textbook exposition is condensed and the list spacing tightened.
\textbf{Image-specific evaluation metrics.} \revise{\textbf{Pixel, feature, and identity fidelity.} \textit{Mean squared error (MSE)} is squared pixel error; \textit{peak signal-to-noise ratio (PSNR)} is its logarithmic ratio; and \textit{pixelwise similarity} is \(1-\mathrm{MSE}\) after \([0,1]\) normalization. \textit{Feature distance} is \(L_2\) distance between pixel vectors or, more commonly, evaluation-model embeddings; \textit{$K$-nearest-neighbor distance} averages distances to the \(K\) nearest references. Identity-evaluator Acc.\ measures recognizability, not pixel fidelity. \textit{Fr\'{e}chet Inception Distance (FID)}~\citep{heusel2017gans} compares feature distributions, not target recovery. \textit{Learned Perceptual Image Patch Similarity (LPIPS)}~\citep{zhang2018unreasonable} is deep-feature distance; \textit{Structural Similarity Index Measure (SSIM)}~\citep{wang2004image} measures image structure; \textit{Natural Image Quality Evaluator (NIQE)}~\citep{mittal2013niqe} measures no-reference naturalness; and the \textit{Contrastive Language--Image Pre-training score (CLIP Score)} is embedding cosine similarity. Distance metrics are minimized, similarity metrics maximized.}

% This paragraph provides a detailed overview of metrics designed to evaluate the MIA and defense performances in the text domain. 
% [SECOND-CUT] definitions, formulas, and citations preserved; exposition condensed.
\textbf{Text-specific evaluation metrics.} \revise{\textbf{Lexical overlap and likelihood.} \textit{Bilingual Evaluation Understudy (BLEU)}~\citep{papineni2002bleu} combines modified \(n\)-gram precision and a brevity penalty; \textit{Recall-Oriented Understudy for Gisting Evaluation (ROUGE)}~\citep{lin2004rouge}, including ROUGE-N, measures reference-content recall. \textit{Perplexity (PPL)} is exponentiated negative autoregressive log-likelihood, and \textit{pseudo-log-likelihood (PLL)} is its masked-language-model analogue; both measure fluency, not privacy recovery. \textit{Named Entity Recovery Ratio (NERR)}, \textit{Stop Word Ratio}~\citep{li2023sentence}, and \textit{Recovery Rate (RR)}~\citep{zhang2022text} quantify recovered content. \textit{Hamming distance per token}~\citep{hamming1950error} measures token mismatch, \textit{cosine similarity (CS)} compares representations, and \textit{large-language-model evaluation (LLM-Eval)}~\citep{lin2023llm} assesses appropriateness, content, grammar, and relevance. Since paraphrases can have low overlap, lexical metrics should be paired with content-recovery and semantic measures.}

% This paragraph focuses on key metrics used within the graph domain to assess the integrity and accuracy of graph-based data structures.
% [SECOND-CUT] definitions, formulas, and citations preserved; exposition condensed.
\textbf{Graph-specific evaluation metrics.} \revise{\textbf{Link and structure metrics.} \textit{Area under the receiver operating characteristic curve (AUC)} ranks recovered links across thresholds; \textit{average precision (AP)} summarizes the precision--recall curve and is more informative under class imbalance, with an edge as the positive event. \textit{Joint degree distribution} gives endpoint degrees, and \textit{relative Frobenius error} \(\|A-\hat A\|_F/\|A\|_F\) compares adjacency or \textit{positive pointwise mutual information (PPMI)} matrices. Agreement between true and reconstructed graphs is also reported as relative graph-property error, cosine similarity, Wasserstein or Jensen--Shannon distance, or the Weisfeiler--Lehman (WL) kernel. Aggregate similarity does not prove exact sensitive-edge recovery, so link-level and structural metrics should be paired.}

\revise{\textbf{Comparability and evaluator bias.} Evaluator false positives, miscalibration, and domain shift can overstate recovery. Results are not comparable across evaluators, preprocessing, targets, attacks, splits, or privacy budgets. A defensible protocol pairs identity Acc. with target-level and distributional image quality; RR/NERR with overlap, semantics, and fluency for text; or AUC/AP with structural similarity for graphs, while reporting utility and cost.}

\FloatBarrier % keep the dataset tables and evaluation floats inside Sec. 7
\section{Further Discussions on Future Directions}
\label{sec: discussion}

% So far, we have introduced a well-defined research problem and discussed several effective objectives and principles, as well as advanced solutions in each domain. In general, the core idea of conducting the MIAs is to utilize the prior knowledge as much as possible to extract more information from the target model and then generate more realistic samples that reveal the training data. In contrast, defending against the MIAs aims to store less information about training data in the model, in which way the adversary can have a hard time recovering the private data from the target model. 
\revise{\textbf{Problem Settings and Advanced Techniques.} Open problems include imperfect side information, generative, reinforcement, and federated learning, and fine-grained attribute recovery. Optimization and generative priors require evaluation under domain knowledge, query limits, efficiency, and legitimate utility in deployment.}

\revise{\textbf{Deployment Environments.} Cloud, edge, split, and federated systems expose outputs, parameters/state, features, and gradients/activations. Tab.~\ref{tab:deployment-matrix} maps them to constraints, controls, and residual risk: MLaaS output perturbation does not protect split features, and secure aggregation does not protect released embeddings in practice.}

\revise{\textbf{Cost and Query-Budget Constraints.} Optimization-based attacks pay per target, whereas training-based attacks amortize auxiliary-data and training costs. Neither is uniformly costlier. Reports should include target count, iterations/samples, surrogate training, hardware/API cost, rate limits, wall-clock time, GPU-hours, queries, reconstruction count, and quality for reproducibility (\eg SMILE~\citep{li2025headtotail}, BREP-MI~\citep{BREPMI}, and LOKT~\citep{LOKT}).}
\revise{Detectability is a further constraint. High-volume query campaigns are visible at the API boundary as rate and input-distribution anomalies; an active split-inference server deviates from the protocol in auditable ways; and assembling an auxiliary corpus may itself be costly or traceable. Whether a defender can flag inversion queries without hurting legitimate users remains open.}

% Model inversion attacks work by exploiting the fact that machine learning models, especially those using techniques like deep learning, learn complex relationships between inputs and outputs during training. By carefully probing the model's predictions, an adversary may be able to "invert" this process and recover plausible inputs that could have generated the observed outputs. The specific techniques for carrying out model inversion can vary but often involve optimization-based approaches that search for inputs that minimize the discrepancy between the model's predictions and the target output.

% Defending against model inversion is an active area of research in machine learning security. Potential mitigation strategies include differential privacy techniques to obscure the training data, model obfuscation methods, and careful monitoring of model outputs to detect suspicious activity. As machine learning systems become more ubiquitous, understanding and addressing model inversion attacks will be crucial for preserving individual privacy.

\section{Conclusion}
\label{conclusion}

\revise{This survey has organized model inversion attacks and defenses around the four questions that decide whether reconstruction succeeds: which interface the adversary observes, what knowledge and prior it may assume, what it can recover, and under what conditions it fails. Across the image, text, and graph domains, reconstruction quality follows the strength of the generative prior and the richness of the exposed interface, and the interfaces worth defending now reach beyond released classifiers to split features, shared embeddings, and vision-language responses. Defenses have not kept pace, and each trades exposed information against accuracy, fidelity, or computation. Progress will depend on improving that trade-off, and on evaluation that reflects the deployment and query budgets an attacker actually faces.}

%%
%% The acknowledgments section is defined using the "acks" environment
%% (and NOT an unnumbered section). This ensures the proper
%% identification of the section in the article metadata, and the
%% consistent spelling of the heading.

% \begin{acks}
% To Robert, for the bagels and explaining CMYK and color spaces.
% \end{acks}

%%
%% The next two lines define the bibliography style to be used, and
%% the bibliography file.

% Keep URLs available in the bibliography; do not suppress them.
% Render the DOI/official URL as a single clickable word "URL".
\def\showURL#1{\begingroup\def\url##1{\href{##1}{URL}}#1\endgroup}
% Bibliography formatting is left exactly as the acmart Surveys style defines
% it. Earlier drafts overrode \bibfont (\scriptsize), \linespread and \bibsep to
% save pages; those are deviations from the required style and were removed.
% Do not re-add them: the 35-page limit is defined "when formatted using the
% Surveys style", so page savings must come from content, not from formatting.
% Record where the bibliography begins for navigation and local checks.
\label{loc:references}
\bibliographystyle{ACM-Reference-Format}
\bibliography{references}

\end{document}